\documentclass{elsarticle}
\makeatletter
\def\ps@pprintTitle{%
 \let\@oddhead\@empty
 \let\@evenhead\@empty
 \def\@oddfoot{\parbox{\textwidth}{Extended preprint of a paper accepted for publication in \emph{Pattern Recognition}. © 2021. This manuscript version is made available under the CC-BY-NC-ND 4.0 license \hbox{\url{http://creativecommons.org/licenses/by-nc-nd/4.0/.}}}\def\@oddfoot{\centerline{\thepage}}}%
 \def\@evenfoott{\centerline{\thepage}}}
\makeatother
\usepackage{url}
\usepackage{amsmath,amssymb,amsfonts}
\usepackage{enumitem}
\usepackage{graphicx}
\usepackage{textcomp}
\usepackage{xcolor}
\usepackage{tikz}
\usepackage{ifthen}
\usepackage{subcaption}
\usepackage{multirow}
\usepackage[colorlinks=true]{hyperref}
\usepackage{float}
\usepackage{algorithm}
\usepackage{algorithmicx}
\usepackage{algpseudocode}

\DeclareMathOperator*{\argmin}{arg\,min}

\algnewcommand{\IfThen}[2]{
  \State \algorithmicif\ #1\ \algorithmicthen\ #2}

\newcounter{StartValA}
\newcounter{EndValA}

\newcounter{ColourFlag}
\newcounter{Sum}
\newcounter{Min}
\newcounter{Max}
\newcounter{MinI}
\newcounter{MinJ}
\newcounter{Val}
\newboolean{IsValid}

\newcounter{arraycard}
\def\arrayLength#1{%
\setcounter{arraycard}{0}%
\foreach \x in #1{%
\stepcounter{arraycard}%
}%
 \the\value{arraycard}%
 } 

\def\dist#1#2{int(abs(#1-#2)*abs(#1-#2))}

\newcounter{N}			
\newcounter{M}			

\def\drawmatrix{
	\node at (\N/2,1.5) {$A$};
	\node at (-1.5,-\M/2) {$B$};
          \foreach \i in {0,...,\value{N}} {
		\pgfmathsetmacro{\val}{int(\A[\i])}
		\node at (\i+.5,0.5) {\val};
		\pgfmathsetmacro{\val}{int(\B[\i])}
		\node at (-0.5,-\i-0.5) {\val};
	}
	\draw[step=1] (0,0) grid (\N,-\M);
	\ifthenelse{\W<\value{N}}{
		\setcounter{EndValA}{\M}
		\addtocounter{EndValA}{-\W}
		\addtocounter{EndValA}{-2}
		\foreach \j in {0,...,\value{EndValA}} {
			\setcounter{StartValA}{\j}
			\addtocounter{StartValA}{\W}
			\addtocounter{StartValA}{1}
			\foreach \i in {\value{StartValA},...,\value{N}} {
				\draw[fill=black!50] (\i,-\j) rectangle (\i+1,-\j-1);
		}}
		\setcounter{StartValA}{\W}
		\addtocounter{StartValA}{1}
		\foreach \j in {\value{StartValA},...,\value{M}} {
			\setcounter{EndValA}{\j}
			\addtocounter{EndValA}{-\W}
			\addtocounter{EndValA}{-1}
			\foreach \i in {0,...,\value{EndValA}} {
				\draw[fill=black!50] (\i,-\j) rectangle (\i+1,-\j-1);
		}}
	}{}
	\foreach \i in {0,...,\value{N}} {
		\setcounter{StartValA}{\i}
		\addtocounter{StartValA}{-\W}
		\ifthenelse{\value{StartValA}<0}{\setcounter{StartValA}{0}}{}
		\setcounter{EndValA}{\i}
		\addtocounter{EndValA}{\W}
		\ifthenelse{\value{EndValA}>\value{M}}{\setcounter{EndValA}{\value{M}}}{}
		\foreach \j in {\value{StartValA},...,\value{EndValA}} {
			\pgfmathtruncatemacro{\distance}{\dist{\A[\i]}{\B[\j]}}
			\node at (\i+.5,-\j-.5)  {\distance};
		}
	}
}

\def\drawcostmatrix{
	\node at (\N/2,1.5) {$A$};
	\node at (-1.5,-\M/2) {$B$};
	\pgfmathtruncatemacro{\NX}{int(\N-1)}
          \foreach \i in {0,...,\NX} {
		\pgfmathsetmacro{\val}{int(\A[\i])}
		\node at (\i+.5,0.5) {\val};
		\pgfmathsetmacro{\val}{int(\B[\i])}
		\node at (-0.5,-\i-0.5) {\val};
	}
	\draw[step=1] (0,0) grid (\N,-\M);
	\foreach \i in {0,...,\NX} {
		\foreach \j in {0,...,\NX} {
			\pgfmathsetmacro{\thisval}{\thematrix[\j][\i]}
			\ifthenelse{\thisval=999}{
				\draw[fill=black!50] (\i,-\j) rectangle (\i+1,-\j-1);
			}{
				\node at (\i+.5,-\j-.5)  {\thisval};
			}
		}
	}
}

\def\dotos#1#2{
	\ifthenelse{\value{ColourFlag}=0}{
		\gdef\backgroundcolour{blue!20}
	}{
		\gdef\backgroundcolour{green!20}
	}
	\ifthenelse{#1=0}{
		\def\i{0}
		\def\j{0}
		\draw[fill=\backgroundcolour] (\i,-\j)  rectangle (1+\i,-1-\j);
		\pgfmathsetmacro{\distance}{\dist{\A[\i]}{\B[\j]}}
		\setcounter{Min}{\distance}
		\node at (\i+.5,-\j-.5)  {\bf \distance};
	}{
		\setcounter{Val}{#1}
		\setcounter{MinI}{-1}
		\setcounter{MinJ}{1}
		\pgfmathsetmacro{\firsti}{int(max(#1-\W,0))}
		\pgfmathsetmacro{\lastj}{int(min(#1,\M))}
		\setboolean{IsValid}{true}
		\ifthenelse{\equal{#2}{mm}}{
			\setcounter{Min}{9999}
			\setcounter{Max}{-9999}
			\foreach \i in {\firsti,...,\value{Val}} {
				\pgfmathsetmacro{\Ai}{\A[\i]}
				\ifthenelse{\Ai<\value{Min}}{\setcounter{Min}{\Ai}}{}
				\ifthenelse{\Ai>\value{Max}}{\setcounter{Max}{\Ai}}{}
			}
			\pgfmathsetmacro{\Bval}{\B[\value{Val}]}
			\ifthenelse{\Bval>\value{Max}}{
			}{
				\ifthenelse{\Bval<\value{Min}}{
				}{
					\setboolean{IsValid}{false}				
				}
			}
			\setcounter{Min}{9999}
			\setcounter{Max}{-9999}
			\foreach \j in {\firsti,...,\lastj} {
				\pgfmathsetmacro{\Bj}{\B[\j]}
				\ifthenelse{\Bj<\value{Min}}{\setcounter{Min}{\Bj}}{}
				\ifthenelse{\Bj>\value{Max}}{\setcounter{Max}{\Bj}}{}
			}
			\pgfmathsetmacro{\Aval}{\A[\value{Val}]}			
			\ifthenelse{\Aval>\value{Max}}{
			}{
				\ifthenelse{\Aval<\value{Min}}{
				}{
					\setboolean{IsValid}{false}				
				}
			}
		}{
		}
		\ifthenelse{\boolean{IsValid}}{
		}{
			\gdef\backgroundcolour{black!15}
		}
		\setcounter{Min}{9999}
		\let\j\lastj
		\foreach \i in {\firsti,...,\value{Val}} {
			\draw[fill=\backgroundcolour] (\i,-\j)  rectangle (1+\i,-1-\j);
			\pgfmathsetmacro{\distance}{\dist{\A[\i]}{\B[\j]}}
			\ifthenelse{\distance<\value{Min}}{\setcounter{Min}{\distance}\setcounter{MinI}{\i}\setcounter{MinJ}{\j}}{}
			\node at (\i+.5,-\j-.5)  {\distance};
		}
		\def\i{\value{Val}}
		\foreach \j in {\firsti,...,\lastj} {
			\draw[fill=\backgroundcolour] (\i,-\j)  rectangle (1+\i,-1-\j);
			\pgfmathsetmacro{\distance}{\dist{\A[\i]}{\B[\j]}}
			\ifthenelse{\distance<\value{Min}}{\setcounter{Min}{\distance}\setcounter{MinI}{\i}\setcounter{MinJ}{\j}}{}
			\node at (\i+.5,-\j-.5)  {\distance};
		}
		\ifthenelse{\boolean{IsValid}}{
			\node at (\value{MinI}+.5,-\value{MinJ}-.5)  {\bf \arabic{Min}};
		}{
			\setcounter{Min}{0}
		}
	}
	\ifthenelse{\value{ColourFlag}=0}{\setcounter{ColourFlag}{1}}{\setcounter{ColourFlag}{0}}
	\addtocounter{Sum}{\value{Min}}
}

\def\dolos#1#2{
	\ifthenelse{\value{ColourFlag}=0}{
		\gdef\backgroundcolour{blue!20}
	}{
		\gdef\backgroundcolour{green!20}
	}
	\pgfmathsetmacro{\d}{\N-\M}	
	\setcounter{Val}{#1}
	\ifthenelse{\value{Val}=\N}{
		\let\i\N
		\let\j\N
		\draw[fill=\backgroundcolour] (\i,-\j)  rectangle (1+\i,-1-\j);
		\pgfmathsetmacro{\distance}{\dist{\A[\i]}{\B[\j]}}
		\setcounter{Min}{\distance}
		\node at (\i+.5,-\j-.5)  {\bf \distance};
	}{
		\setcounter{Min}{1000}
		\setcounter{MinI}{-1}
		\setcounter{MinJ}{1}
		\pgfmathsetmacro{\firstj}{int(max(0,\value{Val}-\d))}
		\pgfmathsetmacro{\lastj}{int(min(#1+\W,\value{M}))}
		\pgfmathsetmacro{\lasti}{int(min(\firstj+\W,\value{N}))}
		\setboolean{IsValid}{true}
		\ifthenelse{\equal{#2}{mm}}{
			\setcounter{Min}{9999}
			\setcounter{Max}{-9999}
			\foreach \i in {\value{Val},...,\lasti} {
				\pgfmathsetmacro{\Ai}{\A[\i]}
				\ifthenelse{\Ai<\value{Min}}{\setcounter{Min}{\Ai}}{}
				\ifthenelse{\Ai>\value{Max}}{\setcounter{Max}{\Ai}}{}
			}
			\pgfmathsetmacro{\Bval}{\B[\value{Val}+\M-\N]}
			\ifthenelse{\Bval>\value{Max}}{
			}{
				\ifthenelse{\Bval<\value{Min}}{
				}{
					\setboolean{IsValid}{false}				
				}
			}
			\setcounter{Min}{9999}
			\setcounter{Max}{-9999}
			\foreach \j in {\firstj,...,\lastj} {
				\pgfmathsetmacro{\Bj}{\B[\j]}
				\ifthenelse{\Bj<\value{Min}}{\setcounter{Min}{\Bj}}{}
				\ifthenelse{\Bj>\value{Max}}{\setcounter{Max}{\Bj}}{}
			}
			\pgfmathsetmacro{\Aval}{\A[\value{Val}]}			
			\ifthenelse{\Aval>\value{Max}}{
			}{
				\ifthenelse{\Aval<\value{Min}}{
				}{
					\setboolean{IsValid}{false}				
				}
			}
		}{
		}
		\ifthenelse{\boolean{IsValid}}{
		}{
			\gdef\backgroundcolour{black!15}
		}
		\setcounter{Min}{9999}
		\let\j\firstj
		\foreach \i in {\value{Val},...,\lasti} {
			\draw[fill=\backgroundcolour] (\i,-\j)  rectangle (1+\i,-1-\j);
			\pgfmathsetmacro{\distance}{\dist{\A[\i]}{\B[\j]}}
			\ifthenelse{\distance<\value{Min}}{\setcounter{Min}{\distance}\setcounter{MinI}{\i}\setcounter{MinJ}{\j}}{}
			\node at (\i+.5,-\j-.5)  {\distance};
		}
		\pgfmathsetmacro{\i}{\value{Val}}
		\foreach \j in {\firstj,...,\lastj} {
			\draw[fill=\backgroundcolour] (\i,-\j)  rectangle (1+\i,-1-\j);
			\pgfmathsetmacro{\distance}{\dist{\A[\i]}{\B[\j]}}
			\ifthenelse{\distance<\value{Min}}{\setcounter{Min}{\distance}\setcounter{MinI}{\i}\setcounter{MinJ}{\j}}{}
			\node at (\i+.5,-\j-.5)  {\distance};
		}
		\ifthenelse{\boolean{IsValid}}{
			\node at (\value{MinI}+.5,-\value{MinJ}-.5)  {\bf \arabic{Min}};
		}{
			\setcounter{Min}{0}
		}
	}
	\ifthenelse{\value{ColourFlag}=0}{\setcounter{ColourFlag}{1}}{\setcounter{ColourFlag}{0}}
	\pgfmathsetmacro{\Min}{\value{Min}}
	\addtocounter{Sum}{\Min}
}

\def\doAos#1#2{
	\ifthenelse{\value{ColourFlag}=0}{
		\gdef\backgroundcolour{blue!20}
	}{
		\gdef\backgroundcolour{green!20}
	}
	\setcounter{Val}{#1}
	\setcounter{MinI}{-1}
	\setcounter{MinJ}{1}
	\pgfmathsetmacro{\firstj}{int(max(\value{Val}-\W,0))}
	\pgfmathsetmacro{\lastj}{int(min(\value{Val}+\W,\value{M}))}
	\pgfmathsetmacro{\i}{\value{Val}}
	\setboolean{IsValid}{true}
	\ifthenelse{\equal{#2}{mm}}{
		\setcounter{Min}{9999}
		\setcounter{Max}{-9999}
		\foreach \j in {\firstj,...,\lastj} {
			\pgfmathsetmacro{\Bj}{\B[\j]}
			\ifthenelse{\Bj<\value{Min}}{\setcounter{Min}{\Bj}}{}
			\ifthenelse{\Bj>\value{Max}}{\setcounter{Max}{\Bj}}{}
		}
		\pgfmathsetmacro{\Aval}{\A[\value{Val}]}			
		\ifthenelse{\Aval>\value{Max}}{
		}{
			\ifthenelse{\Aval<\value{Min}}{
			}{
				\setboolean{IsValid}{false}				
			}
		}
	}{
	}
	\ifthenelse{\boolean{IsValid}}{
	}{
		\gdef\backgroundcolour{black!15}
	}
	\setcounter{Min}{9999}
	\foreach \j in {\firstj,...,\lastj} {
			\ifthenelse{\value{ColourFlag}=0}{
				\draw[fill=\backgroundcolour] (\i,-\j)  rectangle (1+\i,-1-\j);
			}{
				\draw[fill=\backgroundcolour] (\i,-\j) rectangle (1+\i,-1-\j);
			}
			\pgfmathsetmacro{\distance}{\dist{\A[\i]}{\B[\j]}}
			\ifthenelse{\distance<\value{Min}}{\setcounter{Min}{\distance}\setcounter{MinI}{\i}\setcounter{MinJ}{\j}}{}
			\node at (\i+.5,-\j-.5)  {\distance};
	}
	\ifthenelse{\boolean{IsValid}}{
		\node at (\value{MinI}+.5,-\value{MinJ}-.5)  {\bf \arabic{Min}};
	}{
		\setcounter{Min}{0}
	}
	\ifthenelse{\value{ColourFlag}=0}{\setcounter{ColourFlag}{1}}{\setcounter{ColourFlag}{0}}
	\pgfmathsetmacro{\Min}{\value{Min}}
	\addtocounter{Sum}{\Min}
}

\def\doAosGrey#1{
	\gdef\backgroundcolour{black!15}
	\setcounter{Val}{#1}
	\setcounter{MinI}{-1}
	\setcounter{MinJ}{1}
	\pgfmathsetmacro{\firstj}{int(max(\value{Val}-\W,0))}
	\pgfmathsetmacro{\lastj}{int(min(\value{Val}+\W,\value{M}))}
	\pgfmathsetmacro{\i}{\value{Val}}
	\setboolean{IsValid}{true}
	\setcounter{Min}{9999}
	\foreach \j in {\firstj,...,\lastj} {
			\ifthenelse{\value{ColourFlag}=0}{
				\draw[fill=\backgroundcolour] (\i,-\j)  rectangle (1+\i,-1-\j);
			}{
				\draw[fill=\backgroundcolour] (\i,-\j) rectangle (1+\i,-1-\j);
			}
			\pgfmathsetmacro{\distance}{\dist{\A[\i]}{\B[\j]}}
			\ifthenelse{\distance<\value{Min}}{\setcounter{Min}{\distance}\setcounter{MinI}{\i}\setcounter{MinJ}{\j}}{}
			\node at (\i+.5,-\j-.5)  {\distance};
	}
	\ifthenelse{\value{ColourFlag}=0}{\setcounter{ColourFlag}{1}}{\setcounter{ColourFlag}{0}}
}

\def\doBos#1#2{
	\ifthenelse{\value{ColourFlag}=0}{
		\gdef\backgroundcolour{blue!20}
	}{
		\gdef\backgroundcolour{green!20}
	}
	\setcounter{Val}{#1}
	\setcounter{MinI}{-1}
	\setcounter{MinJ}{1}
	\pgfmathsetmacro{\firsti}{int(max(\value{Val}-\W,0))}
	\pgfmathsetmacro{\lasti}{int(min(\value{Val}+\W,\value{N}))}
	\pgfmathsetmacro{\j}{\value{Val}}
	\setboolean{IsValid}{true}
	\ifthenelse{\equal{#2}{mm}}{
		\setcounter{Min}{9999}
		\setcounter{Max}{-9999}
		\foreach \i in {\firsti,...,\lasti} {
			\pgfmathsetmacro{\Ai}{\A[\i]}
			\ifthenelse{\Ai<\value{Min}}{\setcounter{Min}{\Ai}}{}
			\ifthenelse{\Ai>\value{Max}}{\setcounter{Max}{\Ai}}{}
		}
		\pgfmathsetmacro{\Bval}{\B[\value{Val}]}
		\ifthenelse{\Bval>\value{Max}}{
		}{
			\ifthenelse{\Bval<\value{Min}}{
			}{
				\setboolean{IsValid}{false}				
			}
		}
	}{
	}
	\ifthenelse{\boolean{IsValid}}{
	}{
		\gdef\backgroundcolour{black!15}
	}
	\setcounter{Min}{9999}
	\foreach \i in {\firsti,...,\lasti} {
		\ifthenelse{\value{ColourFlag}=0}{
			\draw[fill=\backgroundcolour] (\i,-\j)  rectangle (1+\i,-1-\j);
		}{
			\draw[fill=\backgroundcolour] (\i,-\j) rectangle (1+\i,-1-\j);
		}
		\pgfmathsetmacro{\distance}{\dist{\A[\i]}{\B[\j]}}
		\ifthenelse{\distance<\value{Min}}{\setcounter{Min}{\distance}\setcounter{MinI}{\i}\setcounter{MinJ}{\j}}{}
		\node at (\i+.5,-\j-.5)  {\distance};
	}
	\ifthenelse{\boolean{IsValid}}{
		\node at (\value{MinI}+.5,-\value{MinJ}-.5)  {\bf \arabic{Min}};
	}{
		\setcounter{Min}{0}
	}
	\ifthenelse{\value{ColourFlag}=0}{\setcounter{ColourFlag}{1}}{\setcounter{ColourFlag}{0}}
	\addtocounter{Sum}{\arabic{Min}}
}

\newtheorem{observation}{Observation}
\newtheorem{proof}{Proof}
\newtheorem{theorem}{Theorem}

\def\path{\ensuremath{\mathcal{A}}}
\def\pathlength{\ensuremath{P}}

\def\reqsetcost(#1){\ensuremath{\mathcal{C}(#1)}}

\def\downrs(#1){\ensuremath{\mathcal{O}_A(#1)}}
\def\uprs(#1){\ensuremath{\mathcal{O}_B(#1)}}
\def\lowerbound(#1){\ensuremath{\mathrm{LB}(#1)}}
\def\lb(#1,#2){\ensuremath{\textrm{LB}(#1,#2)}}
\def\dtw(#1,#2){\ensuremath{\textrm{DTW}(#1,#2)}}
\def\cdtw(#1,#2){\ensuremath{\textrm{DTW}_w(#1,#2)}}
\def\leftrs(#1){\ensuremath{\mathcal{L}^w_{#1}}}
\def\rightrs(#1){\ensuremath{\mathcal{R}^w_{#1}}}

\def\leftset{\ensuremath{\mathcal{L}}}
\def\rightset{\ensuremath{\mathcal{R}}}

\def\lowerenv{\ensuremath{\mathbb{L}}}
\def\upperenv{\ensuremath{\mathbb{U}}}

\def\tslen{\ensuremath{\ell}}

\newcommand*{\Tpair}{$A$ and $B$}
\newcommand{\lbkim}{\textsc{LB\_Kim}}

\newcommand{\lbkeogh}{\textsc{LB\_Keogh}}

\newcommand{\lbimproved}{\textsc{LB\_Improved}}
\newcommand{\lbnew}{\textsc{LB\_New}}
\newcommand{\lbenhanced}{\textsc{LB\_Enhanced}}

\newcommand{\lbtight}{\textsc{LB\_Petitjean}}
\newcommand{\lbtightnolr}{\textsc{LB\_Petitjean\_NoLR}}
\newcommand{\lbfast}{\textsc{LB\_Webb}}
\newcommand{\lbfastmonotone}{\ensuremath{\textsc{LB\_Webb}^*}}
\newcommand{\projection}{\ensuremath{\Omega}}
\newcommand{\lbfastnolr}{\textsc{LB\_Webb\_NoLR}}
\newcommand{\lbenhancedfast}{\textsc{LB\_Webb\_Enhanced}}

\DeclareMathOperator{\LBKeogh}{\textsc{LB\_Keogh}}

\DeclareMathOperator{\LBImproved}{\textsc{LB\_Improved}}

\DeclareMathOperator{\DTW}{\textsc{DTW}}

\DeclareMathOperator{\freeAbove}{\mathrm{F}\raisebox{1.75pt}{$\uparrow$}\!}
\DeclareMathOperator{\freeBelow}{\mathrm{F}\!\!\downarrow}
\DeclareMathOperator{\minlrpaths}{\mathrm{MinLRPaths}}

\graphicspath{{images/}}

\def\A{{9,5,8,5,1,1,4,7,8,7,3,2}}
\def\B{{0,1,1,6,4,1,4,2,0,4,5,6}}
\def\N{12}
\def\M{12}
\def\W{4}

\makeatletter
\def\myeqno#1{\refstepcounter{equation}(\arabic{equation})\hss%
\immediate\write\@auxout{%
  \string\newlabel{#1}{{\arabic{equation}}{\thepage}{}{}{}}%
}%
}
\makeatother

\begin{document}
\begin{frontmatter}
\title{Tight lower bounds for Dynamic Time Warping}

\author{Geoffrey I.\ Webb\corref{cor1}%
}
\ead{geoff.webb@monash.edu}
\address{Monash University, Clayton, Victoria, 3800, Australia}
\author{Fran\c{c}ois Petitjean}
\ead{francois.petitjean@monash.edu}

\address{Monash University, Clayton, Victoria, 3800, Australia}
\begin{abstract}
Dynamic Time Warping ($\DTW$) is a popular similarity measure for aligning and comparing time series. Due to $\DTW$'s high computation time, lower bounds are often employed to screen poor matches.
Many alternative lower bounds have been proposed, providing a range of different trade-offs between tightness and computational efficiency.  $\lbkeogh$ provides a useful trade-off in many applications.  Two recent lower bounds, $\lbimproved$ and $\lbenhanced$, are substantially tighter than $\lbkeogh$.  All three have the same worst case computational complexity---linear with respect to series length and constant with respect to window size. We present four new $\DTW$ lower bounds in the same complexity class.  $\lbtight$ is substantially tighter than $\lbimproved$, with only modest additional computational overhead.  $\lbfast$ is more efficient than $\lbimproved$, while often providing a tighter bound. $\lbfast$ is always tighter than $\lbkeogh$. The parameter free $\lbfast$ is usually tighter than $\lbenhanced$. A 
parameterized variant, \lbenhancedfast, is always tighter than $\lbenhanced$. A further variant, $\lbfastmonotone$, is useful for some constrained distance functions. In extensive experiments, $\lbfast$ proves to be very effective for nearest neighbor search.
\end{abstract}
\end{frontmatter}

\section{Introduction}
Dynamic Time Warping ($\DTW$) is a time series similarity measure. From its origins in speech recognition \cite{sakoe1972dynamic}, it has spread to a broad spectrum of  further uses, recent examples  of which include gesture recognition \cite{cheng2016image},  signature verification \cite{OKAWA2020107227}, shape matching \cite{yasseen2016shape}, road surface monitoring \cite{singh2017smart},  neuroscience \cite{cao2016real} and medical diagnosis \cite{varatharajan2018wearable}.  DTW measures similarity by summing pairwise-distances between  points in two series.  However, it allows flexibility in which points are aligned, to adjust for the way related events may unfold at different paces.

In such applications, lower bounding is often employed to discard potential candidate matches without need to compute the full $\DTW$ measure \cite{ratanamahatana2005three}.  Numerous such lower bounds have been derived \cite{keogh2005exact,kim2001index,lemire2009faster,shen2018accelerating,yi1998efficient,TanEtAl19}.  These provide differing trade-offs between computational cost and tightness, ranging from loose constant time $\lbkim$ \cite{kim2001index} to the relatively tight $\lbnew$ \cite{shen2018accelerating}, which requires $O(\tslen\cdot w)$ memory and  $O(\tslen\cdot \log w)$ time to compute a lower bound for a pair of series, where $\tslen$ is the length of the series and $w$ is the window size.  These differing trade-offs will each be useful in different applications.  The~tighter the bound, the less frequent the need to compute the full $\DTW$ distance. However, the more compute resource needed to compute the bound, the lower the saving if $\DTW$ is not computed.

\begin{figure}
\begin{minipage}{0.475\columnwidth}%
\includegraphics[width=\columnwidth]{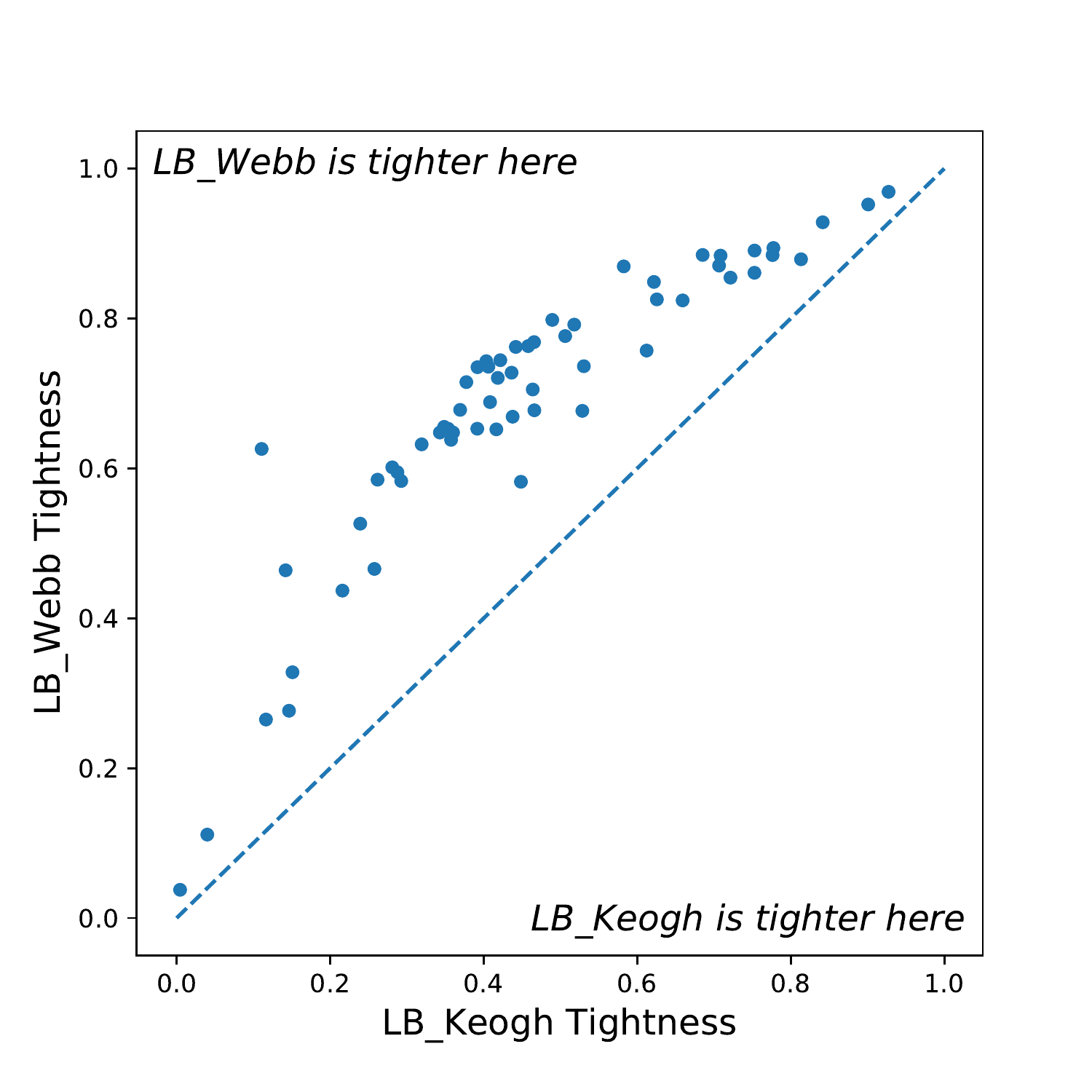}\vspace*{-10pt}
\caption{Relative tightness of $\lbfast$ and $\lbkeogh$}\label{fig:tightKeoghFast}
\end{minipage}
\begin{minipage}{0.475\columnwidth}%
\includegraphics[width=\columnwidth]{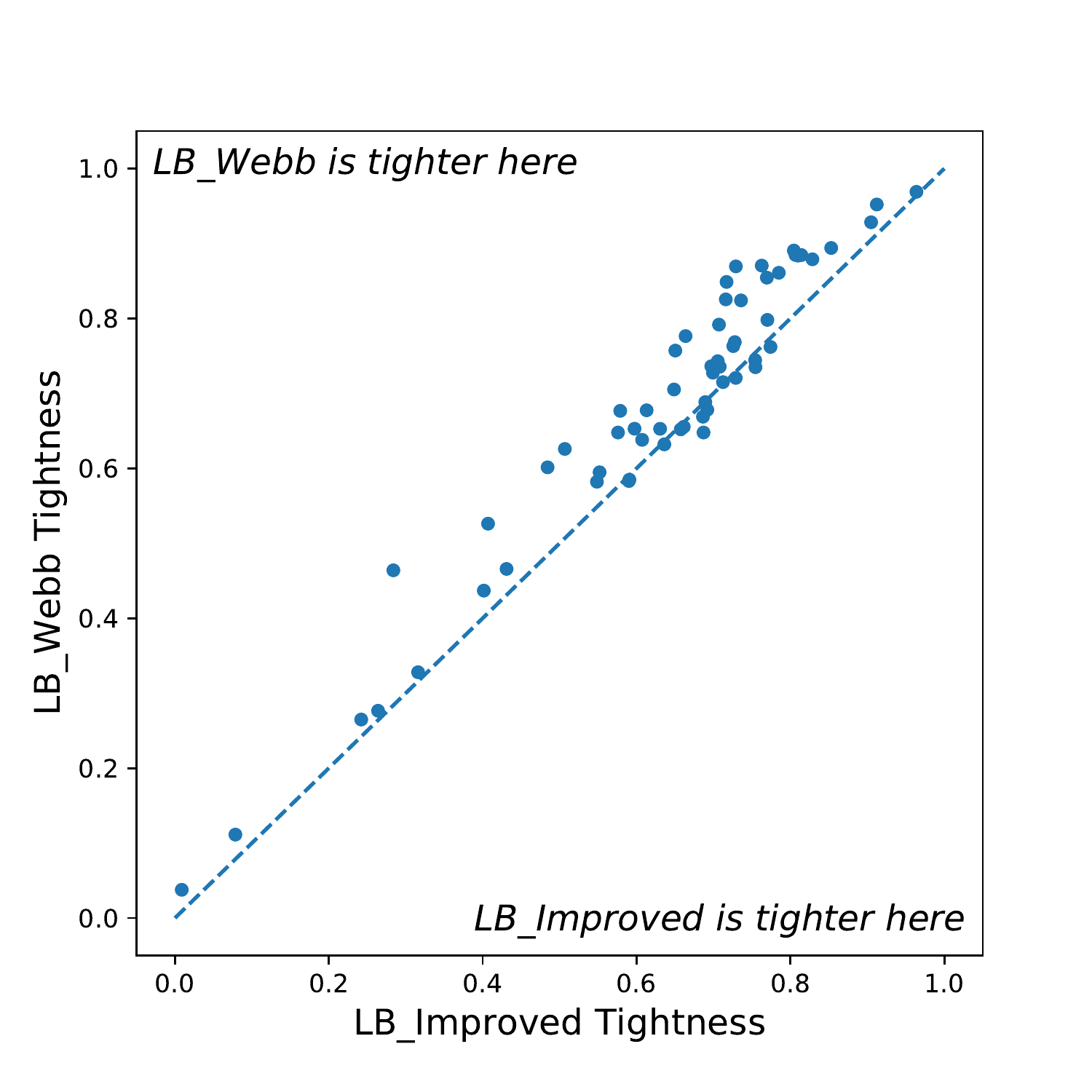}\vspace*{-10pt}
\caption{Relative tightness of $\lbfast$ and $\lbimproved$}\label{fig:tightImprovedWebb}
\end{minipage}\hfil
\end{figure}

This paper presents four new bounds.  To the best of our knowledge, the first of these new bounds, $\lbtight$, is the tightest known $\DTW$ lower bound that has linear complexity with respect to series length and constant complexity with respect to window size.
The second, $\lbfast$, belongs to the same complexity class, but is nonetheless substantially faster.  Less tight than $\lbtight$, $\lbfast$ is always tighter than $\lbkeogh$, often substantially so (see Figure \ref{fig:tightKeoghFast}), and usually tighter than the previous tightest lower bound in the complexity class, $\lbimproved$ (see Figure \ref{fig:tightImprovedWebb}).
The~third, $\lbenhancedfast$, is a variant of $\lbfast$ that may be useful in the context of large window sizes. The fourth, $\lbfastmonotone$, is a variant of $\lbfast$ suited to some specific pairwise distance distance functions.

The paper is organized as follows.  Section \ref{sec:description} describes $\DTW$.  Section~\ref{sec:bounds} describes key existing bounds.  Section~\ref{sec:petitjean} introduces the first of the new bounds, $\lbtight$ and Section \ref{sec:webb} the second, $\lbfast$ and its variants, $\lbfastmonotone$ and $\lbenhancedfast$. We provide proofs that they are $\DTW$ lower bounds and algorithms for calculating them.  Section \ref{sec:experiments} presents experimental evaluation of the utility of these bounds.  We first compare their tightness to that of key existing bounds.  We next compare their practical value for nearest neighbor search.  We end with discussion and conclusions.

\section{Problem description}\label{sec:description}
\begin{figure}
\centering
\includegraphics[width=1.0\columnwidth]{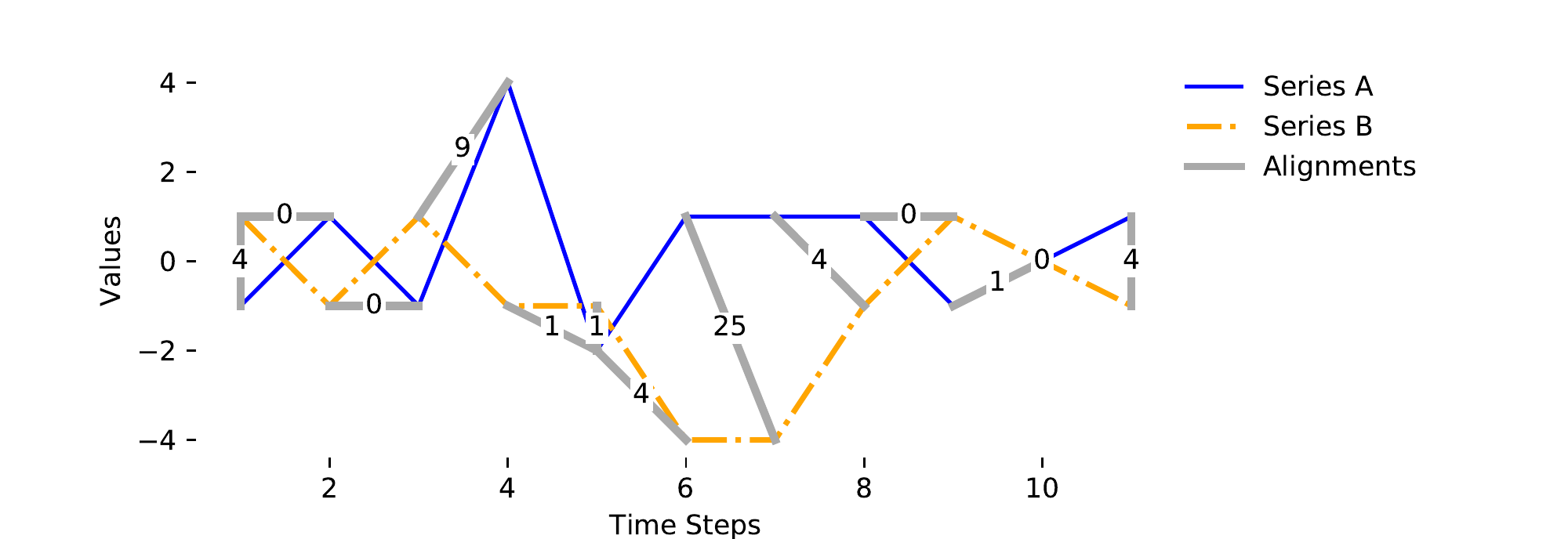}\vspace*{-5pt}
\caption{$\DTW$ warping path for time series $A=\langle -1,1,-1, 4, -2,1,1,1, -1,0,1 \rangle$ and $B=\langle 1,-1,1,-1,-1,-4,-4,-1,1,0,-1 \rangle$ with window $w=1$ and $\delta(A_i,B_j)=(A_i-B_j)^2$. Each alignment is labeled with the distance between the elements that are aligned. The DTW distance is the  sum of these distances (52).}\label{fig:dtw}
\end{figure}

$\DTW$ is a similarity measure  for aligning and comparing time series  \cite{sakoe1972dynamic}.
$\DTW$ finds the global \emph{alignment} of  time series $A=\langle A_1,\ldots, A_{\tslen}\rangle$ and $B=\langle B_1, \ldots,B_{\tslen}\rangle$, as illustrated in Figure~\ref{fig:dtw}.
For ease of exposition, we assume $A$ and $B$ are of the same length. However, it is trivial to extend this work to the case of different length series.
A \emph{warping path} of $A$ and $B$ is a sequence $\path=\langle\path_1,\ldots,\path_\pathlength\rangle$ of \emph{alignments}. 
Each alignment is a pair $\path_k=( i,j)$ indicating that $A_i$ is aligned with $B_j$. 
$\path$~must obey the following constraints: 
\begin{itemize}
  \item {\bf Boundary Conditions}: $\path_1=( 1,1)$ and $\path_\pathlength=(\tslen,\tslen)$.
  \item {\bf Continuity} and {\bf Monotonicity}: for any $\path_k=( i,j)$, $1<k\leq\pathlength$, $\path_{k-1}\in\{( i{-}1,j), ( i,j{-}1), ( i{-}1,j{-}1)\}$.
\end{itemize}

The cost \dtw(A,B)  for series \Tpair{} is the minimal cost of any warping path and is given in Equation~\ref{eqn:cost}, 
where $\delta(A_i,B_j)$ represents the cost of aligning the two elements. Two common such functions are $\delta(A_i,B_j)=|A_i-B_j|$ and $\delta(A_i,B_j)=(A_i-B_j)^2$. $\dtw(A,B) =
D(A_\tslen,B_\tslen)$. 
\begin{equation} \label{eqn:cost}
    D(A_i, B_j) =
    \begin{cases}
	\delta(A_i, B_j)&\mathrm{if~}i=1\wedge j=1\\[5pt]
	\delta(A_i, B_j) +  D(A_i,B_{j{-}1})&\mathrm{if~}i = 1 \wedge1< j\leq\tslen\\[5pt]
	\delta(A_i, B_j) + D(A_{i{-}1},B_{j}) &\mathrm{if~}1< i\leq\tslen  \wedge j=1\\[5pt]
	\delta(A_i, B_j) + \min\left[
       	\begin{array}{l}
		 D(A_{i{-}1},B_{j-1}),\\
       		 D(A_i,B_{j{-}1}),\\
        		D(A_{i{-}1},B_{j}) 
	\end{array}
    	\right]&\mathrm{if~}1< i\leq\tslen  \wedge1< j\leq\tslen.
    \end{cases} 
\end{equation}
The path with minimal cost can be found using dynamic programming by building a cost matrix $D$. 
Each cell $(i,j)$ of the matrix records the minimum cost of aligning $\langle A_1,\ldots,A_i\rangle$ and $\langle B_1,\ldots,B_j\rangle$.

Windowing adds a further constraint, that $A_i$ may only be aligned with $B_j$ if $i-w\leq j \leq i+w$, where $w\in\mathbb{N}$ is the {\it window}.
$\cdtw(A,B) =D_w(A_\tslen,B_\tslen)$ where,
\begin{equation*}
    D_w(A_i, B_j) =
    \begin{cases}
	\delta(A_i, B_j)&\mathrm{if~}i=1\wedge j=1\\[5pt]
	\delta(A_i, B_j) +  D(A_i,B_{j{-}1})&\mathrm{if~}i = 1 \wedge1< j\leq w+1\\[5pt]
	\delta(A_i, B_j) + D(A_{i{-}1},B_{j}) &\mathrm{if~}1< i\leq w+1  \wedge j=1\\[5pt]
	\delta(A_i, B_j) + \min\left[
       	\begin{array}{l}
		 D(A_{i{-}1},B_{j-1}),\\
       		 D(A_i,B_{j{-}1})
	\end{array}
    	\right]&\mathrm{if~}i=j+w  \wedge1< j\leq\tslen\\[12pt]
	\delta(A_i, B_j) + \min\left[
       	\begin{array}{l}
		 D(A_{i{-}1},B_{j-1}),\\
        		D(A_{i{-}1},B_{j}) 
	\end{array}
    	\right]&\mathrm{if~}1< i\leq\tslen  \wedge  j=i+w\\[8pt]
	\delta(A_i, B_j) + \min\left[
       	\begin{array}{l}
		 D(A_{i{-}1},B_{j-1}),\\
       		 D(A_i,B_{j{-}1}),\\
        		D(A_{i{-}1},B_{j}) 
	\end{array}
    	\right]&\begin{array}{l}\mathrm{if~}1< i<j+w\\~~\wedge\,1< j<i+w.\end{array}
    \end{cases} 
\end{equation*}

Figure \ref{fig:costmatrix} shows the cost matrix corresponding to the warping path with window $w=1$ illustrated in Figure \ref{fig:dtw}.
\begin{figure}
\def\A{{-1,1,-1, 4, -2,1,1,1, -1,0,1}}
\def\B{{1,-1,1,-1,-1,-4,-4,-1,1,0,-1}}
\def\N{11}
\def\M{11}
\def\W{1}
\def\thematrix{{{4,0,999,999,999,999,999,999,999,999,999},{0,4,0,999,999,999,999,999,999,999,999},{999,0,4,9,999,999,999,999,999,999,999},{999,999,0,25,1,999,999,999,999,999,999},{999,999,999,25,1,4,999,999,999,999,999},{999,999,999,999,4,25,25,999,999,999,999},{999,999,999,999,999,25,25,25,999,999,999},{999,999,999,999,999,999,4,4,0,999,999},{999,999,999,999,999,999,999,0,4,1,999},{999,999,999,999,999,999,999,999,1,0,1},{999,999,999,999,999,999,999,999,999,1,4}}}

\pgfmathsetcounter{N}{\N-1}
\pgfmathsetcounter{M}{\M-1}
    \centering\footnotesize
		\begin{tikzpicture}[scale=.35]
		\drawcostmatrix
		\end{tikzpicture}
    \vspace*{-2pt}
    \caption{A cost matrix for calculating DTW with window $w=1$.}
    \label{fig:costmatrix}
\end{figure}

The time complexity of calculating DTW with window $w$ is $O(\tslen\cdot w)$.  While this is linear on both $\tslen$ and $w$, when both are relatively large the total computation can be prohibitive for the many repetitions that are entailed by operations such as nearest neighbor search. To this end, it is often desirable to employ  lower bounds, such as the popular $\lbkeogh$ with complexity $O(\tslen)$. These allow some potential nearest neighbors to be discarded without recourse to the expensive process of computing DTW.

\section{Key existing bounds}\label{sec:bounds}
\begin{figure}
\begin{center}
\includegraphics[width=1.0\columnwidth]{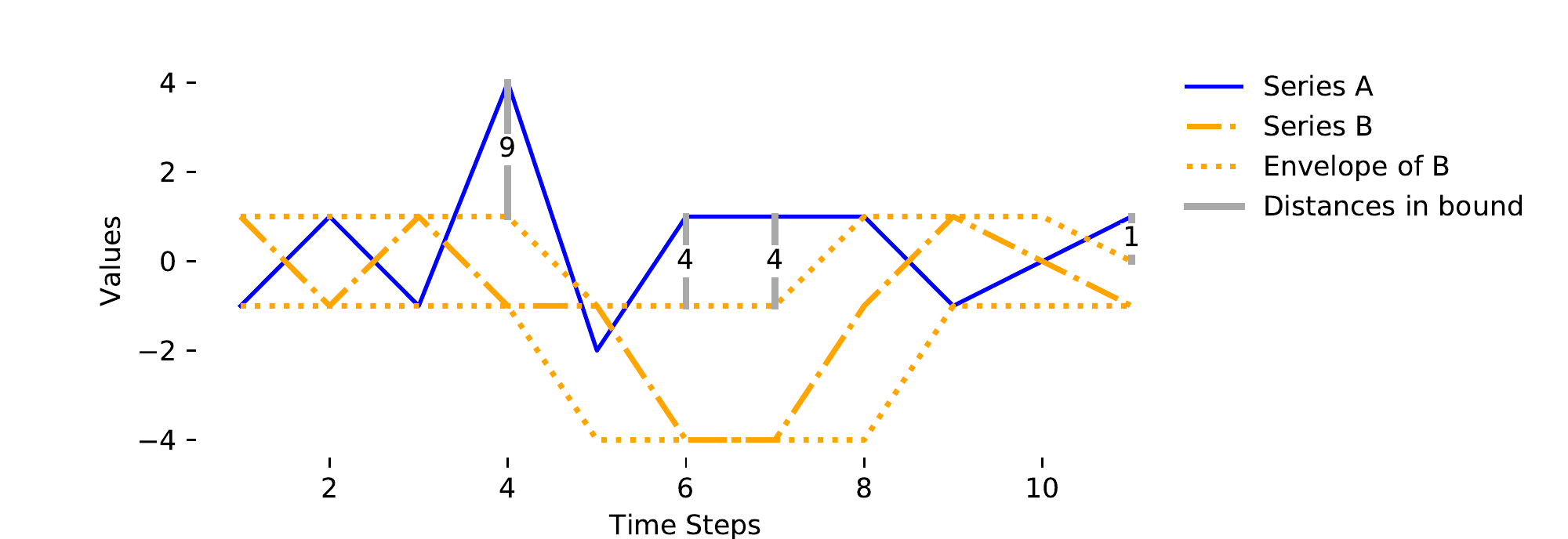}
\end{center}
\vspace*{-15pt}
\caption{Illustration of $\lbkeogh$ with $w=1$ and $\delta(A_i,B_j)=(A_i-B_j)^2$. The gray lines represent the distances $\lbkeogh$ captures.}
\label{fig:lbkeogh}
\end{figure}

$\lbkeogh$  \cite{keogh2005exact} employs a pair of derived series called the \emph{envelopes}. Given window $w$, the upper, $\upperenv^S$, and lower, $\lowerenv^S$, envelopes of time series $S$
 are series representing the maximum and minimum values of $S$ within the window for each point in $S$.

\begin{gather}
    \upperenv_i^S =\max_{\max(1,i{-}w)\leq j \leq \min(\tslen, i{+}w)}(S_j) \label{eqn:upperenvelope} \nonumber\\
    \lowerenv_i^S =\min_{\max(1,i{-}w)\leq j \leq \min(\tslen, i{+}w)}(S_j) \label{eqn:lowerenvelope}\nonumber
\end{gather} 

\begin{equation*}
\LBKeogh_w(A,B) = \sum_{i=1}^{\tslen} 
    \begin{cases}
        \delta(A_i,\upperenv_i^B) & \text{if } A_i > \upperenv_i^B \\
        \delta(A_i,\lowerenv_i^B) & \text{if } A_i < \lowerenv_i^B \\
        0 & \text{otherwise}
    \end{cases} 
\end{equation*}
This bound is illustrated in Figure~\ref{fig:lbkeogh}.

\begin{figure}
\begin{center}
\includegraphics[width=1.0\columnwidth]{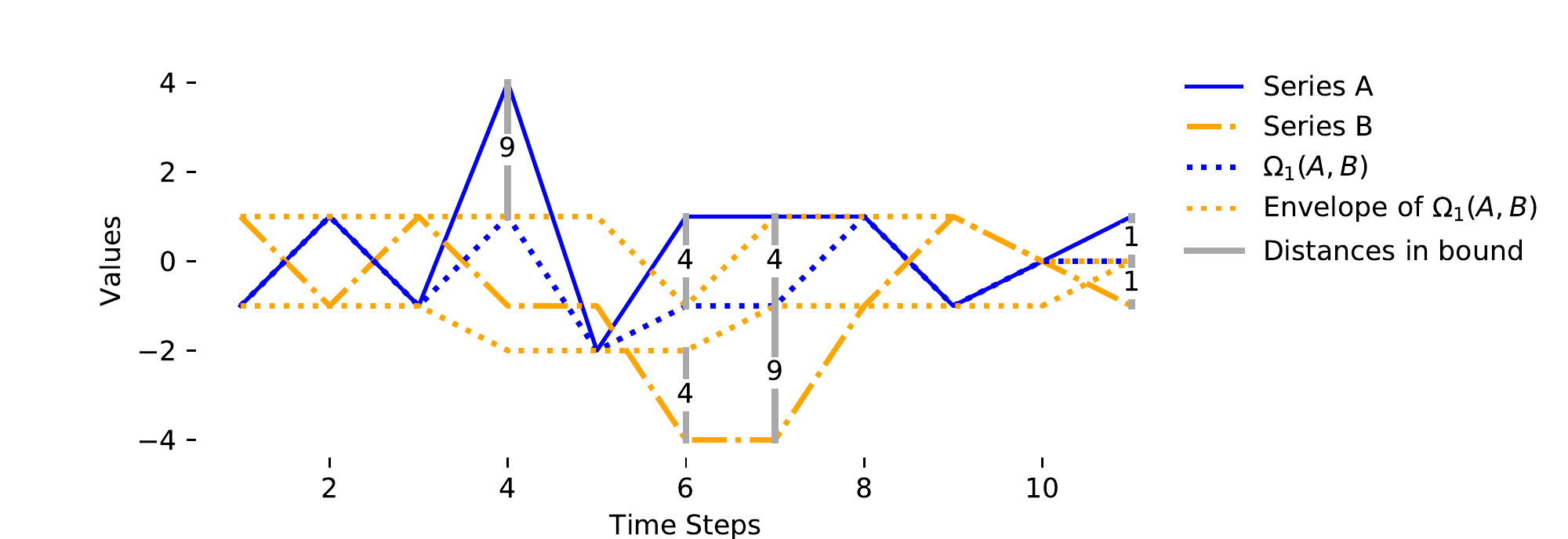}
\end{center}
\vspace*{-10pt}
\caption{Illustration of $\lbimproved$ with $w=1$ and $\delta(A_i,B_j)=(A_i-B_j)^2$. The gray lines represent the distances $\lbimproved$ captures.}
\label{fig:lbimproved}
\end{figure}

A tighter bound is provided by $\lbimproved$ \cite{lemire2009faster}.  This bound augments $\lbkeogh$ by capturing not only distances from series $A$ to the envelope of $B$, but also some distances from $B$ that are not captured by $\lbkeogh$. It~uses the envelopes of a further derived series called the \emph{projection} of $A$.
The projection $\projection_w(A,B)$ of $A$ onto $B$ is a sequence such that for all $i$, $1\leq i\leq \tslen$,
\begin{equation*}
\projection_w(A,B)_i = 
    \begin{cases}
        \upperenv_i^B & \text{if } A_i > \upperenv_i^B \\
        \lowerenv_i^B & \text{if } A_i < \lowerenv_i^B \\
        A_i & \text{otherwise}
    \end{cases} 
\end{equation*}

\begin{equation*}
\LBImproved =
 \LBKeogh_w(A,B)  
+ \sum_{i=1}^{\tslen}
    \begin{cases}
        \delta(B_i,\upperenv_i^{\projection_w(A,B)}) & \text{if } B_i > \upperenv_i^{\projection_w(A,B)} \\
        \delta(B_i,\lowerenv_i^{\projection_w(A,B)}) & \text{if } B_i < \lowerenv_i^{\projection_w(A,B)} \\
        0 & \text{otherwise}
    \end{cases}.
\end{equation*}
This bound is illustrated in Figure~\ref{fig:lbimproved}.

A recently derived bound, $\lbenhanced$ \cite{TanEtAl19}, melds two strategies for establishing a lower bound. It uses the strategy of summing distances to an envelope employed by both $\lbkeogh$ and $\lbimproved$.  It~adds to this  a constant time operation applied to the start and end of the series, where alignments are more constrained. It~employs the concept of a \emph{band}.  This is a continuous path through the cost matrix from a cell at the top or right of the accessible region to a cell at the bottom or left. The sum of the minimum values in any collection of non-overlapping bands forms a DTW lower bound. Two key forms of band are the \emph{left bands}
\begin{align}
\leftset^w_i{=}&\{(\max(1,i{-}w),i),(\max(1,i{-}w)+1,i),\ldots,(i,i),\nonumber\\
&\hspace*{5pt}(i,i-1),\ldots,(i,\max(1,i{-}w)\} \nonumber
\end{align}
and the \emph{right bands}
\begin{align}
\rightset^w_i=&\{(\min(\tslen,i{+}w),i),(\min(\tslen,i{+}w)+1,i),\ldots,(i,i),\nonumber\\
&\hspace*{5pt}(i,i-1),\ldots,(i,\min(\tslen,i{+}w)\} \nonumber.
\end{align}
The use of each of these types of bands to calculate a lower bound in isolation is illustrated in Figures \ref{fig:leftmatrix} and  \ref{fig:rightmatrix}.%
\begin{figure}
\def\A{{-1,1,-1, 4, -2,1,1,1, -1,0,1}}
\def\B{{1,-1,1,-1,-1,-4,-4,-1,1,0,-1}}
\def\N{11}
\def\M{11}
\def\W{1}
\pgfmathsetcounter{N}{\N-1}
\pgfmathsetcounter{M}{\M-1}
    \centering
\footnotesize
		\begin{tikzpicture}[scale=.35]
		\drawmatrix
		\foreach \k in {0,...,\value{N}} {
			\dotos{\k}{}
		}
		\end{tikzpicture}
    \caption{The cost matrix for calculating a lower bound using \emph{left} bands with $w{=}\W$ and $\delta(A_i,B_j)=(A_i-B_j)^2$.  
    Alternating colors distinguish successive bands. The sum over all bands of the minimum distances for that band provides a lower bound (39). 
    }
    \label{fig:leftmatrix}
\end{figure}
\begin{figure}
\def\A{{-1,1,-1, 4, -2,1,1,1, -1,0,1}}
\def\B{{1,-1,1,-1,-1,-4,-4,-1,1,0,-1}}
\def\N{11}
\def\M{11}
\def\W{1}
\pgfmathsetcounter{N}{\N-1}
\pgfmathsetcounter{M}{\M-1}
    \centering
\footnotesize
		\begin{tikzpicture}[scale=.35]
		\drawmatrix
		\foreach \k in {0,...,\value{N}} {
			\dolos{\k}{}
		}
		\end{tikzpicture}
    \caption{The cost matrix for calculating a lower bound using \emph{right} bands with $w{=}\W$ and $\delta(A_i,B_j)=(A_i-B_j)^2$. 
     Alternating colors distinguish successive bands. The sum over all bands of the minimum distances for that band provides a lower bound (36). 
    }
    \label{fig:rightmatrix}
\end{figure}

$\lbenhanced$ uses just the $k$ leftmost left bands and rightmost right bands, as these are the smallest bands and hence will usually contribute most to the lower bound.  The distance between these bands is bridged using $\lbkeogh$, as illustrated in Figure \ref{fig:tan-matrix}.
\begin{figure}
\def\A{{-1,1,-1, 4, -2,1,1,1, -1,0,1}}
\def\B{{1,-1,1,-1,-1,-4,-4,-1,1,0,-1}}
\def\N{11}
\def\M{11}
\def\W{1}
\pgfmathsetcounter{N}{\N-1}
\pgfmathsetcounter{M}{\M-1}
    \centering
		\pgfmathtruncatemacro{\lastt}{1}
		\pgfmathtruncatemacro{\firstl}{\N-2}
		\pgfmathtruncatemacro{\firstm}{\lastt+1}
		\pgfmathtruncatemacro{\lastm}{\firstl-1}
\footnotesize
		\begin{tikzpicture}[scale=.35]
		\drawmatrix
		\foreach \k in {0,...,\lastt} {
			\dotos{\k}{}
		}
		\foreach \k in {2,...,2} {
			\doAosGrey{\k}
		}
		\foreach \k in {3,...,3} {
			\doAos{\k}{}
		}
		\foreach \k in {4,...,4} {
			\doAosGrey{\k}
		}
		\foreach \k in {5,...,6} {
			\doAos{\k}{}
		}
		\foreach \k in {7,...,8} {
			\doAosGrey{\k}
		}
		\foreach \k in {\firstl,...,\value{N}} {
			\dolos{\k}{}
		}
		\end{tikzpicture}    \caption{The cost matrix for calculating $\lbenhanced_1^2(A,B)$ with $k{=}2$, $w{=}1$ and $\delta(A_i,B_j)=(A_i-B_j)^2$. 
     Alternating colors distinguish successive bands. The vertical bands represent the regions bridged by $\lbkeogh$.  For the vertical bands in gray $\lowerenv^{B}_i\leq A_i\leq\upperenv^{B}_i$ and hence the values cannot contribute to the bound.
The minimum value in each other band is set in bold.
The sum of all these minimum distances provides a lower bound (25). 
    }
    \label{fig:tan-matrix}
\end{figure}

\begin{align*}
\lbenhanced_w^k(A,B) = &\sum_{i=1}^k\left[\min(\leftset^w_i)+\min(\rightset^w_{\tslen-i+1})\right]\\
&+\sum_{i=k+1}^{\tslen-k} 
    \begin{cases}
        \delta(A_i,\upperenv_i^B) & \text{if } A_i > \upperenv_i^B \\
        \delta(A_i,\lowerenv_i^B) & \text{if } A_i < \lowerenv_i^B \\
        0 & \text{otherwise}
    \end{cases} 
\end{align*}
\begin{figure}
\begin{center}
\includegraphics[width=1.0\columnwidth]{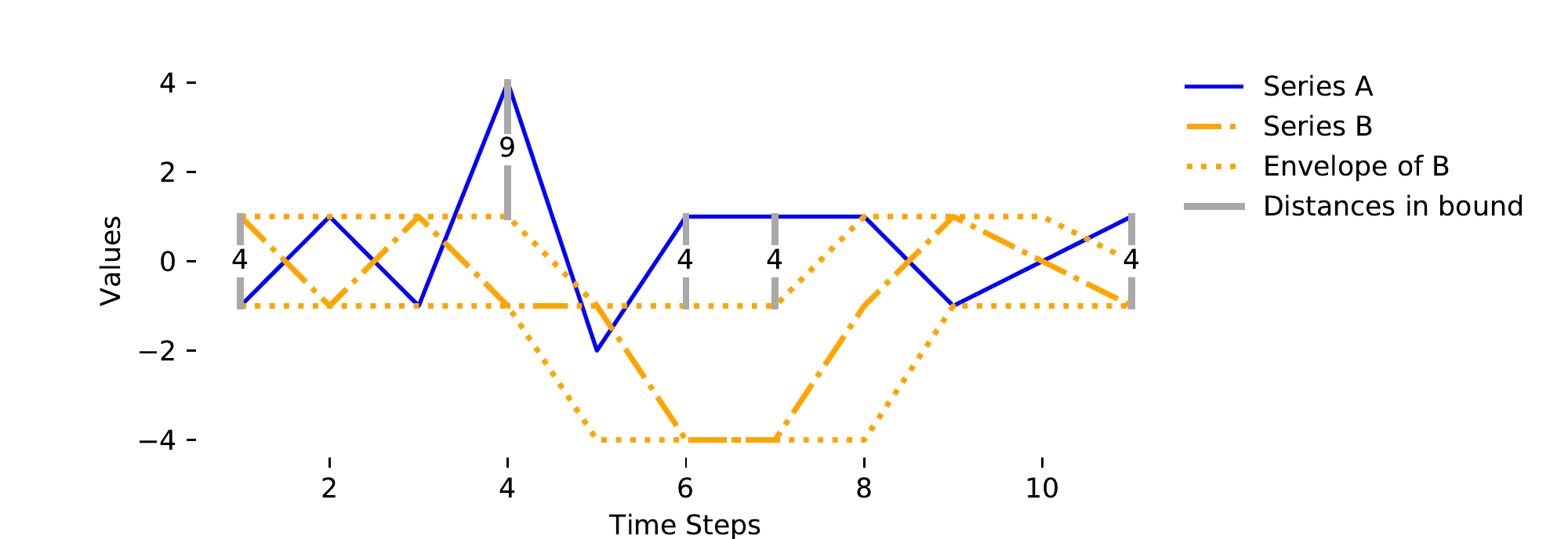}
\end{center}
\vspace*{-10pt}
\caption{Illustration of $\lbenhanced$ with $w=1$, $k=1$ and $\delta(A_i,B_j)=(A_i-B_j)^2$. The gray lines represent the distances $\lbenhanced$ captures. }
\label{fig:lbenhanced}
\end{figure}

\section{The $\lbtight$ lower bound}\label{sec:petitjean}

For some elements of $A$ and $B$, $\lbimproved$ can identify a boundary on the region that $\lbkeogh$ can reach (the envelope of the projection) and add distances from elements in $B$ to this boundary.  Consider an ideal case where the warping path connects $A_i$ and $B_j$ and $\lbimproved$ can incorporate $\delta(A_i, \upperenv_i^B)+\delta(\upperenv_i^B, B_j)$, such as alignment $(A_6,B_7)$ in Figure~\ref{fig:dtw} and the $\lbimproved$ illustration in Figure~\ref{fig:lbimproved}.   With $\delta(A_i,B_j)=(A_i-B_j)^2$, the warping path for these points costs $\delta(A_i,B_j)=25>\delta(A_i, \upperenv_i^B)+\delta(\upperenv_i^B, B_j)=4+9$, which is the allowance by $\lbimproved$.  We present $\lbtight$, a tighter variant of $\lbimproved$ that uses a stronger strategy for adding to $\lbkeogh$ an allowance for points in $B$ that cannot be reached by the distances included in $\lbkeogh$.  It also exploits the constraints on alignments at the start and end of the series, employing a strategy similar to $\lbenhanced$. $\lbtight$ uses the following observations.

\begin{observation}\label{obs:2}
For any alignment $(A_i,B_j)$, if $B_j> \upperenv_j^{\projection}$ then either $A_i<\lowerenv_i^B\leq B_j\leq \upperenv_i^B$ or $\lowerenv_i^B\leq A_i\leq B_j\leq \upperenv_i^B$.
\end{observation}
\begin{proof}
For $A_i$ to be aligned with $B_j$ it is necessary that $i-w\leq j \leq i+w$.  Hence $\lowerenv_i^B\leq B_j\leq \upperenv_i^B$(note, $B$ indexed by $j$, the index with which $A_i$ is aligned, but bounds indexed by $i$).  It cannot be that $A_i>B_j$ because that would require that $\projection_i<A_i$, which can only happen if $A_i> \upperenv_i^B$, in which case $\projection_i= \upperenv_i^B$ which entails that $\upperenv_j^{\projection}\geq\mathbb{U}_i^B$ which contradicts that $B_j> \upperenv_j^{\projection}$.
\end{proof}

\begin{observation}\label{obs:3}
By the same reasoning as Observation~\ref{obs:2}, for any alignment $(A_i,B_j)$, if $B_j< \lowerenv_j^{\projection}$ then either $A_i>\upperenv_i^B\geq B_j$ or $\upperenv_i^B\geq A_i\geq B_j\geq \lowerenv_i^B$.
\end{observation}

\sloppy$\lbtight$ uses these observations to derive tighter bounds than $\lbimproved$.  Returning to the ideal case, where the warping path connects $A_i$ and $B_j$ and $\lbimproved$ can incorporate $\delta(A_i, \upperenv_i^B)+\delta(\upperenv_i^B, B_j)$, $\lbtight$ can instead incorporate the greater amount of $\delta(A_i, \upperenv_i^B)+\delta(\lowerenv_j^A, B_j)-\delta(\upperenv_i^B,\lowerenv_j^A)$. For alignment $(A_6,B_7)$ in Figure~\ref{fig:dtw} this is 21. $B_7=-4$ must align with one of $A_6,A_7$ or $A_8$, all of which have value $1$. Thus, the value of its alignment must equal $(1- -4)^2=25$.  However, the $\lbkeogh$ bound may already have incorporated an allowance for this alignment of up to the distance between the furthest point in $A$ to which $B_7$ might be aligned, $\upperenv_7^A=1$, and the closest point in the envelope of $B$ that is within the window of any point to which $B_7$ could be aligned, $\lowerenv_7^{\projection_7(A,B)}=-1$. This allowance for the largest possible value from $\lbkeogh$ for an alignment with $B_7$, $\delta(\upperenv_7^A,\lowerenv_7^{\projection_7(A,B)})=4$ is subtracted from the distance that is added to the bound, resulting in $25-4=21$.  

$\lbtight$ further exploits the tight constraints on the first and last few alignments in any warping path.  This additional mechanism, called the \emph{left and right paths}, is inspired by $\lbenhanced$. It incorporates the minimum of each of the possible first and last three sequences of alignments in the path, an even tighter mechanism than that of $\lbenhanced$. The length of these initial and final paths are limited to three because there are just seven such options involving just six distances, as illustrated in Figure~\ref{fig:start-end}.  If these paths are increased to length 4, the number of alternatives leaps to 21.  The most efficient manner to compute start and end paths of any greater length than three would almost certainly be to use the same dynamic programming process used to find the full path, an operation of the same complexity as directly finding the path and hence of little practical utility for calculating a lower bound.
\begin{figure}
\def\A{{-1, 4, -2,1,1,1, -1}}
\def\B{{1,-1,-1,-4,-4,-1,1}}
\def\N{7}
\def\M{7}
\def\W{3}
\pgfmathsetcounter{N}{\N-1}
\pgfmathsetcounter{M}{\M-1}
    \centering
		\pgfmathtruncatemacro{\lastt}{1}
		\pgfmathtruncatemacro{\firstl}{\N-2}
		\pgfmathtruncatemacro{\firstm}{\lastt+1}
		\pgfmathtruncatemacro{\lastm}{\firstl-1}
\footnotesize
		\begin{tikzpicture}[scale=.5]
		\drawmatrix
		\draw[very thick,color=blue,->] (0.5,-0.2) -- (1.5,-0.2) --(2.5,-0.5);
		\draw[very thick,color=red,->] (0.5,-0.3) -- (1.5,-0.3) --(2.5,-1.4);
		\draw[very thick,color=blue,->] (0.5,-0.4) -- (1.5,-1.4) --(2.4,-1.5);
		\draw[very thick,color=purple,->] (0.5,-0.5) -- (2.5,-2.5);
		\draw[very thick,color=blue,->] (0.2,-0.5) -- (0.2,-1.5) --(0.5,-2.5);
		\draw[very thick,color=red,->] (0.3,-0.5) -- (0.3,-1.5) --(1.4,-2.5);
		\draw[very thick,color=blue,->] (0.4,-0.5) -- (1.4,-1.5) --(1.5,-2.4);

		\draw[very thick,color=blue,->] (6.5,-6.8) -- (5.5,-6.8) --(4.5,-6.5);
		\draw[very thick,color=red,->] (6.5,-6.7) -- (5.5,-6.7) --(4.5,-5.6);
		\draw[very thick,color=blue,->] (6.5,-6.6) -- (5.5,-5.6) --(4.6,-5.5);
		\draw[very thick,color=purple,->] (6.5,-6.5) -- (4.5,-4.5);
		\draw[very thick,color=blue,->] (6.8,-6.5) -- (6.8,-5.5) --(6.5,-4.5);
		\draw[very thick,color=red,->] (6.7,-6.5) -- (6.7,-5.5) --(5.6,-4.5);
		\draw[very thick,color=blue,->] (6.6,-6.5) -- (5.6,-5.5) --(5.5,-4.6);
		\end{tikzpicture}    \caption{The limited number of potential minimal cost start and end paths of length three.}
    \label{fig:start-end}
\end{figure}
\begin{align*}
\minlrpaths(A,B)=\hspace*{-80pt}\\
~~~&\delta(A_1,B_1)+\delta(A_\tslen,B_\tslen)\\
&+\min[\delta(A_1,B_2){+}\delta(A_1,B_3), \delta(A_1,B_2){+}\delta(A_2,B_3), \\
&\hspace{31pt}\delta(A_2,B_2){+}\delta(A_2,B_3), \delta(A_2,B_2){+}\delta(A_3,B_3), \\
&\hspace{31pt}\delta(A_2,B_2){+}\delta(A_3,B_2), \delta(A_2,B_1){+}\delta(A_3,B_2), \\
&\hspace{31pt}\delta(A_2,B_1){+}\delta(A_3,B_1)]\\
&+\min[\delta(A_\tslen,B_{\tslen{-}1}){+}\delta(A_\tslen,B_{\tslen{-}2}), \delta(A_\tslen,B_{\tslen{-}1}){+}\delta(A_{\tslen{-}1},B_{\tslen{-}2}),  \\
&\hspace{31pt}\delta(A_{\tslen{-}1},B_{\tslen{-}1}){+}\delta(A_{\tslen{-}1},B_{\tslen{-}2}),\delta(A_{\tslen{-}1},B_{\tslen{-}1}){+}\delta(A_{\tslen{-}2},B_{\tslen{-}2}),   \\
&\hspace{31pt}\delta(A_{\tslen{-}1},B_{\tslen{-}1}){+}\delta(A_{\tslen{-}2},B_{\tslen{-}1}),\delta(A_{\tslen{-}1},B_\tslen){+}\delta(A_{\tslen{-}2},B_{\tslen{-}1}),\\
&\hspace{31pt}\delta(A_{\tslen{-}1},B_\tslen){+}\delta(A_{\tslen{-}2},B_\tslen)]
\end{align*}

$\lbtight$ assumes that $\forall_{x,y:A_i\leq x\leq y\leq B_j\vee A_i\geq x\geq y\geq B_j}\delta(A_i,B_j) \geq \delta(A_i, y)+\delta(B_j,x)-\delta(x,y)$. This is true of both $\delta=|A_i-B_j|$ and $\delta(A_i,B_j)=(A_i-B_j)^2$.

\begin{theorem}\label{thm:tight}
If $\forall_{x,y:A_i\leq x\leq y\leq B_j\vee A_i\geq x\geq y\geq B_j}\delta(A_i,B_j) \geq \delta(A_i, y)+\delta(B_j,x)-\delta(x,y)$,
\begin{align*}
\lbtight_w(A,B)=&\minlrpaths(A,B)\\
&+\sum_{i=4}^{\tslen-3} 
    \begin{cases}
        \delta(A_i,\upperenv_i^B) & \text{if } A_i > \upperenv_i^B \\
        \delta(A_i,\lowerenv_i^B) & \text{if } A_i < \lowerenv_i^B \\
        0 & \text{otherwise}
    \end{cases}\\
&  + \sum_{j=4}^{\tslen-3}
    \begin{cases}
        \delta(B_j,\upperenv_j^{A})-\delta( \upperenv_j^{\projection}, \upperenv_j^A) & \text{if } B_j > \upperenv_j^{\projection} > \upperenv_j^A\\
        \delta(B_j,\lowerenv_j^{A})-\delta( \lowerenv_j^{\projection}, \lowerenv_j^A) & \text{if } B_j < \lowerenv_j^{\projection}  < \lowerenv_j^A\\
        \delta(B_j,\upperenv_j^{\projection}) & \text{if } B_j > \upperenv_j^{\projection} \leq \upperenv_j^A\\
        \delta(B_j,\lowerenv_j^{\projection}) & \text{if } B_j < \lowerenv_j^{\projection} \geq \lowerenv_j^A \\
        0 & \text{otherwise}
    \end{cases} 
\end{align*}
 is a lower bound on $\DTW_w(A,B)$, where $ \upperenv_i^{\projection}$ denotes $\upperenv_i^{\projection_w(A,B)}$ and $\lowerenv_i^{\projection}$ denotes $\lowerenv_i^{\projection_w(A,B)}$.
\end{theorem}
\allowdisplaybreaks
\begin{proof}
\begin{align}
\DTW_w(A,B)\hspace*{-45pt}\nonumber\\[6pt]
&=\sum_{(i,j)\in\path}\delta(A_i,B_j)\label{eq:dtwdef}\\[6pt]
&\geq\sum_{(i,j)\in\path}
\begin{cases}
	\delta(A_i,B_j) &  \text{if } i{\leq} 3\wedge j{\leq}3 \hfill \myeqno{lr1}\\
	\delta(A_i,B_j) &  \text{if } i{\geq} \tslen- 2\wedge j{\geq} \tslen- 2 \hfill \myeqno{lr2}\\
        \delta(A_i, \lowerenv_i^B) + \delta(B_j,\upperenv_j^{A})-\delta( \upperenv_j^{\projection}, \upperenv_j^A) & \text{if }  A_i < \lowerenv_i^B\wedge B_j > \upperenv_j^{\projection} > \upperenv_j^A \quad\hfill \myeqno{sepup1}\\
        \delta(A_i, \lowerenv_i^B) + \delta(B_j,\upperenv_j^{\projection}) & \text{if } A_i < \lowerenv_i^B\wedge B_j > \upperenv_j^{\projection}\leq \upperenv_j^A \quad\hfill \myeqno{sepup2}\\
        \delta(B_j,\upperenv_j^{A})-\delta( \upperenv_j^{\projection}, \upperenv_j^A) & \text{if }  A_i \geq \lowerenv_i^B\wedge B_j > \upperenv_j^{\projection} > \upperenv_j^A \quad\hfill \myeqno{sepup3}\\
        \delta(B_j,\upperenv_j^{\projection}) & \text{if }A_i \geq \lowerenv_i^B\wedge  B_j > \upperenv_j^{\projection}\leq \upperenv_j^A \quad\hfill \myeqno{sepup4}\\
        \delta(A_i, \lowerenv_i^B) & \text{if }  A_i < \lowerenv_i^B\wedge  B_j \leq \upperenv_j^{\projection}\hfill \myeqno{sepup5}\\
        \delta(A_i, \upperenv_i^B) +  \delta(B_j,\lowerenv_j^{A})-\delta( \lowerenv_j^{\projection}, \lowerenv_j^A) & \text{if }A_i >\upperenv_i^B\wedge   B_j < \lowerenv_j^{\projection}  < \lowerenv_j^A \hfill \myeqno{sepdown1}\\
        \delta(A_i, \upperenv_i^B) +  \delta(B_j, \lowerenv_j^{\projection}) & \text{if }A_i >\upperenv_i^B\wedge   B_j < \lowerenv_j^{\projection}  \geq \lowerenv_j^A \hfill \myeqno{sepdown2}\\
       \delta(B_j,\lowerenv_j^{A})-\delta( \lowerenv_j^{\projection}, \lowerenv_j^A) & \text{if }A_i \leq\upperenv_i^B\wedge   B_j < \lowerenv_j^{\projection}  < \lowerenv_j^A \hfill \myeqno{sepdown3}\\
       \delta(B_j,\lowerenv_j^{\projection}) & \text{if }A_i \leq\upperenv_i^B\wedge   B_j < \lowerenv_j^{\projection}  \geq \lowerenv_j^A \hfill \myeqno{sepdown4}\\
        \delta(A_i, \upperenv_i^B) & \text{if }A_i >\upperenv_i^B \wedge   B_j \geq \lowerenv_j^{\projection}\hfill \myeqno{sepdown5}\\
        0& \text{otherwise} \hfill \myeqno{default}
\end{cases}\nonumber\\[6pt]
&\geq \minlrpaths(A,B) \quad\quad\quad\quad\quad\quad\quad\quad\quad\quad\quad\quad\hfill \myeqno{lrpaths}\nonumber\\
&\hspace*{10pt}+\sum_{i=4}^{\tslen-3}
	\begin{cases}
		\delta(A_i, \lowerenv_i^B)&\text{if } A_i <  \lowerenv_i^B \quad\quad\quad\quad\quad\quad\quad\hfill \myeqno{keogh2.1}\\
		\delta(A_i, \upperenv_i^B)&\text{if } A_i >  \upperenv_i^B \hfill \myeqno{keogh2.2}\\
		0 & \text{otherwise} \hfill \myeqno{keogh2.3}
	\end{cases}\nonumber\\[4pt]
&\hspace*{10pt} + \sum_{j=4}^{\tslen-3}
    \begin{cases}
        \delta(B_j,\upperenv_j^{A})-\delta( \upperenv_j^{\projection}, \upperenv_j^A) & \text{if } B_j > \upperenv_j^{\projection}  > \upperenv_j^A \quad\hfill \myeqno{new2.1}\\
        \delta(B_j,\lowerenv_j^{A})-\delta( \lowerenv_j^{\projection}, \lowerenv_j^A) & \text{if } B_j < \lowerenv_j^{\projection}  < \lowerenv_j^A \hfill \myeqno{new2.2}\\
        \delta(B_j,\upperenv_j^{\projection}) & \text{if } B_j > \upperenv_j^{\projection}  \hfill \myeqno{new2.3}\\
        \delta(B_j,\lowerenv_j^{\projection}) & \text{if } B_j < \lowerenv_j^{\projection}  \hfill \myeqno{new2.4}\\
        0 & \text{otherwise.} \hfill \myeqno{new2.5}
    \end{cases} \nonumber
\end{align}
\qed
\end{proof}
{\bf Notes}: 

(\ref{eq:dtwdef}) repeats the definition of $\DTW$ as a sum over all alignments in $\path$. 

(\ref{lrpaths}) to (\ref{new2.5}) are the clauses of Theorem~\ref{thm:tight}. (\ref{lrpaths}) adds $\minlrpaths(A,B)$, a quantity no greater than the sum of the alignments between the first three elements of each series and between the last three elements of each. (\ref{keogh2.1}) to (\ref{keogh2.3}) add allowances for each $A_i:4\leq i\leq\tslen{-}3$. (\ref{new2.1}) to (\ref{new2.5}) add allowances for each $B_j:4\leq j\leq\tslen{-}3$.

Clauses (\ref{lr1}) to (\ref{default}) repeat the sum over all alignments $(A_i,B_j)\in\path$, separating them by the various possible combinations of a condition in (\ref{keogh2.1}) to(\ref{keogh2.3}) with a condition in (\ref{new2.1}) to (\ref{new2.5}). Each condition implicitly includes \emph{and none of the above}. The key constraints that arise due to this ordering are made explicit. However, it is important to keep in mind that each of (\ref{sepup1}) to (\ref{default}) includes the implicit constraint $4{\leq}i{\leq}\tslen{-}3\vee  4{\leq}j{\leq}\tslen{-}3$.  
Clauses (\ref{lr1}) to (\ref{default}) do not include cases with both $A_i >\upperenv_i^B$ and $B_j > \upperenv_j^{\projection}$ or both $A_i <\lowerenv_i^B$ and $B_j < \lowerenv_j^{\projection}$ because these are mutually inconsistent, as per Observations \ref{obs:2} and \ref{obs:3}.

For each clause in (\ref{lr1}) to (\ref{default}), the clause in (\ref{keogh2.1}) to (\ref{keogh2.3}) that will apply to the specific $A_i$ and the clause in (\ref{new2.1}) to (\ref{new2.5}) that will apply to the specific $B_j$ are uniquely determined and the addition to the sum over alignments is the sum of the values that will be added by clauses (\ref{lrpaths}) to (\ref{new2.5}) for this $A_i$ and $B_i$. As every $A_i$ and $B_i$ must appear in at least one alignment in $\path$, and as the sum of the provisions for each $A_i$ and $B_j$ are no greater than the corresponding $\delta(A_i,B_j)$, $\lbtight_w(A,B)$ must be a lower bound on $\DTW(A,B)$.

The following notes discuss each case in turn.
\begin{itemize}
\item[(\ref{lr1},\ref{lr2}):] these capture the alignments between the first three and between the last three elements of $A$ and $B$. The $\minlrpaths(A,B)$ on line $(\ref{lrpaths})$ contributes an amount not greater than the sum of these.  The remaining alignments can include elements in $\{A_1,\ldots A_3, A_{\tslen-2}, \ldots A_\tslen,\}$, but only aligned with elements outside  $\{B_1,\ldots B_3,  B_{\tslen-2}, \ldots B_\tslen\}$, and vice versa.
\item[(\ref{sepup1}):] this captures alignments $(A_i,B_j)$ where $A_i$  will be covered by case (\ref{keogh2.1})and $B_j$ will be covered by  (\ref{new2.1}).
$A_i < \lowerenv_i^B\vdash\projection_i=\lowerenv_i^B\leq\upperenv_j^{\projection}$. As $B_j > \upperenv_j^{\projection} > \upperenv_j^A$, 
\begin{align}
\delta(A_i,B_j) &\geq \delta(A_i, \upperenv_j^{\projection})+\delta(B_j,\upperenv^A_j)-\delta( \upperenv_j^{\projection}, \upperenv_j^A)\label{eq:ineq}\\
&\geq \delta(A_i, \lowerenv_i^B) + \delta(B_j,\upperenv_j^{A})-\delta( \upperenv_j^{\projection}, \upperenv_j^A).\nonumber
\end{align}
\item[(\ref{sepup2}):] this captures alignments   $(A_i,B_j)$ where $A_i$  will  be covered by case (\ref{keogh2.1})  and $B_j$ will be covered by  (\ref{new2.3}).
$A_i < \lowerenv_i^B\vdash\projection_i=\lowerenv_i^B\leq\upperenv_j^{\projection}$. As $B_j > \upperenv_j^{\projection}$, 
\begin{align*}
\delta(A_i,B_j) &> \delta(A_i,\upperenv_j^{\projection})+\delta(B_j,\upperenv_j^{\projection})\\
&\geq \delta(A_i, \lowerenv_i^B) + \delta(B_j, \upperenv_j^{\projection}).
\end{align*}
\item[(\ref{sepup3}):] this captures alignments   $(A_i,B_j)$ where $A_i$  will  be covered by case (\ref{keogh2.3}) and $B_j$ will be covered by  (\ref{new2.1}).
\begin{equation*}B_j>\upperenv^{A}_j\vdash\delta(A_i,B_j) \geq\delta(B_j,\upperenv^A_j)\geq \delta(B_j,\upperenv_j^{A})-\delta( \upperenv_j^{\projection}, \upperenv_j^A).
\end{equation*}
\item[(\ref{sepup4}):] this captures alignments   $(A_i,B_j)$ where $A_i$  will  be covered by case (\ref{keogh2.1})  and $B_j$ will be covered by  (\ref{new2.3}).
$A_i \geq \lowerenv_i^B\wedge B_j >\upperenv_j^{\projection}\vdash\projection_i=B_i\leq\upperenv_j^{\projection}$. Hence,
\begin{equation*}
\delta(A_i,B_j) > \delta(B_j,\upperenv_j^{\projection})
\end{equation*}
\item[(\ref{sepup5}):] this captures alignments   $(A_i,B_j)$ where $A_i$  will  be covered by case (\ref{keogh2.3})  and $B_j$ will be covered by  (\ref{new2.5}).
\begin{equation*}
A_i \leq \lowerenv_i^B\vdash\delta(A_i,B_j) > \delta(A_i,\lowerenv^B_i).
\end{equation*}
\item[(\ref{sepdown1}-\ref{sepdown5}):] these are equivalent to (\ref{sepup1}-\ref{sepup5}), addressing clauses (\ref{new2.2}) and (\ref{new2.4}) in place of (\ref{new2.1}) and (\ref{new2.3}) and exchanging upper envelopes for lower and vice versa.
\item[(\ref{default}):] this captures alignments $(A_i,B_j)$ where $A_i$  will  be covered by case (\ref{keogh2.3})  and $B_j$ will be covered by  (\ref{new2.5}), both of which add zero to the lower bound.
\end{itemize}
\begin{figure}
\begin{center}
\includegraphics[width=1.0\columnwidth]{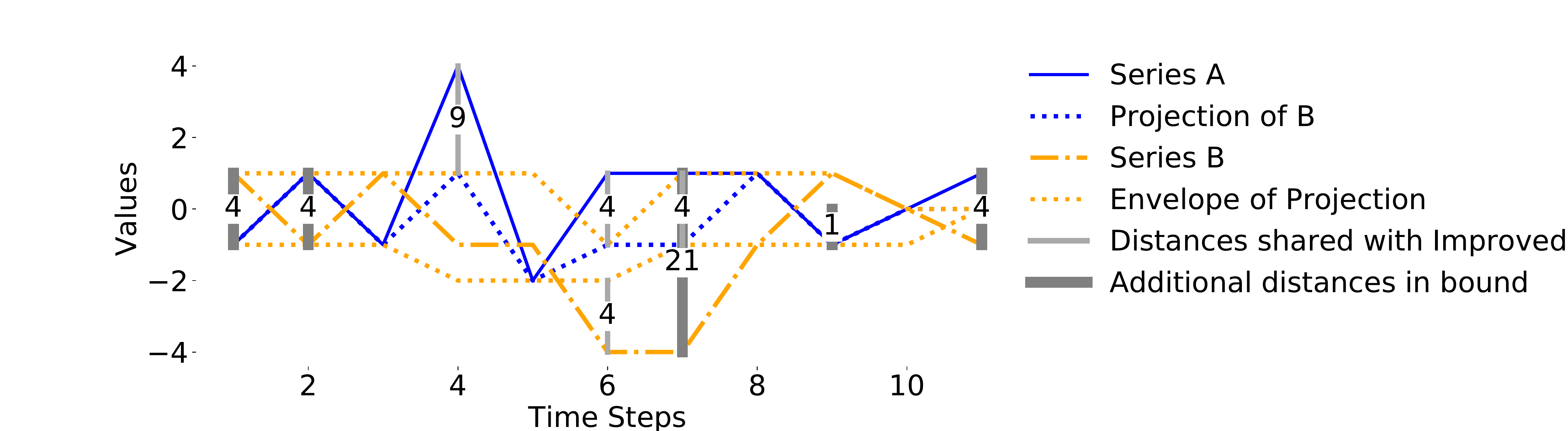}
\end{center}
\vspace*{-15pt}
\caption{Illustration of $\lbtight_1$ with $\delta(A_i,B_j)=(A_i-B_j)^2$. The dark  gray lines represent the points where $\lbtight_1$ captures greater value than \lbimproved. The medium gray lines are values also captured by \lbimproved.}
\label{fig:lbtight}
\end{figure}
This bound is illustrated in Figure~\ref{fig:lbtight}.

When an observation holds irrespective of window size we omit the subscript from $\lbtight$.

A variant of $\lbtight$ that omits the left and right paths, 
\interdisplaylinepenalty=10000 
\begin{align*}
\lbtightnolr_w(A,B)=\hspace*{-20pt}\\
&\sum_{i=1}^{\tslen} 
    \begin{cases}
        \delta(A_i,\upperenv_i^B) & \text{if } A_i > \upperenv_i^B \\
        \delta(A_i,\lowerenv_i^B) & \text{if } A_i < \lowerenv_i^B \\
        0 & \text{otherwise}
    \end{cases}\\
&  + \sum_{i=1}^{\tslen}
    \begin{cases}
        \delta(B_i,\upperenv_i^{A})-\delta( \upperenv_i^{\projection}, \upperenv_i^A) & \text{if } B_i > \upperenv_i^{\projection} > \upperenv_i^A\\
        \delta(B_i,\lowerenv_i^{A})-\delta( \lowerenv_i^{\projection}, \lowerenv_i^A) & \text{if } B_i < \lowerenv_i^{\projection}  < \lowerenv_i^A\\
        \delta(B_i,\upperenv_i^{\projection}) & \text{if } B_i > \upperenv_i^{\projection} \leq \upperenv_i^A\\
        \delta(B_i,\lowerenv_i^{\projection}) & \text{if } B_i < \lowerenv_i^{\projection} \geq \lowerenv_i^A \\
        0 & \text{otherwise}
    \end{cases} 
\end{align*}
is tighter than $\lbimproved$  because if $B_j > \upperenv_j^{\projection}  > \upperenv_j^A$ then $\lbimproved$ will add $\delta(B_j,\upperenv_j^{\projection})$ to the bound whereas $\lbtight$ adds the greater amount $\delta(B_j,\upperenv_j^{A})-\delta( \upperenv_j^{\projection}, \upperenv_j^A)$. Similarly, when $B_j < \lowerenv_j^{\projection}  < \lowerenv_j^A$, $\lbtight$ also adds a greater amount. However, it is possible, but rare in practice, for $\lbtight$ to be less tight than $\lbtightnolr$ and hence possible for it to be less tight than $\lbimproved$.

\begin{algorithm}
\begin{algorithmic}
\Procedure{\lbtight}{series $A$, series $B$, lower envelope of $A$ $\mathit{LA}$, upper envelope  of $A$ $\mathit{UA}$, lower envelope of $B$ $\mathit{LB}$, upper envelope  of $B$ $\mathit{UB}$, window $w$, \#  left-right bands $k$, abandon value $a$}
\State $b\leftarrow\minlrpaths(A,B)$
\For{$i\leftarrow4$ to $\tslen-3$}\Comment{Compute the $\lbkeogh$ bridge.}
		\If{$A_i>\mathit{UB}_i$}
			\State $b\leftarrow b+\delta(A_i,\mathit{UB}_i)$
			\State $P_i\leftarrow\mathit{UB}_i$
		\ElsIf{$A_i<\mathit{LB}_i$}
			\State $b\leftarrow b+\delta(A_i,\mathit{LB}_i)$
			\State $P_i\leftarrow\mathit{LB}_i$
		\Else
			\State $P_i\leftarrow A_i$
		\EndIf
	\IfThen{$b>a$}{\Return{b}}
\EndFor
\State $(LP,UP)\leftarrow\mathrm{compute\_envelopes}(P)$\Comment{Linear time algorithm  \cite{lemire2009faster}.}
\For{$i\leftarrow4$ to $\tslen-3$}\Comment{Allow for $B_i$ that $\lbkeogh$ could not reach.}
		\If{$B_i>\mathit{UP}_i>\mathit{UA}_i$}
			\State $b\leftarrow b+\delta(B_i,\mathit{UA}_i)-\delta(UP_i,UA_i)$
		\ElsIf{$B_i<\mathit{LP}_i<\mathit{LA}_i$}
			\State $b\leftarrow b+\delta(B_i,\mathit{LA}_i)-\delta(LP_i,LA_i)$
		\ElsIf{$B_i>\mathit{UP}_i$}
			\State $b\leftarrow b+\delta(B_i,\mathit{UP}_i)$
		\ElsIf{$B_i<\mathit{LP}_i$}
			\State $b\leftarrow b+\delta(B_i,\mathit{LP}_i)$
		\EndIf
	\IfThen{$b>a$}{\Return{b}}
\EndFor
\State\Return{b}
\EndProcedure
\end{algorithmic}
\caption{Algorithm for computing $\lbtight$}\label{alg:tight}
\end{algorithm}
We present pseudocode for calculating $\lbtight$ in Algorithm \ref{alg:tight}. $\lbtight$ requires calculation of envelopes around both series as well as around the projection.  While the envelopes for the training data can be computed in advance of nearest neighbor search, and the envelope on the query need only be computed once, an envelope on the projection needs to be computed for each   query-training data pair.  This can be computed in $O(\tslen)$ time \cite{lemire2009faster}, and hence $\lbtight$ has $O(\tslen)$ complexity.  Its computation is similar to $\lbimproved$, the major additional overhead being need to compute an envelope on the query and the need to compute two distances for some elements of $B$ rather than one. However, while the complexity is $O(\tslen)$, the constants are large and the additional tightness of the bound relative to $\lbkeogh$ will only compensate for the additional computation in the most demanding of cases.


\section{The $\lbfast$ lower bound}\label{sec:webb}
\begin{figure}
\begin{center}
\includegraphics[width=1.0\columnwidth]{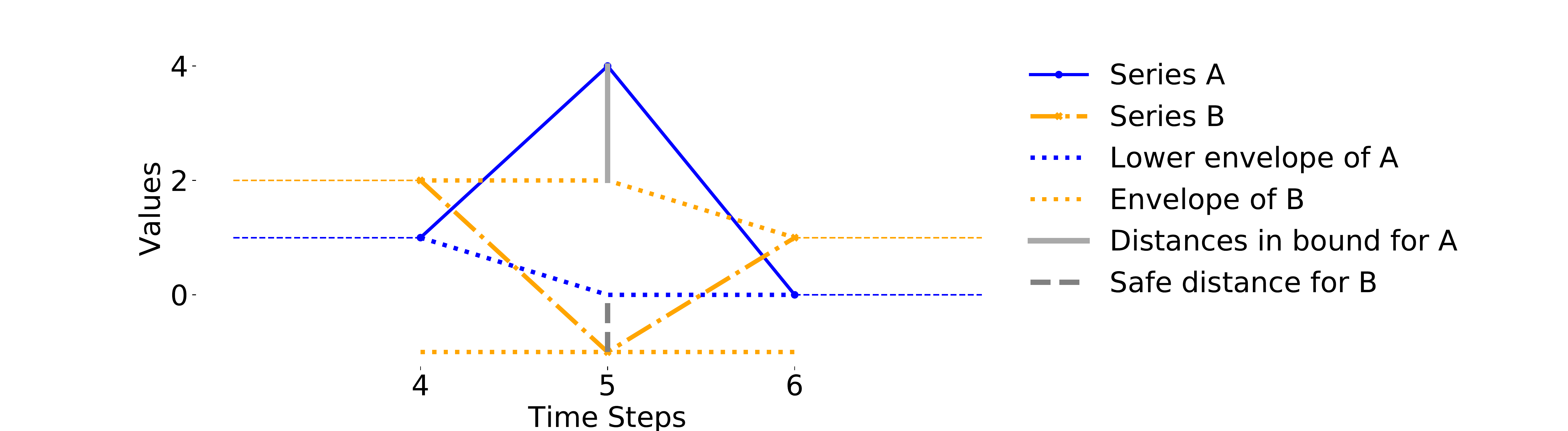}
\end{center}
\vspace*{-15pt}
\caption{Illustration of free below $\lowerenv^A$. $B_5$ is free below $\lowerenv^A$ when $w=1$ because none of the distances included in allowance for $A_i$ within its window extend beyond $\lowerenv^A$. $A_4$ and $A_6$ are both within the envelope of $B$, and so do not contribute to the bound. $A_5$ is above $\upperenv^B$, so contributes $\delta(A_5,\upperenv^B_5)$.  However, as $\upperenv^B_5>\lowerenv^A_5$, it does not extend beyond $\lowerenv^A$.  Note, for computational efficiency, $\lbfast$ uses $\upperenv^{\lowerenv^A}$ rather than $\lowerenv^A$, as if $A_i\geq\upperenv^B_i\geq\upperenv^{\lowerenv^A}_i$ then the allowance for $A_i$ cannot extend beyond $\lowerenv^A_j$ for any $j$ within the window of $i$. Hence, it is safe to include allowance from $B_j$ to $\lowerenv^A_j$ when this is true for all $A_i$ within the window of $j$.}
\label{fig:freeBelow}
\end{figure}

An approximation of $\lbtight$ can be computed without recourse to the projection or its envelopes. This more efficient variant, $\lbfast$, uses the concept that an element $B_j$ is \emph{free above} $\upperenv^A$  if all elements $A_i$ within its window are within the envelope of $B$ or cannot access above $\lowerenv^{\upperenv^A}$.
\begin{equation*}
\freeAbove(j)=\forall_i (4\leq i\leq \tslen-3 \wedge j{-}w\leq i \leq j{+}w)\rightarrow (\lowerenv_i^B\leq A_i \leq \upperenv_i^B\vee A_i <\lowerenv^B_i\leq \lowerenv^{\upperenv^A}_i).
\end{equation*}
Similarly, an element $B_j$ is \emph{free below} $\lowerenv^A$  if all elements $A_i$ within its window are within the envelope of $B$ or cannot access below $\upperenv^{\lowerenv^A}$.
\begin{equation*}
\freeBelow(j)=\forall_i (4\leq i\leq \tslen-3 \wedge j{-}w\leq i \leq j{+}w)\rightarrow (\lowerenv_i^B\leq A_i \leq \upperenv_i^B\vee A_i >\upperenv^B_i\geq \upperenv^{\lowerenv^A}_i).
\end{equation*}
This is illustrated in Figure~\ref{fig:freeBelow}.%
If $B_j$ is free above $\upperenv^A$, then $\lbkeogh$ does not reach above $\lowerenv^{\upperenv^A}_j$ within $B_j$'s window and hence $\delta(B_j,\lowerenv^{\upperenv^A}_j)$ can be added to $\lbkeogh$. Respectively, if $B_j$ is free below $\lowerenv^A$, then $\delta(B_j,\upperenv^{\lowerenv^A}_j)$ can be added to $\lbkeogh$.

$\lbfast$ uses only the envelopes of $A$ and $B$, an envelope on the envelope of $B$ and a simple record with respect to each point $B_i$ of whether it is free above $\upperenv^A$ or below $\lowerenv^A$. The latter can be generated as a simple side effect of the calculation of the $\lbkeogh$ bridge.

\begin{theorem}\label{thm:lbfast}
If $\forall_{x,y:A_i\leq x\leq y\leq B_j\vee A_i\geq x\geq y\geq B_j}\delta(A_i,B_j) \geq \delta(A_i, y)+\delta(B_j,x)-\delta(x,y)$,
\begin{align*}
\lbfast_w(A,B)=&\minlrpaths(A,B)\\
&+\sum_{i=4}^{\tslen-3} 
    \begin{cases}
        \delta(A_i,\upperenv_i^B) & \text{if } A_i > \upperenv_i^B \\
        \delta(A_i,\lowerenv_i^B) & \text{if } A_i < \lowerenv_i^B \\
        0 & \text{otherwise}
    \end{cases}\\
&  + \sum_{i=4}^{\tslen-3}
    \begin{cases}
	\delta(B_i,\upperenv^A_i) & \text{if } \freeAbove(i) \wedge B_i > \upperenv^A_i\\
	\delta(B_i,\lowerenv^A_i) & \text{if } \freeBelow(i) \wedge B_i < \lowerenv^A_i\\
        \delta(B_i,\upperenv_i^{A})-\delta( \upperenv_i^{\lowerenv^B}, \upperenv_i^A) & \text{if }\neg\freeAbove(i)\wedge B_i > \upperenv_i^{\lowerenv^B} > \upperenv_i^A\\
        \delta(B_i,\lowerenv_i^{A})-\delta( \lowerenv_i^{\upperenv^B}, \lowerenv_i^A) & \text{if }\neg\freeBelow(i)\wedge B_i < \lowerenv_i^{\upperenv^B}  < \lowerenv_i^A\\
        0 & \text{otherwise}
    \end{cases} 
\end{align*}
 is a lower bound on $\DTW_w(A,B)$, where $\upperenv_i^{\lowerenv^B}$ denotes $i^{\mathrm{th}}$ value of the upper envelope of the lower envelope of $B$ and $ \lowerenv_i^{\upperenv^B}$  denotes  $i^{\mathrm{th}}$ value of the lower envelope of the upper envelope of $B$ and $k$ is an integer $0\leq k\leq \tslen/2$.
\end{theorem}
\begin{proof}
\allowdisplaybreaks
\begin{equation*}
\DTW_w(A,B)\hspace*{\linewidth}
\end{equation*}
\begin{equation*}
=\sum_{(i,j)\in\path}\delta(A_i,B_j)\hspace*{\linewidth}
\end{equation*}
\begin{align*}
&\geq\sum_{(i,j)\in\path}
\begin{cases}
	\delta(A_i,B_j) &  \text{if } i{\leq} 3\wedge j{\leq} 3 \hfill \myeqno{a1}\\
	\delta(A_i,B_j) &  \text{if } i{\geq} \tslen{-} 2\wedge j{\geq} \tslen{-} 3 \quad\quad\hfill \myeqno{a2}\\
      \delta(A_i, \upperenv_i^B) + \delta(B_j,\lowerenv^A_j) & \text{if } A_i>\upperenv^B_i\wedge\freeBelow(j)  \hfill \myeqno{b}\\
      \delta(A_i, \lowerenv_i^B) + \delta(B_j,\upperenv^A_j) & \text{if } A_i<\lowerenv^B_i\wedge\freeAbove(j)  \hfill \myeqno{c}\\
        \delta(B_j,A_i)+\delta(A_i, \upperenv_i^B) -\delta(\upperenv_i^B,A_i) & \text{if }  A_i>\upperenv_i^B  \wedge B_j< \lowerenv_j^{\upperenv^B} < \lowerenv_j^A  \quad \hfill \myeqno{d}\\
         \delta(B_j,A_i)+\delta(A_i, \lowerenv_i^B) -\delta(\lowerenv_i^B,A_i) & \text{if } A_i<\lowerenv_i^B  \wedge B_j> \upperenv_j^{\lowerenv^B} > \upperenv_j^A \hfill \myeqno{e} \\
       \delta(A_i,B_j) & \text{if }  A_i>\upperenv_i^B \hfill \myeqno{f}\\
       \delta(A_i,B_j) & \text{if } A_i<\lowerenv_i^B \hfill \myeqno{g}\\
       \delta(A_i,B_j) & \text{otherwise} \hfill \myeqno{h}
\end{cases}\\[5pt]
&\geq \minlrpaths(A,B)\nonumber \hspace{240pt} \myeqno{i}\\
&\quad+\sum_{i=4}^{\tslen-3} 
    \begin{cases}
        \delta(A_i,\upperenv_i^B) & \text{if } A_i > \upperenv_i^B  \hspace{190pt} \myeqno{j}\\
        \delta(A_i,\lowerenv_i^B) & \text{if } A_i < \lowerenv_i^B  \hfill \myeqno{k}\\
        0 & \text{otherwise} \hfill \myeqno{l}
    \end{cases}\\
&\quad  + \sum_{j=4}^{\tslen-3}
    \begin{cases}
	\delta(B_j,\upperenv^A_j) & \text{if } \freeAbove(j) \wedge B_j > \upperenv^A_j \hspace{90pt} \myeqno{m}\\
	\delta(B_j,\lowerenv^A_j) & \text{if } \freeBelow(j) \wedge B_j < \lowerenv^A_j \hfill \myeqno{n}\\
        \delta(B_j,\lowerenv_j^{A})-\delta( \lowerenv_j^{\upperenv^B}, \lowerenv_j^A) & \text{if } B_j < \lowerenv_j^{\upperenv^B}  < \lowerenv_j^A \hfill \myeqno{o}\\ 
               \delta(B_j,\upperenv_j^{A})-\delta( \upperenv_j^{\lowerenv^B}, \upperenv_j^A) & \text{if } B_j > \upperenv_j^{\lowerenv^B} > \upperenv_j^A \hfill \myeqno{p}\\
        0 & \text{otherwise}\hfill\myeqno{q}
    \end{cases} 
\end{align*}
\end{proof}
{\bf Notes.}
\begin{itemize}
\item[ (\ref{a1},\ref{a2}):] these capture the alignments between the first three and between the last three elements of $A$ and $B$. The $\minlrpaths(A,B)$ on line $(\ref{i})$ contributes an amount not greater than the sum of these.

\item[$(\ref{b})$:] this captures alignments where $A_i>\upperenv^B_i\geq\lowerenv^A_j>B_j$ and $(\ref{j})$ will add $\delta(A_i,\upperenv_i^B)$ and $(\ref{n})$ will add $\delta(B_i,\lowerenv^A_i)$, which sum to less than $\delta(A_i, B_j)$.

\item[$(\ref{c})$:] this captures alignments where $A_i<\lowerenv^B_i\leq\upperenv^A_j<B_j$ and $(\ref{j})$ will add $\delta(A_i,\lowerenv_i^B)$ and $(\ref{n})$ will add $\delta(B_i,\upperenv^A_i)$, which sum to less than $\delta(A_i, B_j)$.

\item[$(\ref{d})$:] this captures alignments for which $(\ref{j})$ and $(\ref{o})$ will be counted.

\item[$(\ref{e})$:] this captures alignments for which $(\ref{k})$ and $(\ref{p})$ will be counted.

\item[$(\ref{f})$:] this captures alignments for which $(\ref{j})$ and $(\ref{q})$ will apply.

\item[$(\ref{g})$:] this captures alignments for which $(\ref{k})$ and $(\ref{q})$ will apply.

\item[$(\ref{h})$:] this captures alignments for which $(\ref{l})$ and one of $(\ref{m})$, $(\ref{n})$, $(\ref{o})$ or $(\ref{p})$ will apply.

\item[$(\ref{i})$:] $\minlrpaths(A,B)$ equals the minimum value of a path through the first and last three elements of $A$ and $B$.  It cannot be greater than the value of the alignments at $(a)$ and $(\ref{b})$. 

\item[$(\ref{j})$:] this adds the distance from $A_i$ to the upper envelope of $B$ for elements $A_i$ that have alignments captured at $(\ref{b})$, $(\ref{d})$ and $(\ref{f})$.

\item[$(\ref{k})$:]  this adds the distance from $A_i$ to the lower envelope of $B$ for elements $A_i$ that have alignments captured at $(\ref{c})$, $(\ref{e})$ and $(\ref{g})$.

\item[$(\ref{l})$:] this adds zero for elements $A_i$ that fall within the envelope of $B$, whose alignments are captured at $(\ref{h})$.

\item[$(\ref{m})$:] this applies to elements $B_j$ for which all alignments $(A_i,B_j)$ are of type $(\ref{c})$ or $(\ref{h})$. If $(\ref{c})$, $(\ref{k})$ added $\delta(A_i,\lowerenv_i^B)$ and $\delta(A_i,B_j)\leq \delta(A_i,\lowerenv_i^B)+\delta(B_j,\upperenv^A_j)$ so it is safe to add the latter term. If $(\ref{h})$ then $(\ref{j})$ applied, no allowance was added for $A_i$ and hence it is also safe to add $\delta(B_j,\upperenv^A_j)$.

\item[$(\ref{n})$:] this applies to elements $B_j$ for which all alignments $(A_i,B_j)$ are of type $(\ref{b})$ or $(\ref{h})$. If $(\ref{b})$, $(\ref{j})$ added $\delta(A_i,\upperenv_i^B)$ and $\delta(A_i,B_j)\leq \delta(A_i,\upperenv_i^B)+\delta(B_j,\lowerenv^A_j)$ so it is safe to add the latter term.  If $(\ref{h})$ then $(\ref{j})$ applied, no allowance was added for $A_i$ and hence it is also safe to add $\delta(B_j,\lowerenv^A_j)$.

\item[$(\ref{o})$:] this applies to elements $B_j$ for which at least one alignment is of type $(\ref{d})$.  $\delta(A_i,\upperenv_i^B)$ is added at $(\ref{j})$, leaving $\delta(B_j,A_i)-\delta(\upperenv_i^B,A_i)\leq \delta(B_j,\lowerenv_j^{A})-\delta( \lowerenv_j^{\upperenv^B}, \lowerenv_j^A)$.

\item[$(\ref{p})$:] this applies to elements $B_j$ for which at least one alignment is of type $(\ref{e})$.  $\delta(A_i,\lowerenv_i^B)$ is added at $(\ref{k})$, leaving $\delta(B_j,A_i)-\delta(\lowerenv_i^B,A_i)\leq \delta(B_j,\upperenv_j^{A})-\delta( \upperenv_j^{\lowerenv^B}, \upperenv_j^A)$.

\item[$(\ref{q})$:] this applies to elements of $B$ for which all alignments are of type $(\ref{h})$.
\end{itemize}
\begin{figure}
\begin{center}
\includegraphics[width=1.0\columnwidth]{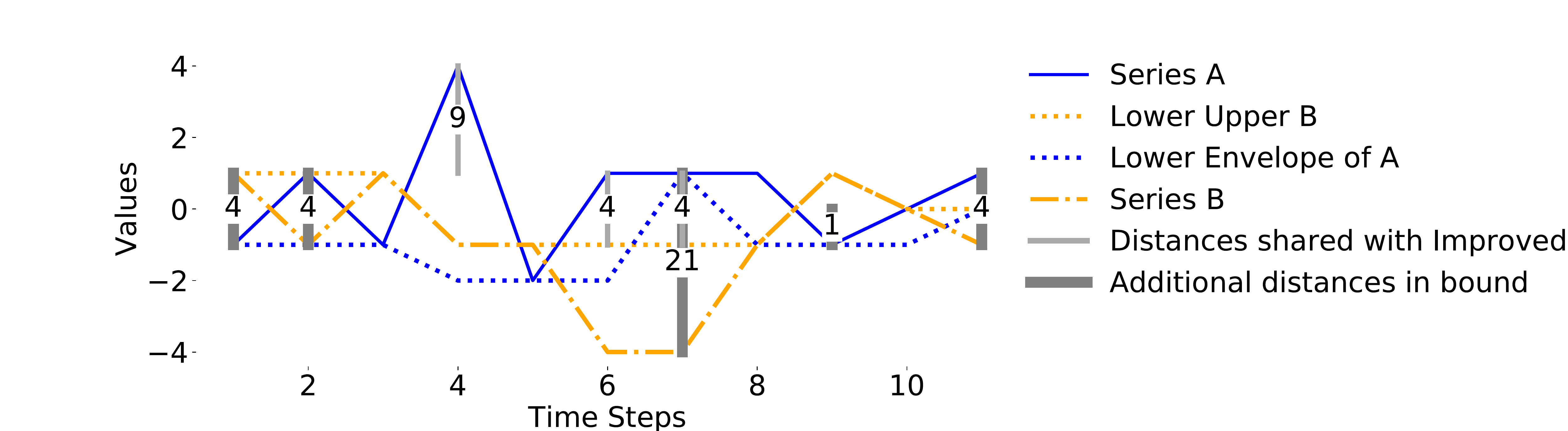}
\end{center}
\vspace*{-15pt}
\caption{Illustration of $\lbfast_1$ with $\delta(A_i,B_j)=(A_i-B_j)^2$. The dark  gray lines represent the points where $\lbfast_1$ captures greater value than \lbimproved. The medium gray areas are those captured by all of \lbkeogh, $\lbimproved$ and $\lbfast^1$. }
\label{fig:lbfast}
\end{figure}
$\lbfast$ is illustrated in Figure~\ref{fig:lbfast}. Pseudocode is presented in Algorithm~\ref{alg:fast}. It is more efficient to compute than either $\lbimproved$ or $\lbtight$.  It is tighter than $\lbkeogh$ and $\lbenhanced$ and often tighter than $\lbimproved$. It is less tight than $\lbtight$.

\begin{algorithm}
\setlength{\baselineskip}{14.75pt}
\begin{algorithmic}
\Procedure{\lbfast}{series $A$, series $B$, lower envelope of $A$ $\mathit{LA}$, upper envelope  of $A$ $\mathit{UA}$, lower envelope of $B$ $\mathit{LB}$, upper envelope  of $B$ $\mathit{UB}$, lower envelope of $UB$ $\mathit{LUB}$, upper envelope  of $LB$ $\mathit{ULB}$, window $w$, \#  left-right bands $k$, abandon value $a$}
\State $b \leftarrow\minlrpaths(A,B)$
\State $c\!\uparrow\, \leftarrow w$	\Comment{Count of the number of $\freeAbove$ elements to the left of $i$.}
\State $c\!\downarrow\, \leftarrow w$	\Comment{Count of the number of $\freeBelow$ elements to the left of $i$.}
\State $F\!\uparrow \,\leftarrow \langle\mathbf{false}\rangle^w$\Comment{$\mathbf{true}$ if $\freeAbove(i)$. Initialize all elements  as $\mathbf{false}$.}
\State $F\!\downarrow \,\leftarrow \langle\mathbf{false}\rangle^w$\Comment{$\mathbf{true}$ if $\freeBelow(i)$. Initialize all elements  as $\mathbf{false}$.}
\For{$i \leftarrow k+1$ to $\tslen-k$}\Comment{Compute the $\lbkeogh$ bridge.}
		\If{$A_i>\mathit{UB}_i$}
			\State $b \leftarrow b+\delta(A_i,\mathit{UB}_i)$
			\State $c\!\uparrow \,\leftarrow 0; c\!\downarrow \leftarrow c\!\downarrow+1$
		\ElsIf{$A_i<\mathit{LB}_i$}
			\State $b \leftarrow b+\delta(A_i,\mathit{LB}_i)$
			\State $c\!\downarrow\, \leftarrow 0; c\!\uparrow  \leftarrow c\!\uparrow+1$
		\Else
			\State $c\!\uparrow  \,\leftarrow c\!\uparrow+1; c\!\downarrow  \leftarrow c\!\downarrow+1$
		\EndIf
		\IfThen{$c\!\uparrow>2\times w$}{$\freeAbove(i)  \leftarrow\mathbf{true}$}
		\IfThen{$c\!\downarrow>2\times w$}{$\freeBelow(i)  \leftarrow\mathbf{true}$}
	\IfThen{$b>a$}{\Return{b}}
\EndFor
\For{$i \leftarrow\max(1,\tslen-k-c\!\uparrow+w)$ to $\tslen$}\Comment{Remaining free elements.}
		\State $\freeAbove(i)  \leftarrow\mathbf{true}$
\EndFor
\For{$i=\max(1,\tslen-k-c\!\downarrow+w)$ to $\tslen$}\Comment{Remaining free elements.}
		\State $\freeBelow(i)  \leftarrow\mathbf{true}$
\EndFor
\For{$ \leftarrow k+1$ to $\tslen-k$}\Comment{Allow for $B_i$ that $\lbkeogh$ could not reach.}
		\If{$\freeAbove(i)\wedge B_i>\mathit{UA}_i$}
			\State $b \leftarrow b+\delta(B_i,\mathit{UA}_i)$
		\ElsIf{$\freeBelow(i)\wedge B_i<\mathit{LA}_i$}
			\State $b \leftarrow b+\delta(B_i,\mathit{LA}_i)$
		\ElsIf{$B_i>\mathit{ULB}_i\geq \mathit{UA}_i$}
			\State $b \leftarrow b+\delta(B_i,\mathit{UA}_i)-\delta(\mathit{ULB}_i,\mathit{UA}_i)$
		\ElsIf{$B_i<\mathit{LUB}_i\leq \mathit{LA}_i$}
			\State $b \leftarrow b+\delta(B_i,\mathit{LA}_i)-\delta(\mathit{LUB}_i,\mathit{LA}_i)$
		\EndIf
	\IfThen{$b>a$}{\Return{b}}
\EndFor
\State\Return{b}
\EndProcedure
\end{algorithmic}
\caption{Algorithm for computing $\lbfast$}\label{alg:fast}
\end{algorithm}

\subsection{$\lbfastmonotone$}
In some cases a simplified variant of $\lbfast$ can be deployed. Where $\delta(A_i,B_j)=|A_i-B_j|$, $\lbfast_w(A,B)=$
\begin{align*}
\lbfastmonotone_w(A,B)=&\minlrpaths(A,B)\\
&+\sum_{i=4}^{\tslen-3} 
    \begin{cases}
        \delta(A_i,\upperenv_i^B) & \text{if } A_i > \upperenv_i^B \\
        \delta(A_i,\lowerenv_i^B) & \text{if } A_i < \lowerenv_i^B \\
        0 & \text{otherwise}
    \end{cases}\\
&  + \sum_{i=4}^{\tslen-3}
    \begin{cases}
	\delta(B_i,\upperenv^A_i) & \text{if } \freeAbove(i) \wedge B_i > \upperenv^A_i\\
	\delta(B_i,\lowerenv^A_i) & \text{if } \freeBelow(i) \wedge B_i < \lowerenv^A_i\\
        \delta(B_i,\upperenv_i^{\lowerenv^B}) & \text{if }\neg\freeAbove(i)\wedge B_i > \upperenv_i^{\lowerenv^B} > \upperenv_i^A\\
        \delta(B_i,\lowerenv_i^{\upperenv^B}) & \text{if }\neg\freeBelow(i)\wedge B_i < \lowerenv_i^{\upperenv^B}  < \lowerenv_i^A\\
        0 & \text{otherwise}
    \end{cases} 
\end{align*}
$\lbfastmonotone(A,B)$ does not require that $\forall_{x,y:A_i\leq x\leq y\leq B_j\vee A_i\geq x\geq y\geq B_j}\delta(A_i,B_j) \geq \delta(A_i, y)+\delta(B_j,x)-\delta(x,y)$ and is a lower bound for $\DTW$ where $\delta(A_i,B_j)$ increases monotonically with $|A_i-B_j|$. This is the class of $\delta$ for which $\lbkeogh$, $\lbimproved$ and $\lbenhanced$ are lower bounds of $\DTW$.

\subsection{$\lbenhancedfast$}\label{sec:enhancedwebb}
As the $\DTW$ window increases in size, lower bounds based on envelopes, such as $\lbkeogh$, $\lbimproved$, $\lbtight$ and $\lbfast$, are likely to decline in tightness due to each point in the envelope representing a maximum or minimum over an ever increasing proportion of a whole series. In this case, the method underlying $\lbenhanced$ is likely to excel, as it does not use envelopes. To this end, a parameterized variant of $\lbfast$ that employs the left and right bands of $\lbenhanced$ is likely to come to the fore.

\begin{align*}
&\lbenhancedfast_w^k(A,B)=\\
&~~~~~~~~~~\sum_{i=1}^k\left[\min(\leftset^w_i)+\min(\rightset^w_{\tslen-i+1})\right]\\
&~~~~~~~~~~+\sum_{i=k+1}^{\tslen-k} 
    \begin{cases}
        \delta(A_i,\upperenv_i^B) & \text{if } A_i > \upperenv_i^B \\
        \delta(A_i,\lowerenv_i^B) & \text{if } A_i < \lowerenv_i^B \\
        0 & \text{otherwise}
    \end{cases}\\
&~~~~~~~~~~  + \sum_{i=k+1}^{\tslen-k}
    \begin{cases}
	\delta(B_i,\upperenv^A_i) & \text{if } \freeAbove(i) \wedge B_i > \upperenv^A_i\\
	\delta(B_i,\lowerenv^A_i) & \text{if } \freeBelow(i) \wedge B_i < \lowerenv^A_i\\
        \delta(B_i,\upperenv_i^{A})-\delta( \upperenv_i^{\lowerenv^B}, \upperenv_i^A) & \text{if }\neg\freeAbove(i)\wedge B_i > \upperenv_i^{\lowerenv^B} > \upperenv_i^A\\
        \delta(B_i,\lowerenv_i^{A})-\delta( \lowerenv_i^{\upperenv^B}, \lowerenv_i^A) & \text{if }\neg\freeBelow(i)\wedge B_i < \lowerenv_i^{\upperenv^B}  < \lowerenv_i^A\\
        0 & \text{otherwise}
    \end{cases} 
\end{align*}

\section{Experiments}\label{sec:experiments}
In order to assess the practical merits of the new lower bounds relative to the prior state of the art, we compare their performance on the 85 dataset Bakeoff Paper \cite{bagnall2017great} version of the widely utilized UCR benchmark time series datasets  \cite{ucrarchive}. We use this version because of the availability of comparative results of many techniques  for it.  

All experiments are performed using single threaded implementations of the respective algorithms in Java and executed on Intel Xeon CPU E5-2680 v3  2.50GHz CPUs.  In the interests of reproducible science the source code is made available at \url{ https://github.com/GIWebb/DTWBounds}.

The experiments are run in a heterogeneous grid environment.  While the amount of RAM may differ from experiment to experiment, all comparative runs for a single dataset were performed on a single machine, ensuring that comparative results for each dataset are commensurable even though results between datasets may not be. 

We use $\delta=(A_i-B_j)^2$.

\subsection{Tightness}

We first seek to quantify the relative tightness of our new bounds relative to $\lbkeogh$ and $\lbimproved$.  Relative tightness will vary greatly depending on window size.  To obtain an evaluation that is relevant to real world practice, we use for each dataset the window size recommended by the archive. These recommended window sizes are those that provide most accurate nearest neighbor classification using leave-one-out cross-validation on the training set. Some recommended window sizes are $0$.  There is no value in computing a linear time lower bound for a window size of zero, as it is quicker to simply compute the full distance.  Hence we do not include these datasets in this evaluation and use only the 60 datasets with recommended window sizes of one or more.

\begin{figure}[t]
\begin{minipage}{0.475\columnwidth}%
\includegraphics[width=\columnwidth]{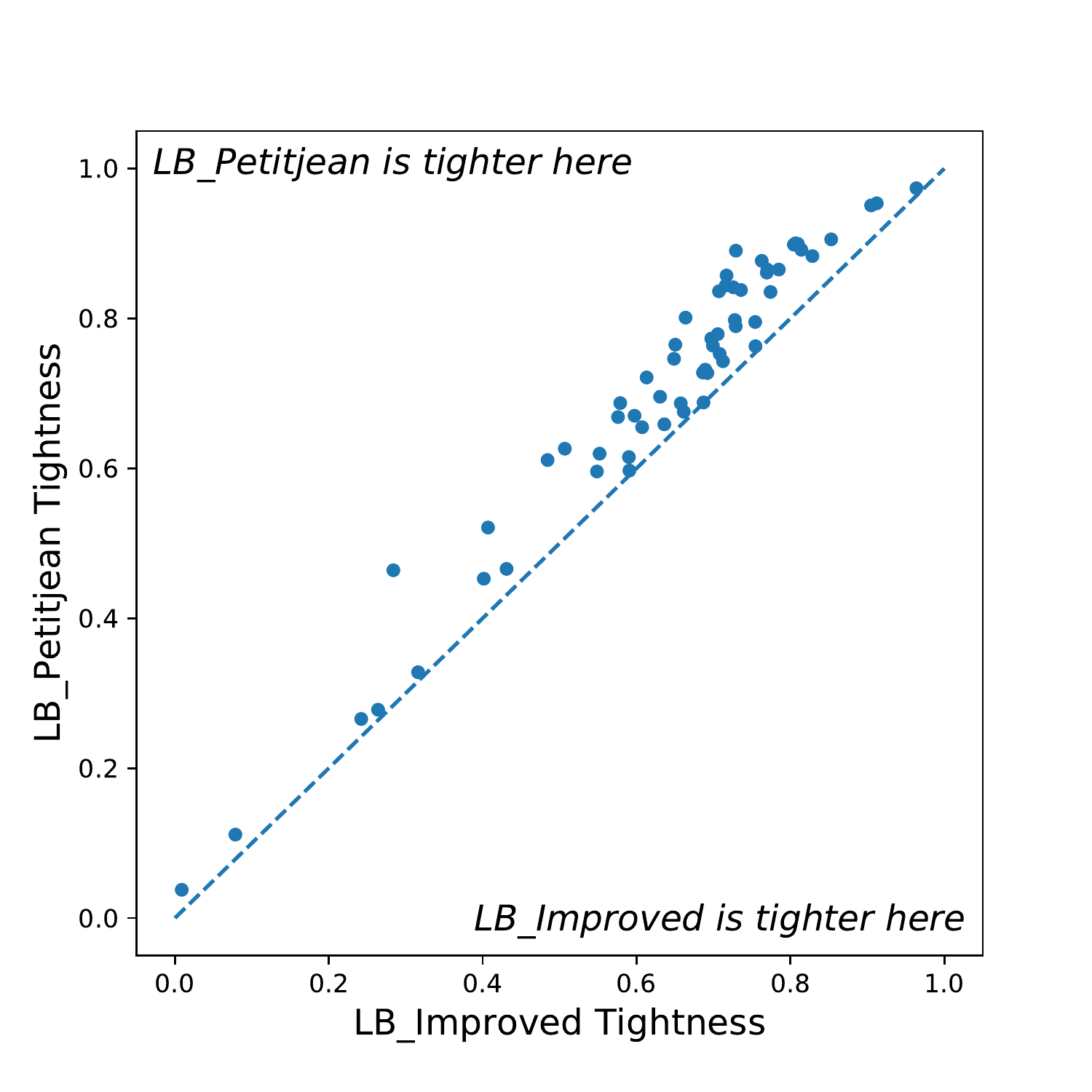}
\vspace*{-30pt}
\caption{Relative tightness of $\lbtight$ and $\lbimproved$}\label{fig:tightImpTight}
\end{minipage}\hspace*{.05\columnwidth}
\begin{minipage}{0.475\columnwidth}%
\includegraphics[width=\columnwidth]{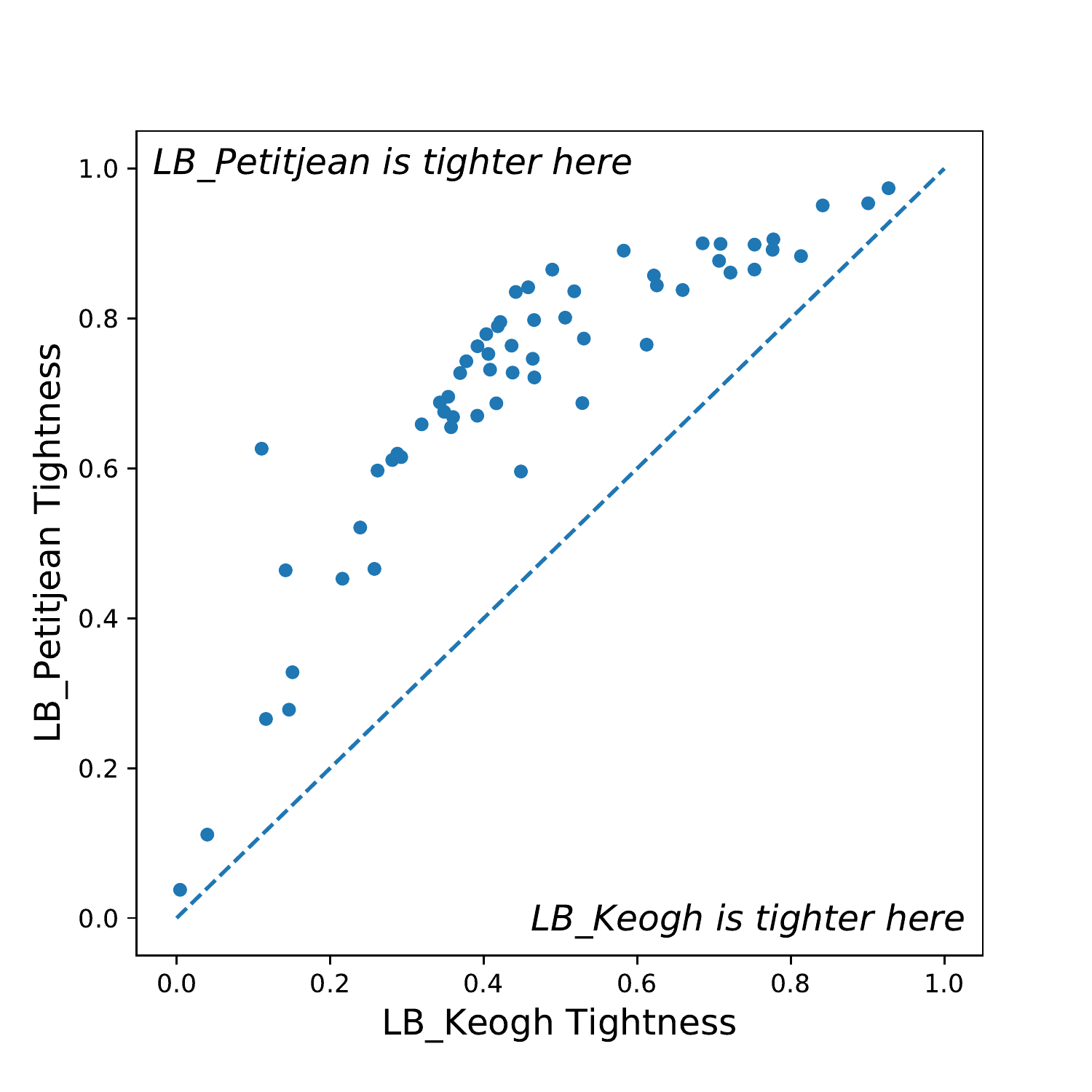}
\vspace*{-30pt}\caption{Relative tightness of $\lbtight$ and $\lbkeogh$}
\end{minipage}\hspace*{.05\columnwidth}
\end{figure}

\begin{figure}[t]
\begin{minipage}{0.475\columnwidth}%
\includegraphics[width=\columnwidth]{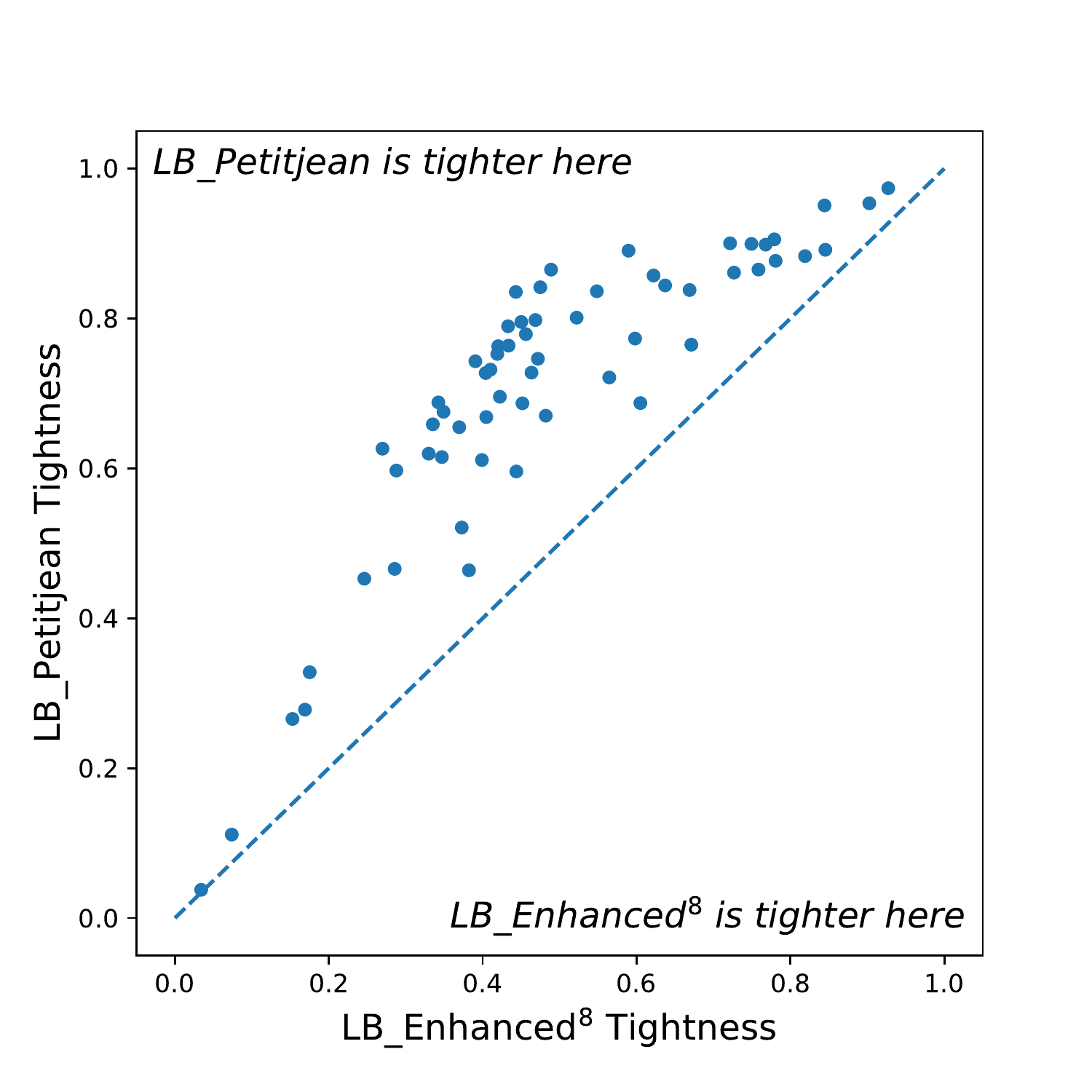}
\vspace*{-30pt}\caption{Relative tightness of $\lbtight$ and $\lbenhanced^8$}\label{fig:tightEnhancedFastPetitjean}
\end{minipage}\hspace*{.05\columnwidth}
\begin{minipage}{0.475\columnwidth}%
\includegraphics[width=\columnwidth]{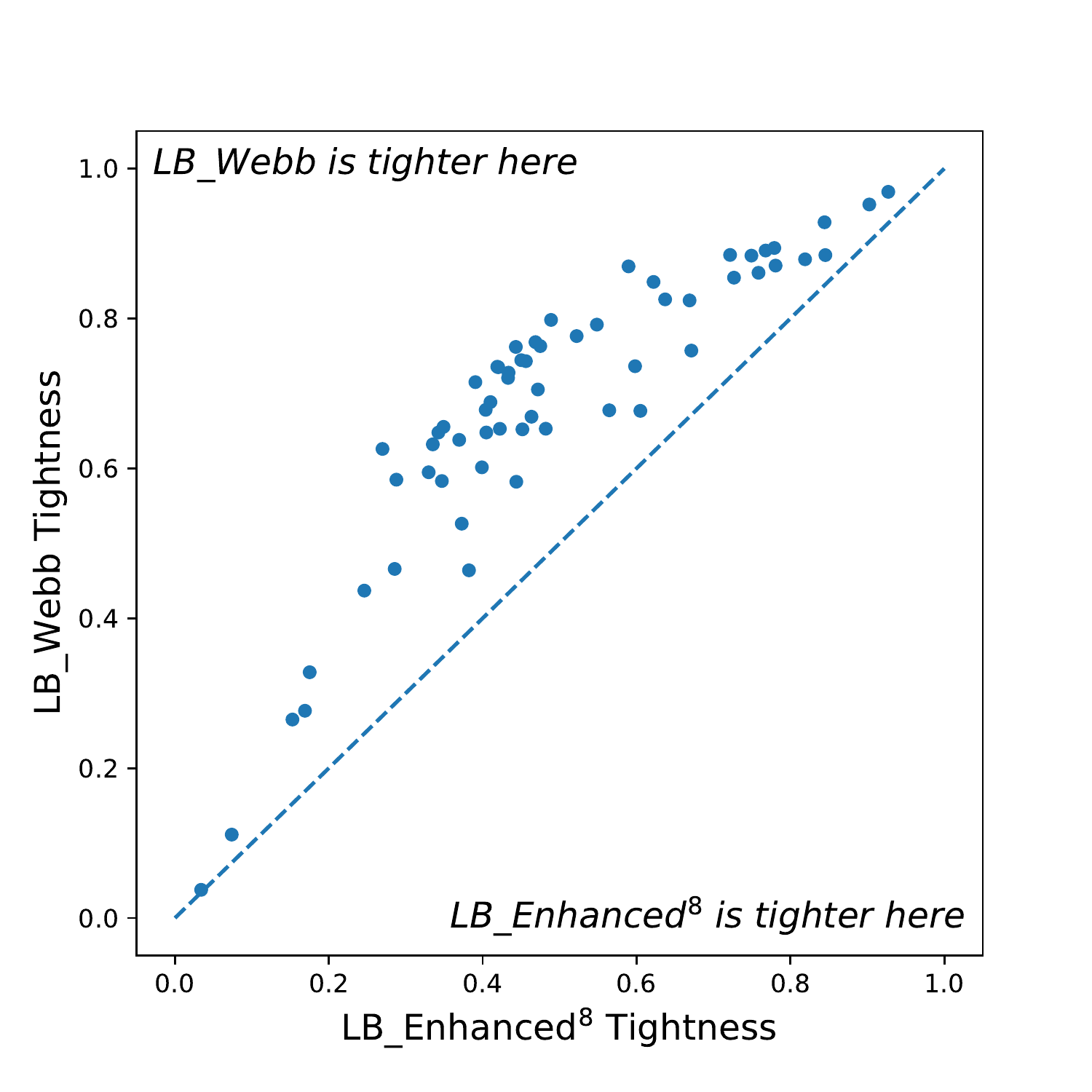}
\vspace*{-30pt}\caption{Relative tightness of $\lbfast$ and $\lbenhanced^8$}\label{fig:tightEnhancedFastWebb}
\end{minipage}
\end{figure}

We compute tightness for every pair of a training ($T$) and a test ($Q$) series.  We calculate the tightness of a lower bound $\lambda_w(Q,T)$ on $\cdtw(Q,T)$ as $\lambda_w(Q,T)/\cdtw(Q,T)$.  We exclude pairs $(Q,T)$ for which $\cdtw(Q,T)=0.0$.  We compare the average tightness on each dataset for each of $\lbtight$ and $\lbfast$ against each of $\lbimproved$ and $\lbkeogh$ in Figures~\ref{fig:tightKeoghFast}, \ref{fig:tightImprovedWebb} and \ref{fig:tightImpTight} to \ref{fig:tightEnhancedFastWebb}.
$\lbtight$ is always tighter than $\lbimproved$ and often substantially tighter.  In the most extreme case, for ShapeletSim, $\lbtight$ has average tightness of $0.038$ while $\lbimproved$ has average tightness of only $0.009$.  The advantage of $\lbtight$ relative to $\lbkeogh$ is even greater.  

$\lbfast$ is also necessarily tighter than $\lbkeogh$. Figure \ref{fig:tightKeoghFast} shows that the advantage is often substantial. It is tighter on average than $\lbimproved$  for $47$ datasets and less tight for $13$.  Figure \ref{fig:tightImprovedWebb} shows that it is never substantially less tight than $\lbimproved$ and is often substantially tighter.

The tightness of $\lbenhanced$ varies with $k$, tending to, but not always, growing with $k$.  Tan \emph{et. al.}{} \cite{TanEtAl19} identify $k=8$ as providing an effective trade-off between tightness and computation.  Both $\lbtight$ and $\lbfast$ are always tighter than $\lbenhanced^8$. See Figures \ref{fig:tightEnhancedFastPetitjean} and \ref{fig:tightEnhancedFastWebb}.

\subsection{Classification times with optimal windows}
The utility of a lower bound for a given application is determined by the trade-off it provides between tightness and speed.  $\lbtight$ requires slightly more computation than $\lbimproved$, while $\lbfast$ requires substantially less.  To assess the relative utilities of these trade-offs, we next test the efficiency of nearest neighbor search utilizing these bounds, again employing optimal window sizes and hence limiting the evaluation to the 60 datasets for which the optimal window size is greater than zero.  

We conduct two types of nearest neighbor search.  Each finds for each test series $Q$, $\argmin_{T\in\mathcal{T}}\cdtw(Q,T)$, the training series $T$ that is the nearest neighbor to $Q$ using DTW with the optimal window, $w$.  The first approach, described in Algorithm~\ref{fig:unsorted-proc}, tests a test series $Q$ against each training series $T$ in random order, first applying the relevant lower bound and then only computing the full distance if the lower bound is less than the best distance so far.  The second approach, described in Algorithm~\ref{fig:sorted-proc}, for each query $Q$, first computes the lower bound for every training series $T$, then sorts the training series in ascending order and finally computes the full distances on successive training series until the minimum distance found is less than the next lower bound.  

Each approach is repeated ten times for each dataset and average results are presented in order to smooth out variations in time due to extraneous factors and in performance due to randomization for the random order approach. The~envelopes for the training series,  $\lowerenv^T$,  $\upperenv^T$, $\upperenv^{\lowerenv^T}$ and $\lowerenv^{\upperenv^T}$, are  precalculated and the time for calculating these envelopes is not included in the experimental timings.  The calculation of all other envelopes is included in timings.  Calculation of $\lowerenv^{\projection}$ and $\upperenv^{\projection}$ is considered part of a the calculation of the bound and must be done once for each bound calculation.  In contrast, calculation of $\upperenv^Q$,  ${\lowerenv^Q}$, $\lowerenv^{\upperenv^Q}$,  ${\upperenv^{\lowerenv^Q}}$  need only be done once per query series.
\begin{algorithm}[t]
\begin{algorithmic}
\Procedure{RandExp}{set of query series $\mathcal{Q}$, set of training series $\mathcal{T}$, lower bound $\lambda$}
\For{$Q\in \mathcal{Q}$}
	\If{$\lambda$ requires  $\upperenv^Q$ and ${\lowerenv^Q}$}
		\State Calculate and save $\upperenv^Q$ and ${\lowerenv^Q}$ 
	\EndIf
	\State $b\leftarrow \emptyset$
	\For{$T\in \mathcal{T}$}
		\If{$b=\emptyset$}
			\State $d\leftarrow\dtw(Q,T)$
			\State $b\leftarrow T$
		\Else
			\If{$\lambda(Q,T,b)<b$}
				\State $d^\prime\leftarrow\dtw(Q,T)$
				\If{$d^\prime<d$}
					\State $b\leftarrow T$
					\State $d\leftarrow d^\prime$
				\EndIf
			\EndIf
		\EndIf
	\EndFor
\EndFor
\EndProcedure
\end{algorithmic}
\caption{Experimental procedure for nearest neighbor search with random order}\label{fig:unsorted-proc}
\end{algorithm}
\begin{algorithm}[t]
\begin{algorithmic}
\Procedure{SortedExp}{set of query series $Q$, set of training series $\mathcal{T}$, lower bound $\lambda$}
\For{$Q\in \mathcal{Q}$}
	\If{$\lambda$ requires  $\upperenv^Q$ and ${\lowerenv^Q}$}
		\State Calculate and save $\upperenv^Q$ and ${\lowerenv^Q}$ 
	\EndIf
	\For{$T\in \mathcal{T}$}
		\State $D[T]\leftarrow\lambda(Q,T)$
	\EndFor
	\State $d\leftarrow \infty$
	\For{$T\in \mathcal{T}$ in ascending order on $D[T]$ \textbf{until} $D[T]\geq d$ }
		\If{$d= \infty$}
			\State $d\leftarrow\dtw(Q,T)$
			\State $b\leftarrow T$
		\Else
			\If{$D[T]<b$}
				\State $d^\prime\leftarrow\dtw(Q,T)$
				\If{$d^\prime<d$}
					\State $b\leftarrow T$
					\State $d\leftarrow d^\prime$
				\EndIf
			\EndIf
		\EndIf
	\EndFor
\EndFor
\EndProcedure
\end{algorithmic}
\caption{Experimental procedure for nearest neighbor search with sorted series}\label{fig:sorted-proc}
\end{algorithm}

Note that  early abandoning is used for the random order search, whereby the lower bound calculation is abandoned as soon as the cumulative calculation of the lower bound exceeds the distance to the nearest neighbor found so far.  This is not possible for the sorted approach, as the lower bounds are computed before any of the full distances.

\begin{figure}
\begin{minipage}{0.48\columnwidth}%
\vspace*{-20pt}
\includegraphics[width=1.1\columnwidth]{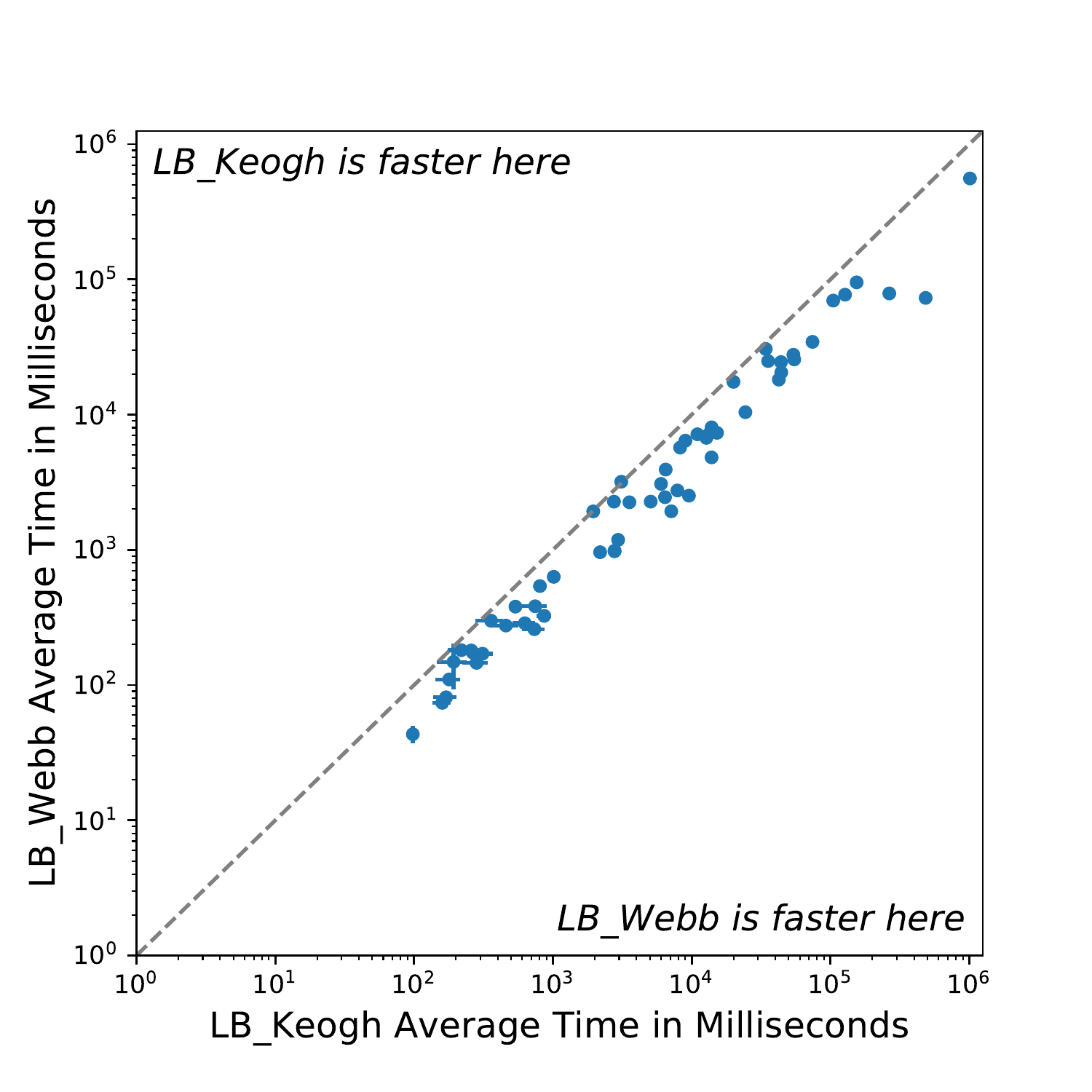}
\vspace*{-20pt}
\caption{Relative compute time for nearest neighbor search in random order using $\lbfast$ and $\lbkeogh$. 
}\label{fig:nosortKeoghWebb}
\end{minipage}\hspace*{.04\columnwidth}
\begin{minipage}{0.48\columnwidth}%
\vspace*{-20pt}
\includegraphics[width=1.1\columnwidth]{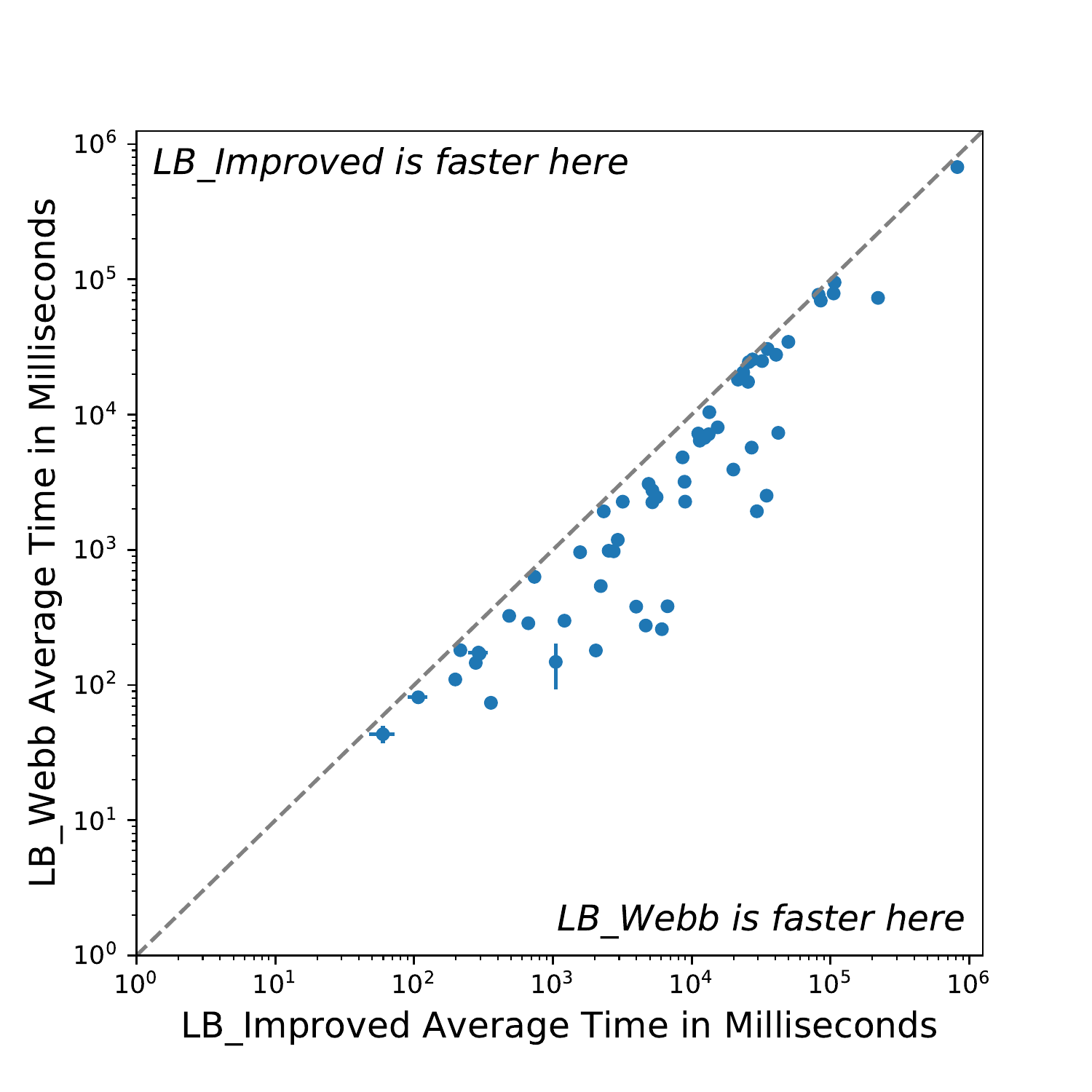}
\vspace*{-20pt}
\caption{Relative compute time for nearest neighbor search in random order using $\lbfast$ and $\lbimproved$. 
}\label{fig:nosortImprovedWebb}
\end{minipage}
\end{figure}

\begin{figure}
\begin{minipage}{0.48\columnwidth}%
\vspace*{-20pt}
\includegraphics[width=1.1\columnwidth]{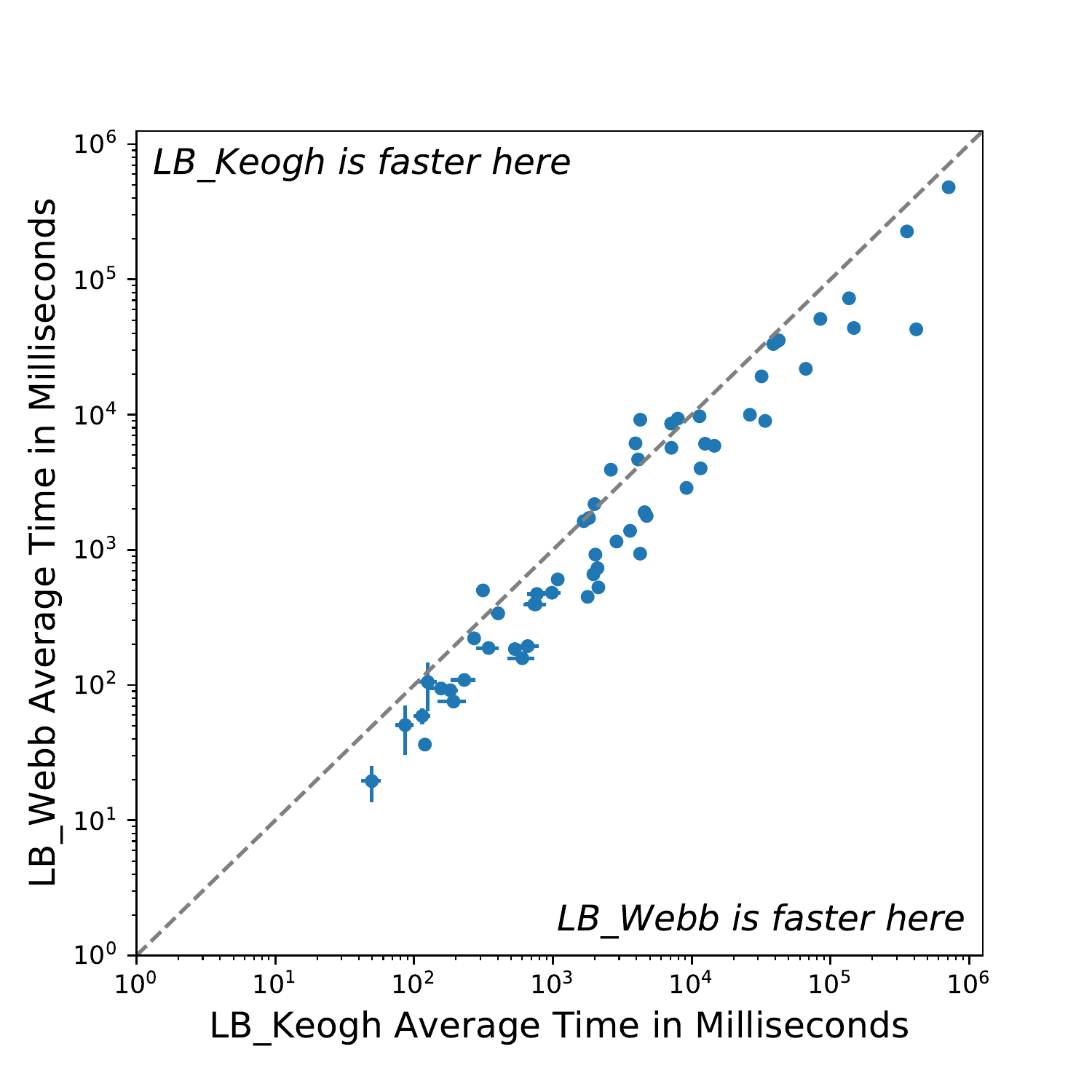}
\vspace*{-20pt}
\caption{Relative compute time for nearest neighbor search in sorted order using $\lbfast$ and $\lbkeogh$. 
}\label{fig:sortKeoghWebb}
\end{minipage}\hspace*{.04\columnwidth}
\begin{minipage}{0.48\columnwidth}%
\vspace*{-20pt}
\includegraphics[width=1.1\columnwidth]{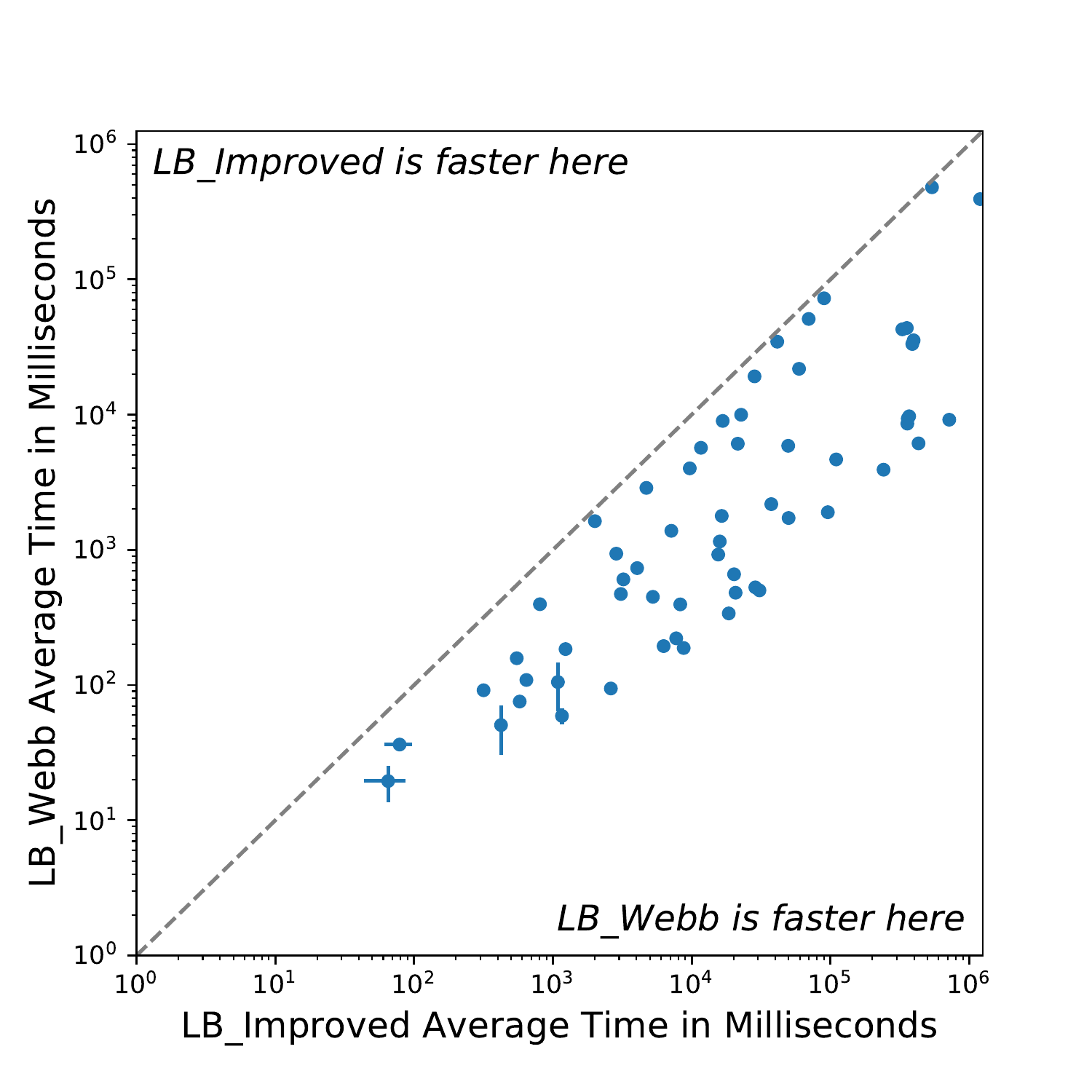}
\vspace*{-20pt}
\caption{Relative compute time for nearest neighbor search in sorted order using $\lbfast$ and $\lbimproved$. 
}\label{fig:sortImprovedWebb}
\end{minipage}
\end{figure}

Figures \ref{fig:nosortKeoghWebb} to \ref{fig:sortImprovedWebb} present the comparisons of $\lbfast$ and $\lbtight$ against each of $\lbkeogh$ and $\lbimproved$.  These and all subsequent relative compute-time scatter plots plot the mean of ten runs together with error bars that show one standard deviation either side of the mean in each dimension.  As the plots use log-log scale, these plots extend further to the left than right and further below than above the point.  In~most cases the error bars are not visible, as they do not extend beyond the dot centered on the mean.

$\lbfast$ delivers faster nearest neighbor DTW search than either $\lbkeogh$ or $\lbimproved$  for the majority of datasets under both approaches. When the training examples are processed in random order, $\lbfast$  delivers faster nearest neighbor DTW search than $\lbkeogh$ for 59 out of 60 datasets.   The greatest difference is for the FordB dataset for which $\lbkeogh$ takes on average $8$ minutes and 4 seconds and $\lbfast$ takes 1 minute and 12  seconds. When the training series are sorted on the lower bounds, $\lbfast$ is faster 52 times and $\lbkeogh$ 8.   The greatest difference is again for the FordB dataset for which $\lbkeogh$ takes on average $6$ minutes and 54 seconds compared with  $42$ seconds for $\lbfast$.  
\begin{figure}
\begin{minipage}{0.48\columnwidth}%
\vspace*{-20pt}
\includegraphics[width=1.1\columnwidth]{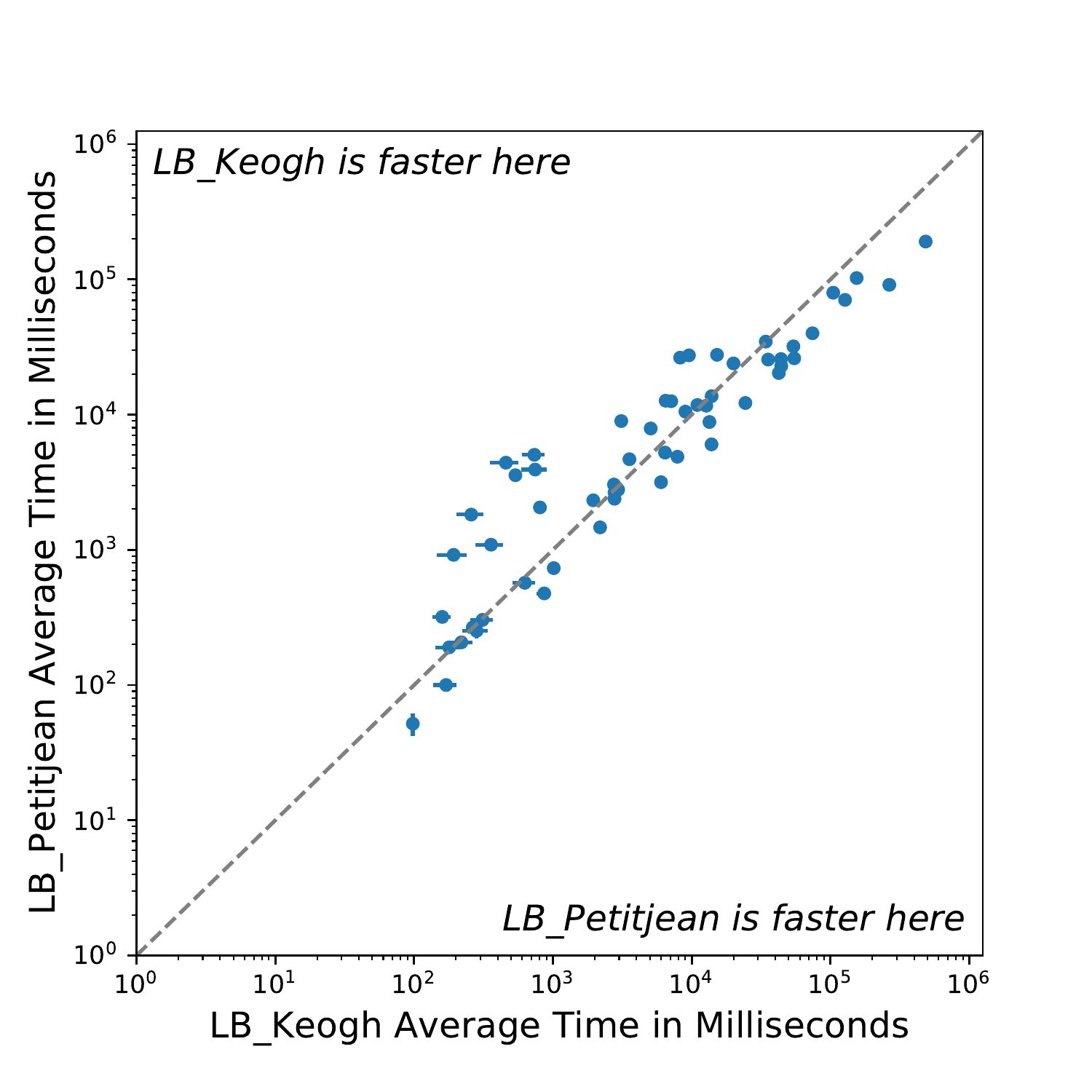}
\vspace*{-20pt}
\caption{Relative compute time for nearest neighbor search in random order using $\lbtight$ and $\lbkeogh$. 
}\label{fig:nosortKeoghPetitjean}
\end{minipage}\hspace*{.04\columnwidth}
\begin{minipage}{0.48\columnwidth}%
\vspace*{-20pt}
\includegraphics[width=1.1\columnwidth]{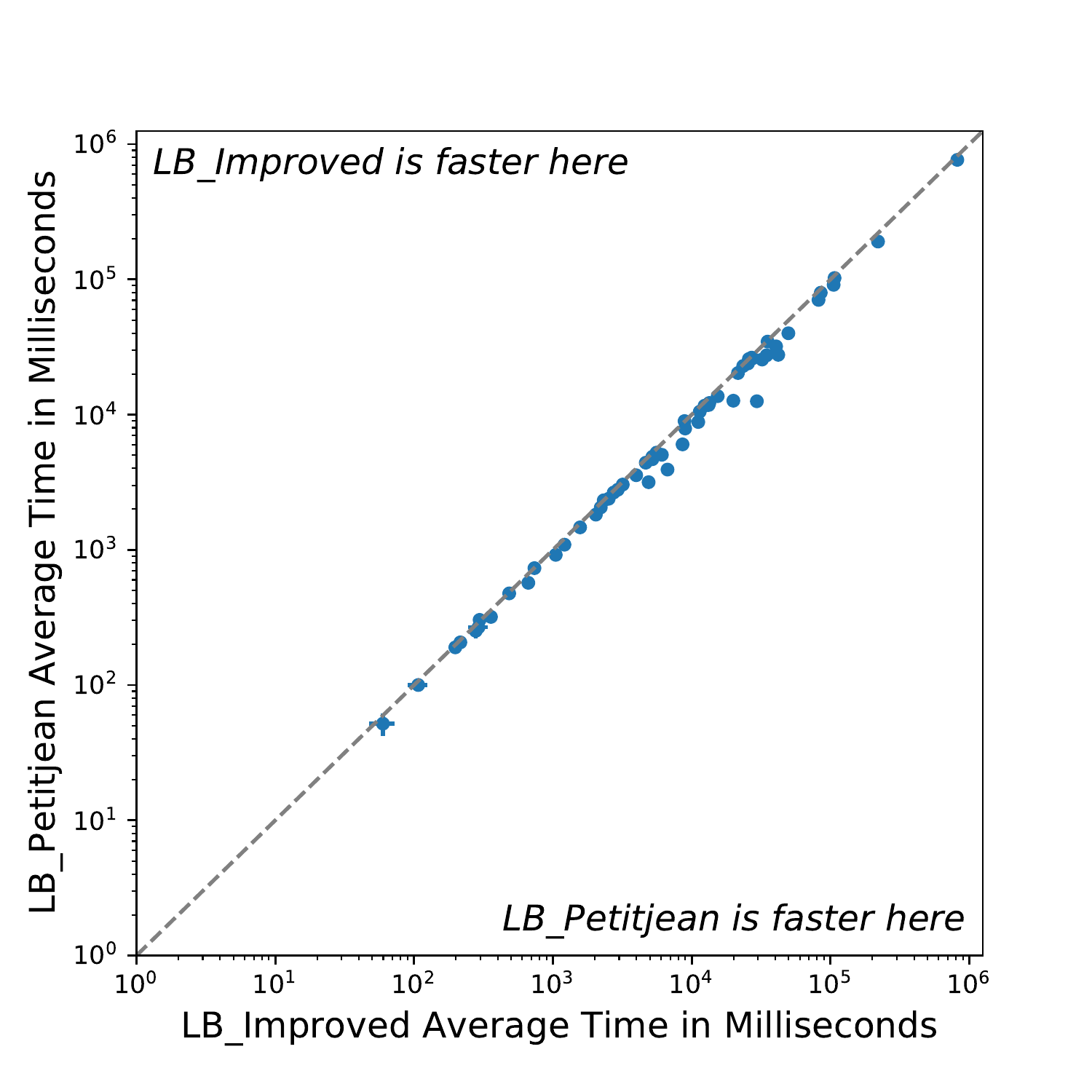}
\vspace*{-20pt}
\caption{Relative compute time for nearest neighbor search in random order using $\lbtight$ and $\lbimproved$. 
}\label{fig:nosortImprovedPetitjean}
\end{minipage}
\end{figure}
\begin{figure}
\begin{minipage}{0.48\columnwidth}%
\vspace*{-20pt}
\includegraphics[width=1.1\columnwidth]{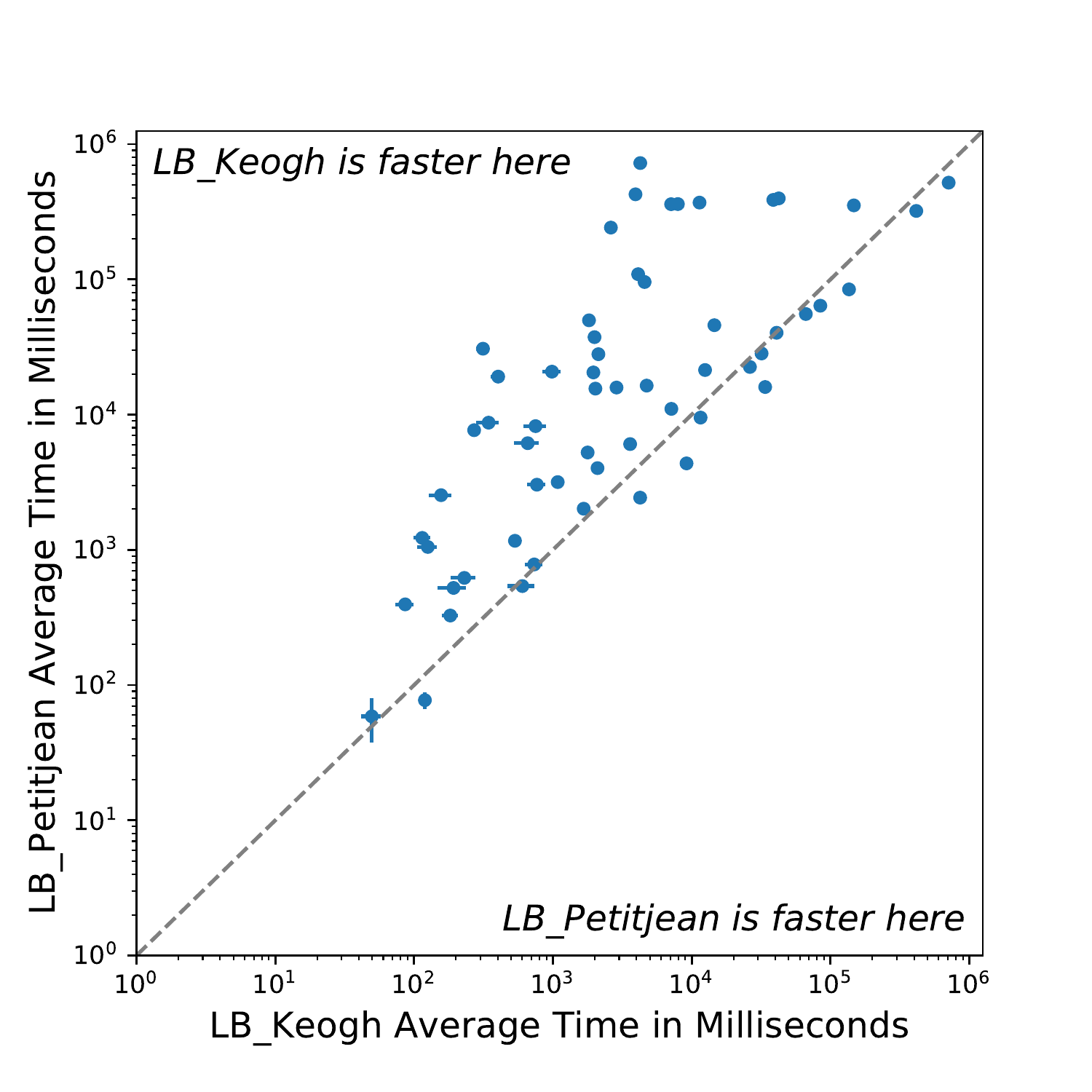}
\vspace*{-20pt}
\caption{Relative compute time for nearest neighbor search in sorted order using $\lbtight$ and $\lbkeogh$.
}\label{fig:sortKeoghWebb}
\end{minipage}\hspace*{.04\columnwidth}
\begin{minipage}{0.48\columnwidth}%
\vspace*{-20pt}
\includegraphics[width=1.1\columnwidth]{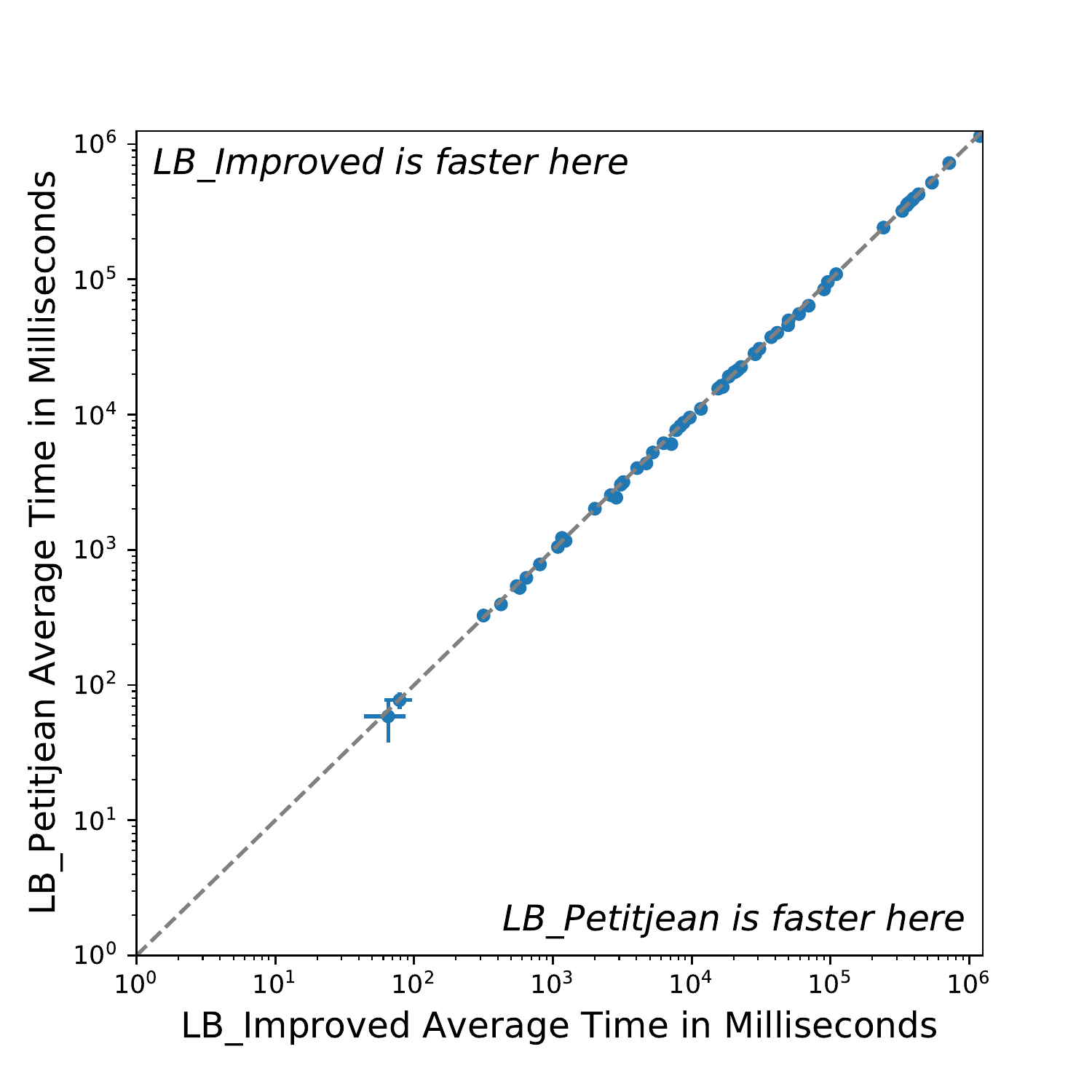}
\vspace*{-20pt}
\caption{Relative compute time for nearest neighbor search in sorted order using $\lbtight$ and $\lbimproved$. 
}\label{fig:sortImprovedPetitjean}
\end{minipage}
\end{figure}

$\lbtight$ supports faster DTW nearest neighbor search than $\lbimproved$ for the majority of datasets when the series are not sorted, due to its improved tightness and similar compute time. When the series are sorted, however, the slight increase in compute time sometimes outweighs the increase in tightness.  It tends to be more efficient than $\lbkeogh$ when the datasets are unsorted, because its more expensive computations can be abandoned as soon as the lower bound is sufficiently tight to allow a candidate to be pruned. It tends to be less efficient than $\lbkeogh$ when the candidates are sorted on the bound, as the bound must be calculated in full for every candidate.

\begin{figure}
\begin{minipage}{0.48\columnwidth}%
\vspace*{-20pt}
\includegraphics[width=1.1\columnwidth]{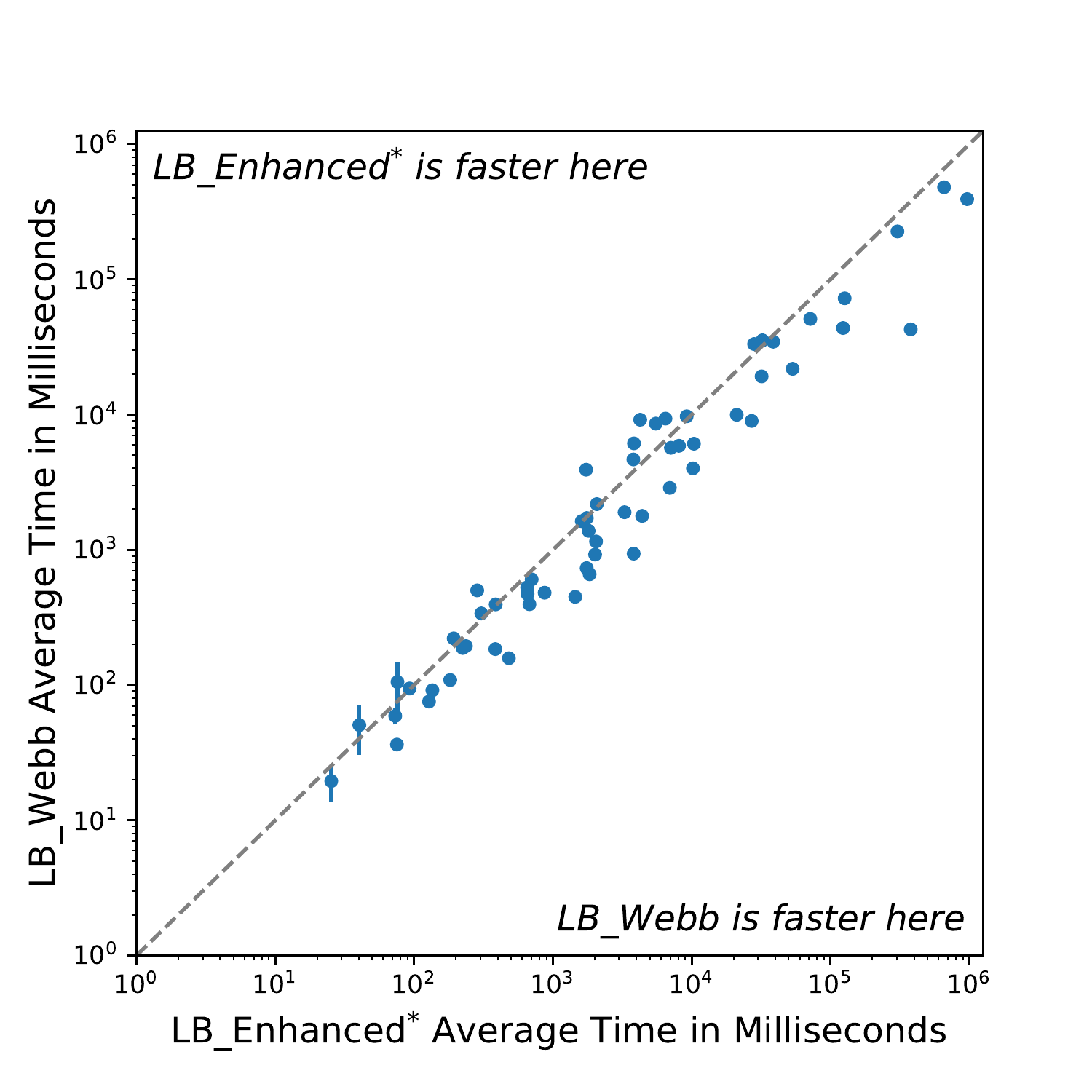}
\vspace*{-20pt}
\caption{Relative compute time for nearest neighbor search in sorted order using $\lbfast$ and $\lbenhanced$ with the best performing value of $k$. 
}\label{fig:sortWebbEnhanced}
\end{minipage}\hspace*{.04\columnwidth}
\begin{minipage}{0.48\columnwidth}%
\vspace*{-20pt}
\includegraphics[width=1.1\columnwidth]{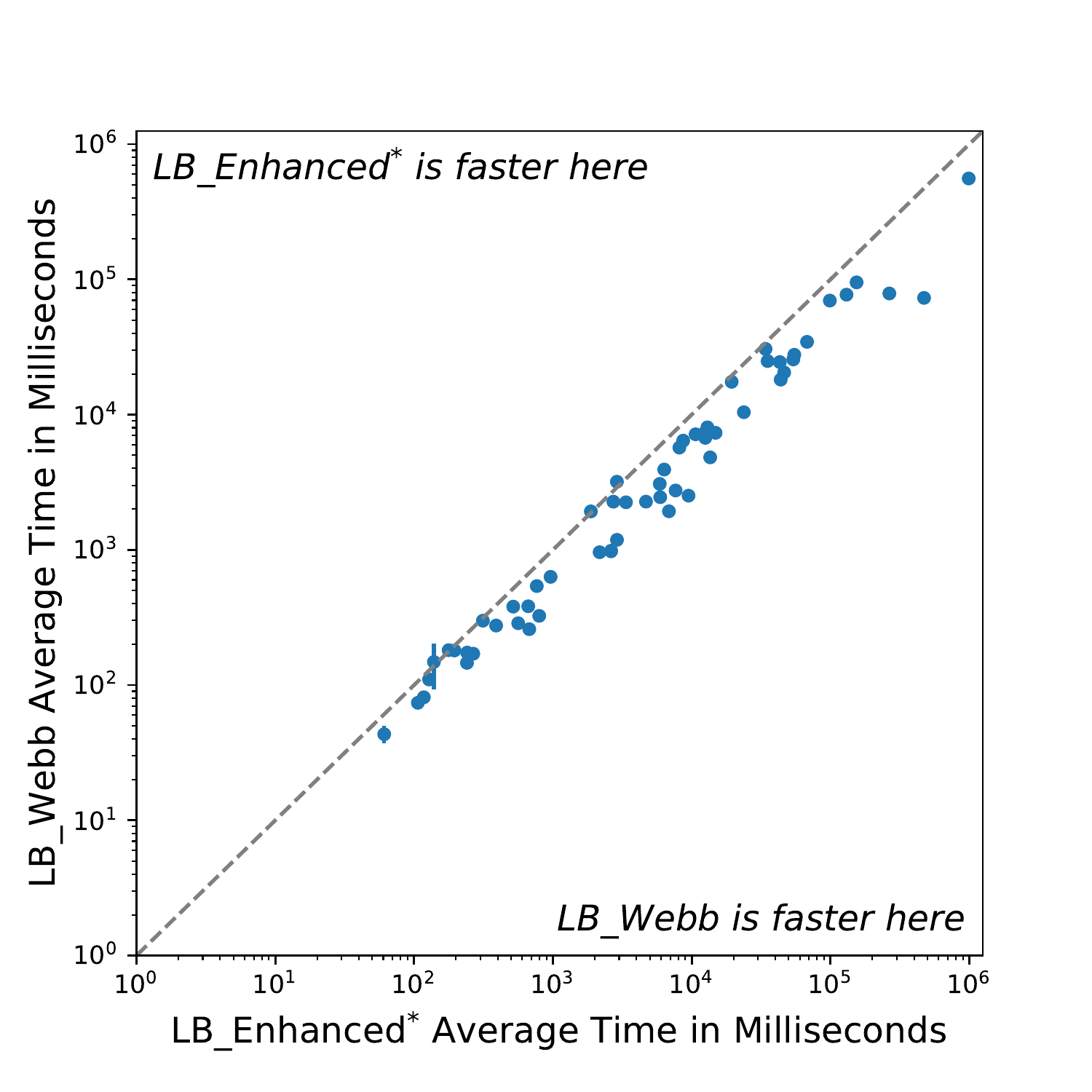}
\vspace*{-20pt}
\caption{Relative compute time for nearest neighbor search in random order using $\lbfast$ and $\lbenhanced$ with the best performing value of $k$. 
}\label{fig:nosortWebbEnhanced}
\end{minipage}
\end{figure}

We next compare $\lbfast$ to $\lbenhanced$ on the 60 datasets with optimal window size greater than zero.  We first look at the case where the datsets are first sorted by lower bound. As the performance of $\lbenhanced$ varies with parameter $k$, we test all values of $k$ up to 16, at which point $\lbenhanced$ is clearly beyond its optimal setting.  The total average (again over ten runs) compute time for $\lbfast$ is 26 minutes. $\lbenhanced$ using the fastest setting of $k$ for each dataset requires on average 49 minutes to obtain the same results. Note that this assessment does not take account of the issue of how the optimal value for $k$ might be predetermined. Figure~\ref{fig:sortWebbEnhanced} shows the scatter plot of times per dataset for $\lbfast$ relative to the best performance of $\lbenhanced$ for any setting of $k$.

The results for processing the datasets in random order are presented in Figure \ref{fig:nosortWebbEnhanced}. $\lbfast$ is faster for  56  datasets and slower for 4.  $\lbfast$  requires on average 54 minutes to classify all 60 test sets while $\lbenhanced$ with optimal $k$ requires 1 hour and 50 minutes. 

For both these tasks, $\lbfast$ is faster and unlike $\lbenhanced$ does not require any parameter tuning.

\subsection{Classification times with varying window sizes}
To explore how the bounds interact with varying window sizes, and to assess whether the advantage to $\lbfast$ is specific to the 60 datasets whose optimal window sizes are greater than zero, we here assess classification time when the training data are sorted on the respective lower bound and the window size is a specified percentage of series length.  We use three window sizes, $1\%$ (Table \ref{tab:g1}), $10\%$ (Table \ref{tab:g10}) and $20\%$ (Table \ref{tab:g20}).  In each case we round fractional values up in order to avoid windows of size zero. In each of the three  tables of results, for each pairwise comparison, we present first a win/loss summary and then the total time taken, on average, in hours, minutes and seconds to classify the entire 85 test sets in the repository, followed by the ratio of the two times. The win/loss summary states the number of datasets for which the first algorithm required less time to classify the test set (wins) and the number for which the second algorithm required less time.  There are no draws. Thus, when the window size is 1\% of the total time series length (Table \ref{tab:g1}), $\lbfast$ requires less time than $\lbkeogh$ on 62 datasets and more on 23 and requires  9  minutes and 13 seconds to classify the entire repository, which is just 37\% of the 24 minutes and 39 seconds required by $\lbkeogh$.  
\begin{table}
\begin{tabular}{lc@{/}cc@{=}r}
\bf Comparison&\bf win&\bf loss&\multicolumn{2}{l}{\bf Total time ratio}\\
\ensuremath{\lbfast} vs {\lbkeogh}& 62 & 23 & 0:09:13/0:24:39 & 0.37\\
\ensuremath{\lbfast} vs {\lbimproved}& 85& 0 & 0:09:13/3:32:25 & 0.04\\
\ensuremath{\lbfast} vs \ensuremath{\lbtight}& 85 &  0 & 0:09:13/3:32:05 & 0.04\\
\ensuremath{\lbfast} vs \ensuremath{\lbenhanced^{*}}& 30& 55 & 0:09:13/0:22:00 & 0.42\\
\ensuremath{\lbtight} vs {\lbkeogh}&  4 & 81 & 3:32:05/0:24:39 & 8.60\\
\ensuremath{\lbtight} vs {\lbimproved}& 56 & 29 & 3:32:05/3:32:25 & 1.00\\
\ensuremath{\lbtight} vs \ensuremath{\lbfast}&  0 & 85 & 3:32:05/0:09:13 & 22.97\\
\ensuremath{\lbtight} vs \ensuremath{\lbenhanced^{*}}&  4 & 81 & 3:32:05/0:22:00 & 9.64\\
\end{tabular}
\vspace*{-8pt}\caption{Results on all UCR datasets, $w=0.01\cdot\tslen$}\label{tab:g1}
\end{table}
\begin{table}
\begin{tabular}{lc@{/}cc@{=}r}
\bf Comparison&\bf win&\bf loss&\multicolumn{2}{l}{\bf Total time ratio}\\
\ensuremath{\lbfast} vs {\lbkeogh}& 84 &  1 & 1:21:45/2:58:00 & 0.46\\
\ensuremath{\lbfast} vs {\lbimproved}& 85 &  0 & 1:21:45/4:53:11 & 0.28\\
\ensuremath{\lbfast} vs \ensuremath{\lbtight}& 85 &  0 & 1:21:45/4:43:24 & 0.29\\
\ensuremath{\lbfast} vs \ensuremath{\lbenhanced^{*}}& 79 &  6 & 1:21:45/2:06:49 & 0.64\\
\ensuremath{\lbtight} vs {\lbkeogh}& 22 & 63 & 4:43:24/2:58:00 & 1.59\\
\ensuremath{\lbtight} vs {\lbimproved}& 66 & 19 & 4:43:24/4:53:11 & 0.97\\
\ensuremath{\lbtight} vs \ensuremath{\lbfast}&  0 & 85 & 4:43:24/1:21:45 & 3.47\\
\ensuremath{\lbtight} vs \ensuremath{\lbenhanced^{*}}& 11 & 74 & 4:43:24/2:06:49 & 2.23\\
\end{tabular}
\vspace*{-8pt}\caption{Results on all UCR datasets, $w=0.10\cdot\tslen$}\label{tab:g10}
\end{table}
\begin{table}
\begin{tabular}{lc@{/}cc@{=}r}
\bf Comparison&\bf win&\bf loss&\multicolumn{2}{l}{\bf Total time ratio}\\
\ensuremath{\lbfast} vs {\lbkeogh}& 85 &  0 & 3:23:09/5:25:55 & 0.62\\
\ensuremath{\lbfast} vs {\lbimproved}& 85 &  0 & 3:23:09/7:04:20 & 0.48\\
\ensuremath{\lbfast} vs \ensuremath{\lbtight}& 85 &  0 & 3:23:09/6:45:17 & 0.50\\
\ensuremath{\lbfast} vs \ensuremath{\lbenhanced^{*}}& 76 &  9 & 3:23:09/3:51:42 & 0.88\\
\ensuremath{\lbtight} vs {\lbkeogh}& 29 & 56 & 6:45:17/5:25:55 & 1.24\\
\ensuremath{\lbtight} vs {\lbimproved}& 76 &  9 & 6:45:17/7:04:20 & 0.96\\
\ensuremath{\lbtight} vs \ensuremath{\lbfast}&  0 & 85 & 6:45:17/3:23:09 & 1.99\\
\ensuremath{\lbtight} vs \ensuremath{\lbenhanced^{*}}& 15 & 70 & 6:45:17/3:51:42 & 1.75\\
\end{tabular}
\vspace*{-8pt}\caption{Results on all UCR datasets, $w=0.20\cdot\tslen$}\label{tab:g20}
\end{table}

As the window size increases from 1 to 10 to 20\%, $\lbfast$  consistently provides an advantage relative to $\lbkeogh$, but the magnitude of that advantage decreases.  Thus, at a window size of 20\% of total series length, $\lbfast$ is faster for all  datasets.  At this window size, $\lbfast$ requires 3 hours and 23 minutes on average to classify all 85 datasets in the repository while $\lbkeogh$ requires 5 hours and 25 minutes.

$\lbfast$ is faster than $\lbimproved$ for all datasets at all three window sizes.  The relative magnitude of the improvement decreases as window size increases.  Nonetheless, $\lbfast$ requires only 3 hours and 23 minutes to classify all 85 datasets at a window size of 20\% of total series length compared to 7 hours and 4 minutes for $\lbimproved$.

\begin{figure}
\begin{minipage}{0.48\columnwidth}%
\vspace*{-20pt}
\includegraphics[width=1.1\columnwidth]{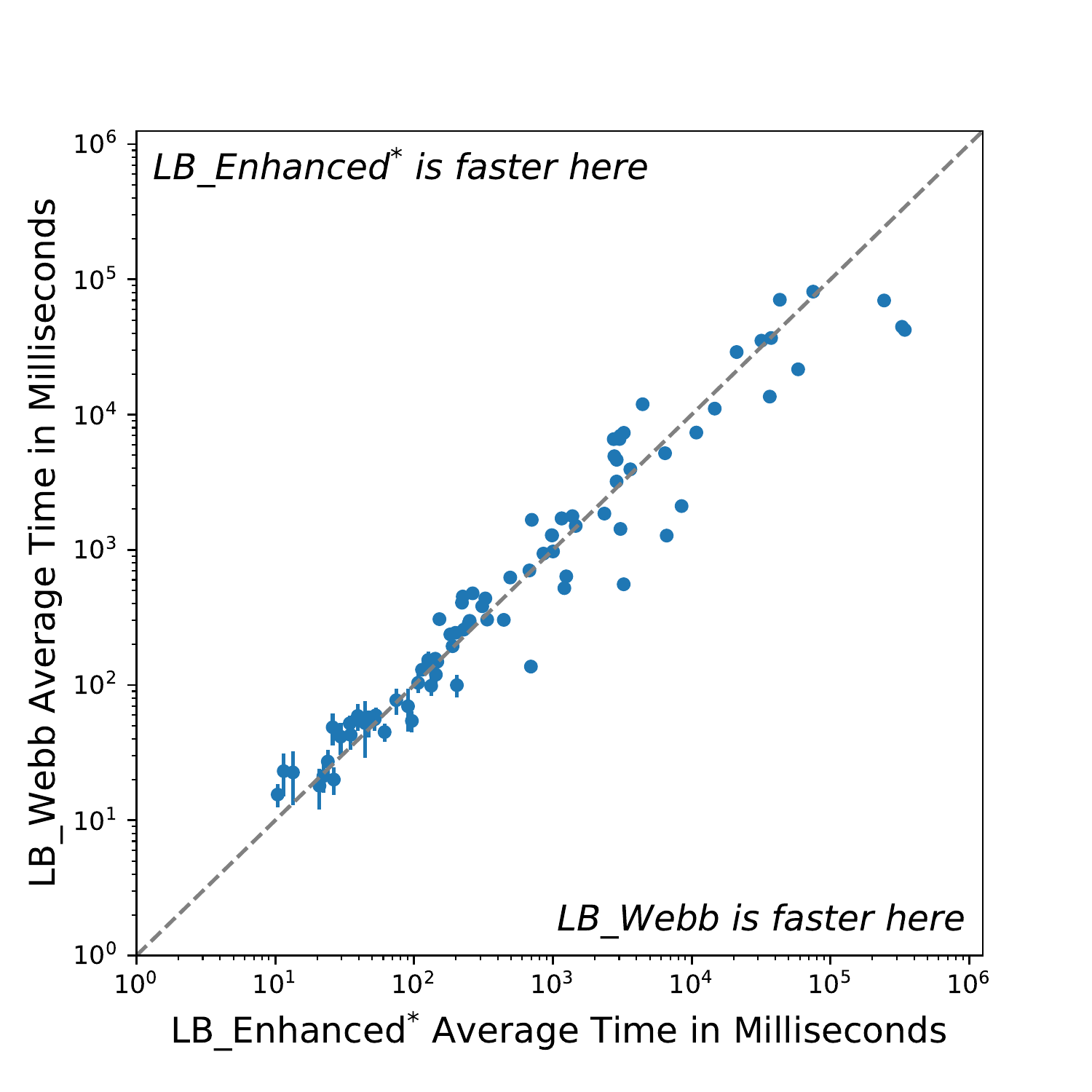}
\vspace*{-20pt}
\caption{Relative compute time  for nearest neighbor search in sorted order with window size set to 1\% of series length.  $\lbfast$ vs $\lbenhanced$ with the most effective value of $k$ for each dataset. $\lbfast$ is faster for  30  datasets and slower for 55. $\lbfast$  requires on average under 9 minutes to classify all 85 test sets while $\lbenhanced$ requires 22 minutes. }\label{fig:nosortImprovedWebbg1}
\end{minipage}\hspace*{.04\columnwidth}
\begin{minipage}{0.48\columnwidth}%
\vspace*{-20pt}
\includegraphics[width=1.1\columnwidth]{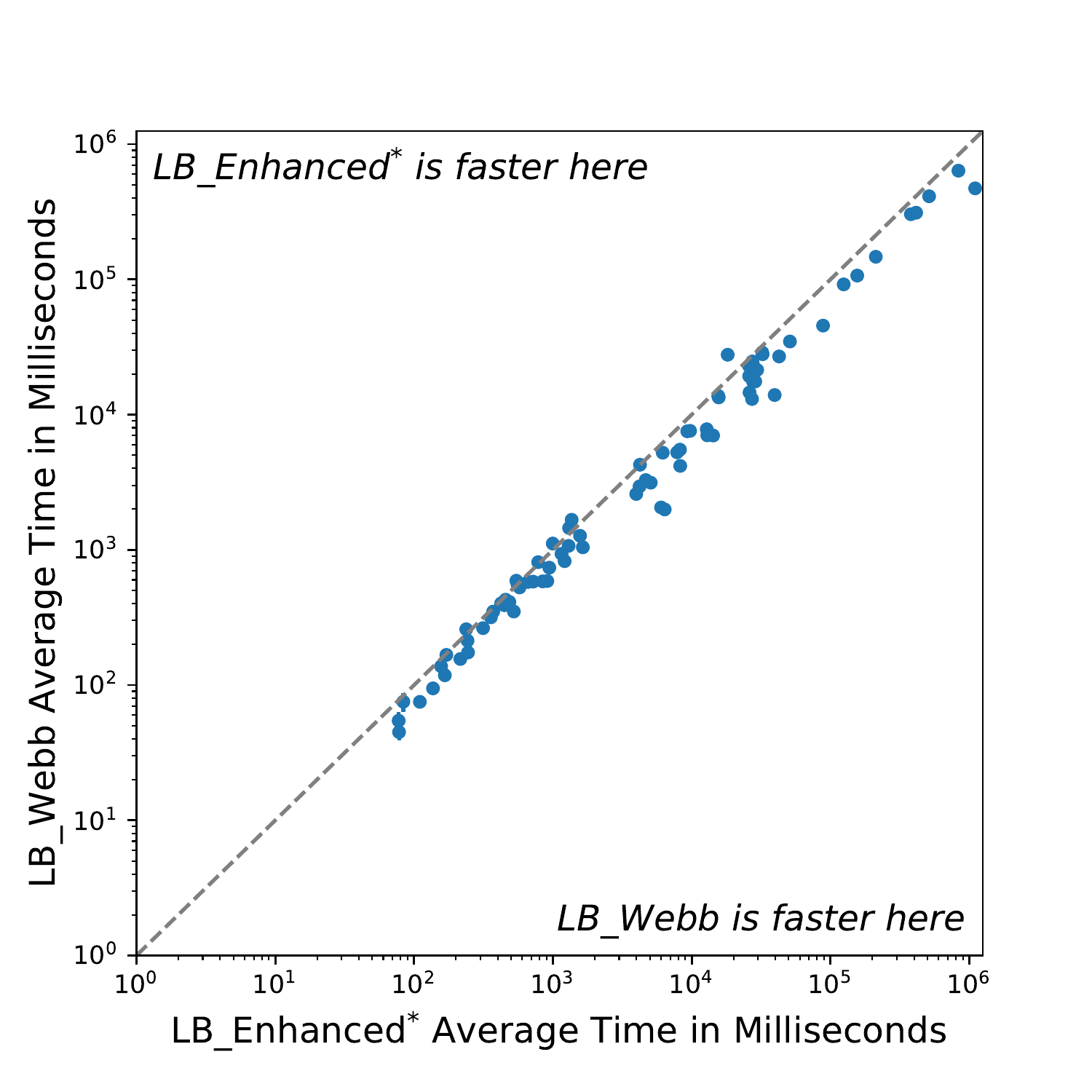}
\vspace*{-20pt}
\caption{Relative compute time  for nearest neighbor search in sorted order with window size set to 20\% of series length.  $\lbfast$ vs $\lbenhanced$ with the most effective value of $k$ for each dataset. $\lbfast$ is faster for76  datasets and slower for 9. $\lbfast$  requires on average 3 hours and 23 minutes to classify all 85 test sets while $\lbenhanced$ requires 3 hours and 51 minutes. }\label{fig:nosortImprovedWebb}
\end{minipage}
\end{figure}

$\lbfast$ is faster than $\lbenhanced$ at the best setting for $k$ for only 30 out of the 85 datasets when the window size is set to 1\% of series length.  Nonetheless it requires less than half the time to classify the full repository.  As illustrated in Figure \ref{fig:nosortImprovedWebbg1}, this is due to the  losses being for datasets that require less computation and the  wins being predominantly for datasets that require more.
As the window size increases, $\lbfast$ wins more often relative to $\lbenhanced$ with the optimal setting of $k$, but the magnitudes of the wins and losses shrink, as illustrated in Figure \ref{fig:nosortImprovedWebb}.%

$\lbtight$ delivers faster nearest neighbor search than $\lbimproved$ for the majority of datasets at all window sizes.  However, the magnitudes of the wins and losses are extremely small.  Neither $\lbtight$ nor $\lbimproved$ is competitive with $\lbkeogh$, $\lbenhanced$ or $\lbfast$ on these tasks. This is because presorting does not offer any chance to early abandon lower bound computation.  As a result, these bounds are often computed to greater precision than required by the task.  As shown in Figure \ref{fig:nosortKeoghPetitjean}, $\lbtight$ is more likely to excel in contexts where early abandon can be deployed.

\section{On the effect of the left and right paths}

In this section we investigate the role of the left and right paths that are incorporated in the two new bounds.  To~this end we compare $\lbfast$ to a variant, $\lbfastnolr$ without the left and right paths and a variant, $\lbenhancedfast$, that replaces the left and right paths with left and right bands of the form used by $\lbenhanced$, presented in Section \ref{sec:enhancedwebb}.  
\begin{align*}
&\lbfastnolr_w(A,B)=\\
&~~~~~~~~~~\sum_{i=1}^{\tslen} 
    \begin{cases}
        \delta(A_i,\upperenv_i^B) & \text{if } A_i > \upperenv_i^B \\
        \delta(A_i,\lowerenv_i^B) & \text{if } A_i < \lowerenv_i^B \\
        0 & \text{otherwise}
    \end{cases}\\
&~~~~~~~~~~  + \sum_{i=1}^{\tslen}
    \begin{cases}
	\delta(B_i,\upperenv^A_i) & \text{if } \freeAbove(i) \wedge B_i > \upperenv^A_i\\
	\delta(B_i,\lowerenv^A_i) & \text{if } \freeBelow(i) \wedge B_i < \lowerenv^A_i\\
        \delta(B_i,\upperenv_i^{A})-\delta( \upperenv_i^{\lowerenv^B}, \upperenv_i^A) & \text{if }\neg\freeAbove(i)\wedge B_i > \upperenv_i^{\lowerenv^B} > \upperenv_i^A\\
        \delta(B_i,\lowerenv_i^{A})-\delta( \lowerenv_i^{\upperenv^B}, \lowerenv_i^A) & \text{if }\neg\freeBelow(i)\wedge B_i < \lowerenv_i^{\upperenv^B}  < \lowerenv_i^A\\
        0 & \text{otherwise}
    \end{cases} 
\end{align*}

\begin{figure}
\begin{minipage}{0.48\columnwidth}%
\vspace*{-20pt}
\includegraphics[width=\columnwidth]{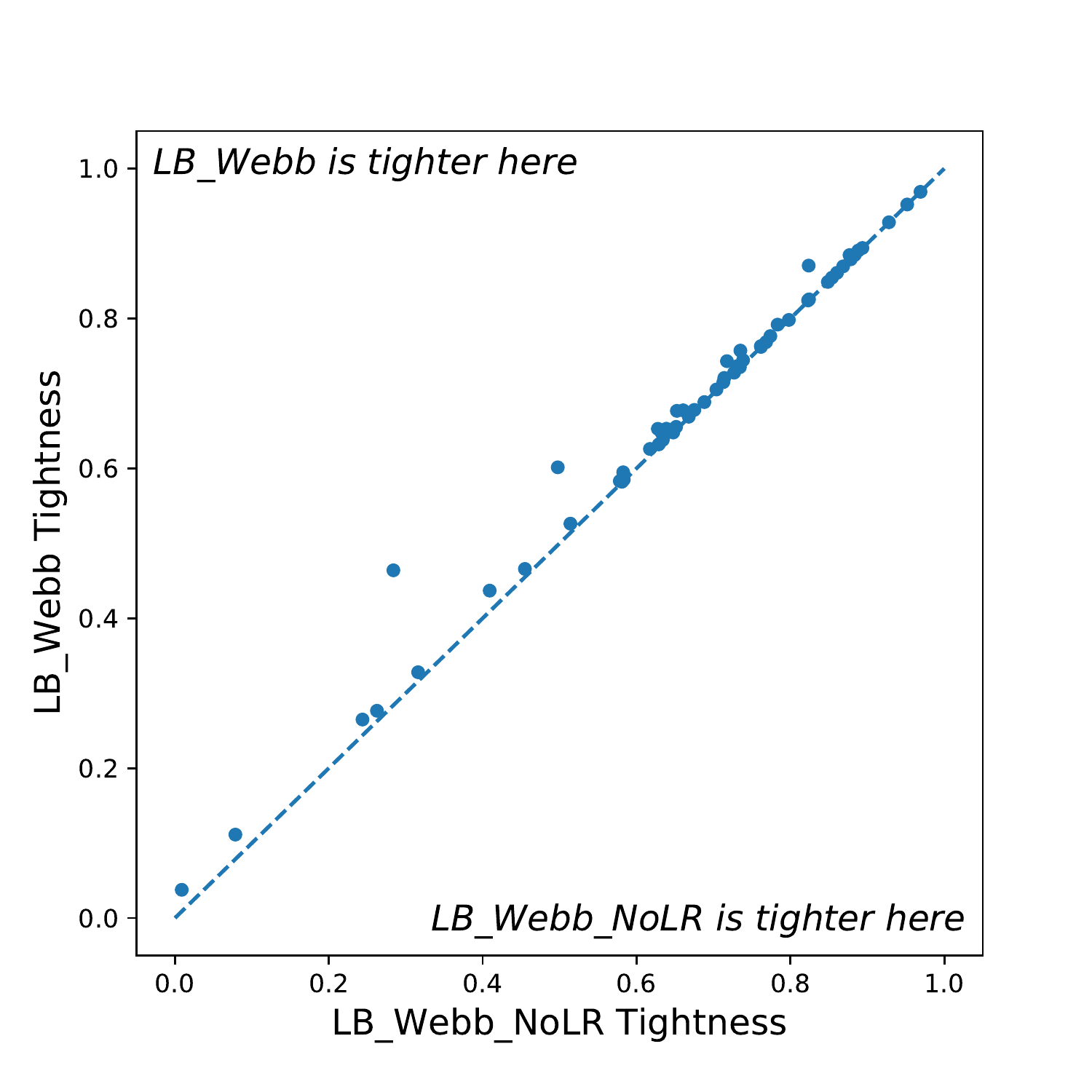}
\vspace*{-20pt}
\caption{Relative tightness of $\lbfast$ and $\lbfastnolr$}\label{fig:fastNoLR}
\end{minipage}\hspace*{.05\columnwidth}
\begin{minipage}{0.48\columnwidth}%
\vspace*{-20pt}
\includegraphics[width=\columnwidth]{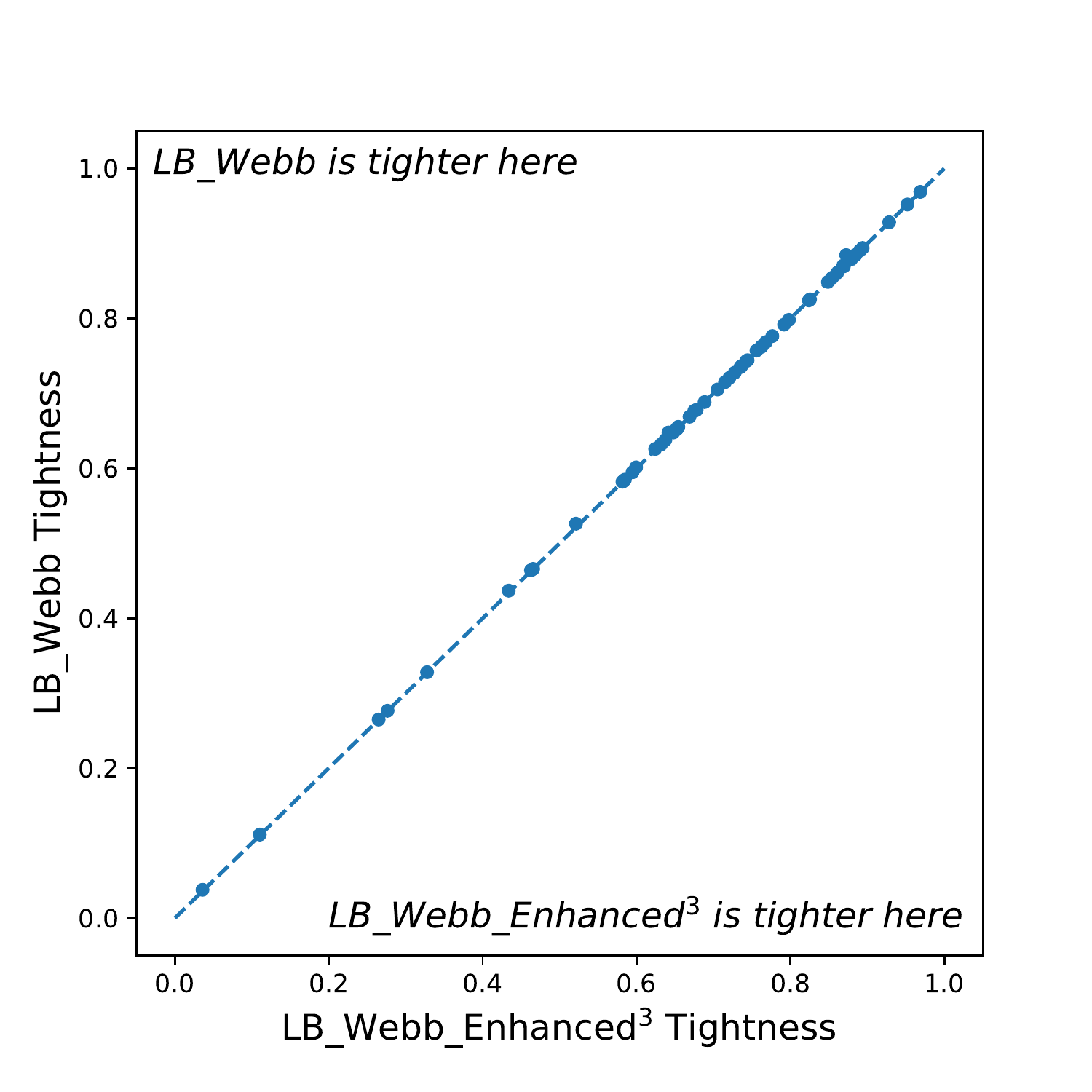}
\vspace*{-20pt}
\caption{Relative tightness of $\lbfast$ and $\lbenhancedfast^3$}\label{fig:enhancedFast}
\end{minipage}
\end{figure}

Figure \ref{fig:fastNoLR} shows the relative tightness of $\lbfast$ and $\lbfastnolr$ using the optimal window size on all 60 datasets for which the optimal window size is greater than zero.  $\lbfast$ provides a tighter lower bound for every dataset except wafer, for which its tightness is  0.96891 versus 0.96904 a difference of just 0.00007.  For many datasets the difference is small, but for a few datasets, where there is considerable variation in the start and end of the series, the difference is substantial. The largest difference is for FacesUCR for which $\lbfast$ has tightness of 0.4639 relative to 0.2839 for $\lbfastnolr$. The average difference between the tightness of the two variants is 0.0124.

Figure \ref{fig:enhancedFast} shows the relative tightness of $\lbfast$ and $\lbenhancedfast^3$ using the optimal window size on all 60 datasets for which the optimal window size is greater than zero.  $\lbfast$ provides a tighter lower bound for every dataset.  However, the difference is always small. The largest difference is for ECG5000 for which $\lbfast$ has tightness of 0.8845 relative to 0.8724 for $\lbenhancedfast^3$. The average difference between the tightness of the two variants is 0.0008.

\begin{figure}
\begin{minipage}{0.48\columnwidth}%
\includegraphics[width=1.1\columnwidth]{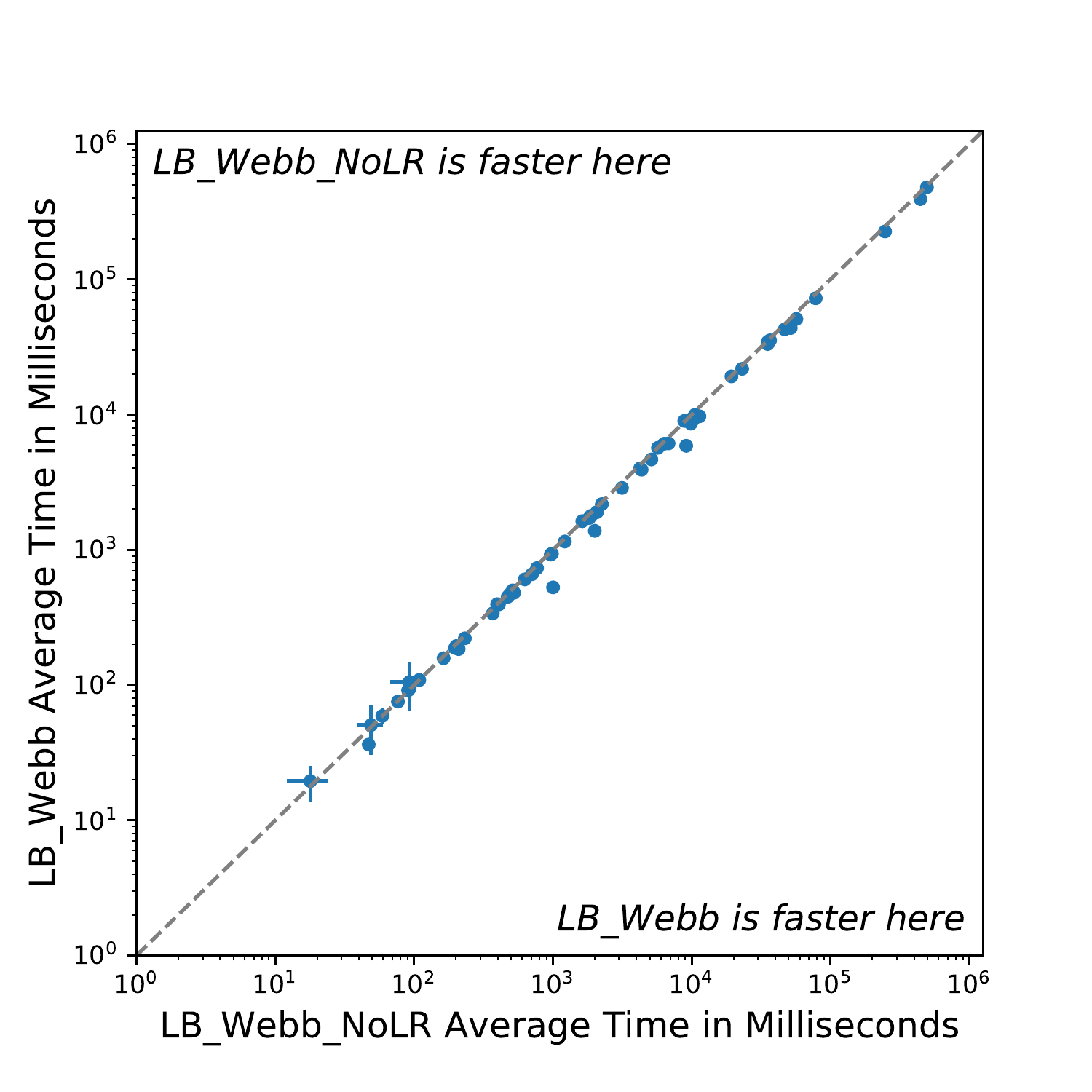}
\vspace*{-20pt}\caption{Relative compute time for nearest neighbor search in sorted order using $\lbfast$ and $\lbfastnolr$. $\lbfast$ is faster for  54  datasets and slower for 6. $\lbfast$ requires on average 35 minutes and 21 seconds to classify all 60 test sets while $\lbfastnolr$ requires 37 minutes and 20 seconds.}\label{fig:sortWebbNoLR}
\end{minipage}\hspace*{.04\columnwidth}
\begin{minipage}{0.48\columnwidth}%
\includegraphics[width=1.1\columnwidth]{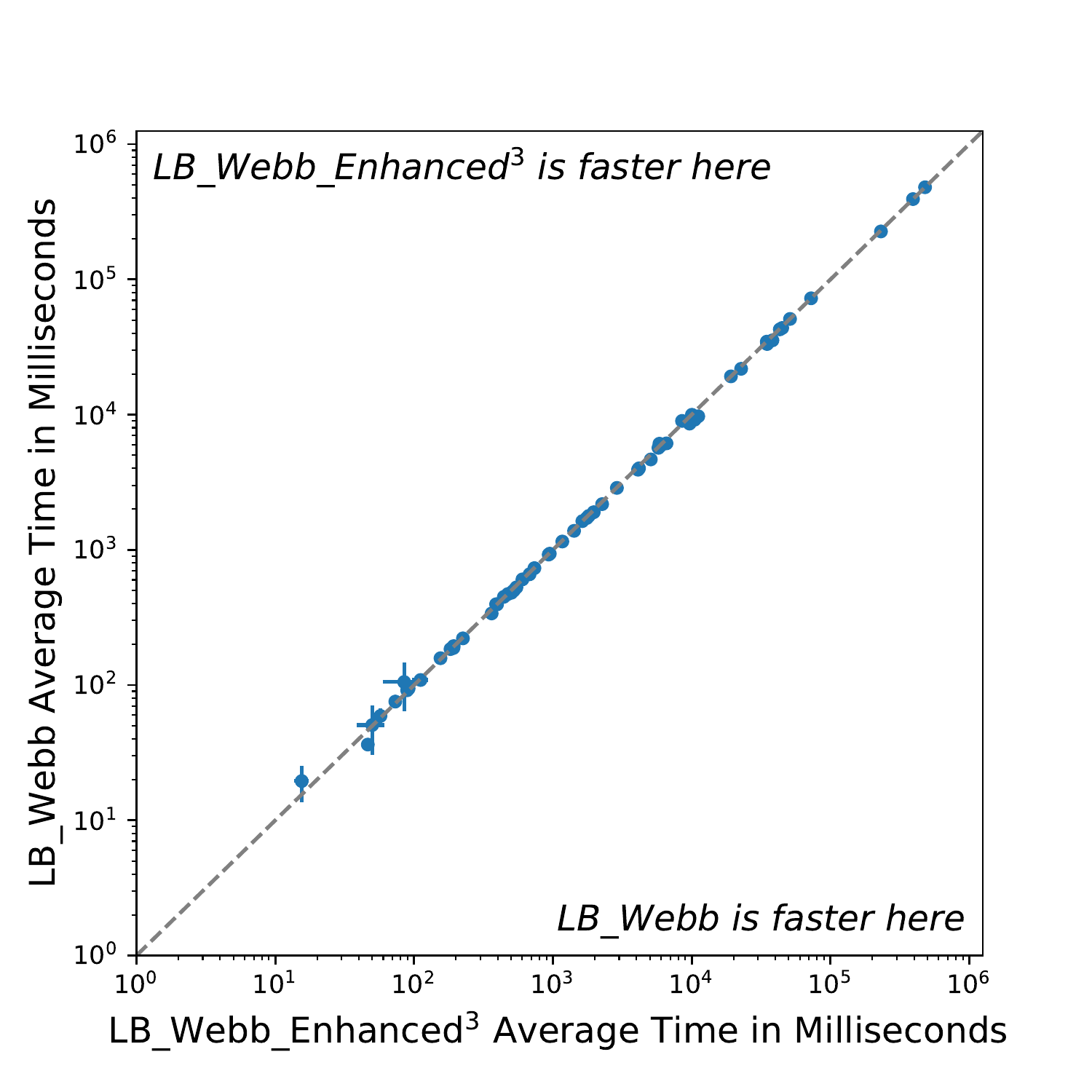}
\vspace*{-20pt}\caption{Relative compute time for nearest neighbor search in sorted order using $\lbfast$ and $\lbenhancedfast^3$. $\lbfast$ is faster for  41 datasets and slower for 19. $\lbfast$ requires on average 35 minutes and 21 seconds to classify all 60 test sets while $\lbenhancedfast^3$ requires 35 minutes and 16 seconds.}\label{fig:sortEnhancedWebb}
\end{minipage}
\end{figure}

Figure \ref{fig:sortWebbNoLR}
shows the relative time with $\lbfast$ and $\lbfastnolr$ for nearest neighbor search using the optimal window size on all 60 datasets for which the optimal window size is greater than zero.   $\lbfast$ is faster for all but 6 datasets.  However, the relative differences are mainly small.  The biggest difference is for ElectricDevices for which $\lbfast$ requires 5 minutes and 39 seconds and $\lbfastnolr$ requires 6 minutes and 17 seconds.  On average $\lbfast$ requires 35 minutes and 21 seconds to classify the entire 60 datasets while $\lbfastnolr$ requires 37 minutes and 20 seconds.
 
Figure \ref{fig:sortEnhancedWebb} shows the relative time with $\lbfast$ and $\lbenhancedfast^3$ for nearest neighbor search using the optimal window size on all 60 datasets for which the optimal window size is greater than zero.   $\lbfast$ is faster for all but 19 datasets.  However, the relative differences are all small.  The biggest difference is for UWaveGestureLibraryAll for which $\lbfast$ requires 57  seconds and $\lbenhancedfast^3$ requires 1 minute and 4 seconds.  On average $\lbfast$ requires 35 minutes and 21 seconds to classify the entire 60 datasets while $\lbenhancedfast^3$ requires 35 minutes and 16 seconds.

In summary, the addition of the left and right paths to $\lbfast$ almost invariably increases tightness. When there is substantial variation in the beginnings and endings of the series it can increase tightness substantially.  It always increases tightness relative to using left and right bands, but in practice this increase in tightness appears to have little impact. These results suggest that in some applications it might be advantageous to use $\lbenhancedfast$ rather than $\lbfast$, if an appropriate value for $k$ can be determined.

\section{Conclusions}

We have derived four new $\DTW$ lower bounds. To the best of our knowledge, $\lbtight$ is the tightest bound that has linear time complexity with respect to series length and is invariant to window size.  $\lbfast$ shares the same complexity, but provides a trade-off between efficiency and tightness that is more effective in many applications.

Both these bounds lend themselves to early abandoning. $\lbtight$ is likely to be most useful in contexts where this can be deployed, such as in a form of nearest neighbor search where the bound is calculated immediately before calculating DTW and thus can be abandoned if a closer candidate has already been encountered.

Both bounds also lend themselves to cascading.  This is a process by which successive bounds providing successive trade-offs between compute time and tightness are employed in succession.  For example, Rakthanmanon and Keogh \cite{rakthanmanon2013data} employ first the constant time $\lbkim$ \cite{kim2001index}, followed by $\lbkeogh$, followed by a second evaluation of $\lbkeogh$ with the order of the two series reversed.  Reversing the order of the two series in $\lbkeogh$ will obtain a tighter bound than applying $\lbkeogh$ in the original order in approximately 50\% of cases, as the order of the series affects the bound, but neither order is a priori superior.  Both $\lbtight$ and $\lbfast$ can be deployed in a similar manner, by first computing the constant time left and right paths, then computing the bridging $\lbkeogh$, before finally computing the additional final pass.  This cascade provides intermediate lower bounds of successive strength that build upon one another, using the value calculated for the looser bound as a starting point for the tighter one and ending with a bound that is likely to be substantially tighter than the best of $\lbkeogh$ under both orders. This is a promising direction for further research.

$\lbenhancedfast$ is a parameterized variant of $\lbfast$ that employs the left and right bands of $\lbenhanced$ in place of $\lbfast$'s left and right paths. This variant, with suitably large values of parameter $k$, may be useful in contexts where bounds based on distance to the envelope are less effective, such as when window sizes are large.

$\lbtight$ and $\lbfast$ require that $\forall_{x,y\in \mathcal{R},\mathcal{R}:A_i\leq x\leq\allowbreak y\leq B_j\vee A_i\geq x\geq y\geq B_j}\delta(A_i,B_j) \geq \delta(A_i, y)+\delta(B_j,x)-\delta(x,y)$, a condition satisfied by the two common pairwise distance measures, $\delta(A_i,B_j)=|A_i-B_j|$ and $\delta(A_i,B_j)=(A_i-B_j)^2$. A further variant, $\lbfastmonotone$, supports faster computation of $\lbfast$ when $\delta(A_i,B_j)=|A_i-B_j|$, and provides a tight lower bound for $\DTW$ so long as $\delta(A_i,B_j)$ increases monotonically with $|A_i-B_j|$, the same class of pairwise distance functions as for which $\lbkeogh$, $\lbimproved$ and $\lbenhanced$ are $\DTW$ lower bounds.

$\lbfast$ has similar tightness to $\lbimproved$, but requires substantially less computation. Our experiments show that it provides a highly effective trade-off between speed and tightness in a wide variety of contexts. 

\section{Acknowledgment}
This research has been supported by the Australian Research Council under award DP210100072. 
The authors would like to also thank Prof Eamonn Keogh and all the contributors to the UCR time series classification archive.

\bibliographystyle{elsarticle-num}
{

\begin{thebibliography}{10}
\expandafter\ifx\csname url\endcsname\relax
  \def\url#1{\texttt{#1}}\fi
\expandafter\ifx\csname urlprefix\endcsname\relax\def\urlprefix{URL }\fi
\expandafter\ifx\csname href\endcsname\relax
  \def\href#1#2{#2} \def\path#1{#1}\fi

\bibitem{sakoe1972dynamic}
H.~Sakoe, S.~Chiba, A dynamic programming approach to continuous speech
  recognition, in: International Congress on Acoustics, Vol.~3, 1971, pp.
  65--69.

\bibitem{cheng2016image}
H.~Cheng, Z.~Dai, Z.~Liu, Y.~Zhao, An image-to-class dynamic time warping
  approach for both 3d static and trajectory hand gesture recognition, Pattern
  Recognition 55 (2016) 137--147.

\bibitem{OKAWA2020107227}
M.~Okawa, Online signature verification using single-template matching with
  time-series averaging and gradient boosting, Pattern Recognition 102 (2020)
  107227.

\bibitem{yasseen2016shape}
Z.~Yasseen, A.~Verroust-Blondet, A.~Nasri, Shape matching by part alignment
  using extended chordal axis transform, Pattern Recognition 57 (2016)
  115--135.

\bibitem{singh2017smart}
G.~Singh, D.~Bansal, S.~Sofat, N.~Aggarwal, Smart patrolling: An efficient road
  surface monitoring using smartphone sensors and crowdsourcing, Pervasive and
  Mobile Computing 40 (2017) 71--88.

\bibitem{cao2016real}
Y.~Cao, N.~Rakhilin, P.~H. Gordon, X.~Shen, E.~C. Kan, A real-time spike
  classification method based on dynamic time warping for extracellular enteric
  neural recording with large waveform variability, Journal of Neuroscience
  Methods 261 (2016) 97--109.

\bibitem{varatharajan2018wearable}
R.~Varatharajan, G.~Manogaran, M.~K. Priyan, R.~Sundarasekar, Wearable sensor
  devices for early detection of alzheimer disease using dynamic time warping
  algorithm, Cluster Computing 21~(1) (2018) 681--690.

\bibitem{chavez2001searching}
E.~Ch{\'a}vez, G.~Navarro, R.~Baeza-Yates, J.~L. Marroqu{\'\i}n, Searching in
  metric spaces, ACM Computing Surveys 33~(3) (2001) 273--321.

\bibitem{lai2007fast}
J.~Z. Lai, Y.-C. Liaw, J.~Liu, Fast k-nearest-neighbor search based on
  projection and triangular inequality, Pattern Recognition 40~(2) (2007)
  351--359.

\bibitem{ratanamahatana2005three}
C.~A. Ratanamahatana, E.~Keogh, Three myths about dynamic time warping data
  mining, in: Proceedings of the 2005 SIAM International Conference on Data
  Mining, SIAM, 2005, pp. 506--510.

\bibitem{keogh2005exact}
E.~Keogh, C.~A. Ratanamahatana, Exact indexing of dynamic time warping,
  Knowledge and Information Systems 7~(3) (2005) 358--386.

\bibitem{kim2001index}
S.-W. Kim, S.~Park, W.~W. Chu, An index-based approach for similarity search
  supporting time warping in large sequence databases, in: 17th International
  Conference on Data Engineering., IEEE, 2001, pp. 607--614.

\bibitem{lemire2009faster}
D.~Lemire, Faster retrieval with a two-pass dynamic-time-warping lower bound,
  Pattern Recognition 42~(9) (2009) 2169--2180.

\bibitem{shen2018accelerating}
Y.~Shen, Y.~Chen, E.~Keogh, H.~Jin, Accelerating time series searching with
  large uniform scaling, in: Proceedings of the 2018 SIAM International
  Conference on Data Mining, SIAM, 2018, pp. 234--242.

\bibitem{yi1998efficient}
B.-K. Yi, H.~Jagadish, C.~Faloutsos, Efficient retrieval of similar time
  sequences under time warping, in: Data Engineering, 1998. Proceedings., 14th
  International Conference on, IEEE, 1998, pp. 201--208.

\bibitem{TanEtAl19}
C.~W. Tan, F.~Petitjean, G.~I. Webb, Elastic bands across the path: A new
  framework and methods to lower bound dtw, in: Proceedings of the 2019 SIAM
  International Conference on Data Mining, 2019, pp. 522--530.

\bibitem{bagnall2017great}
A.~Bagnall, J.~Lines, A.~Bostrom, J.~Large, E.~Keogh, The great time series
  classification bake off: a review and experimental evaluation of recent
  algorithmic advances, Data Mining and Knowledge Discovery 31~(3) (2017)
  606--660.

\bibitem{ucrarchive}
H.~A. Dau, A.~Bagnall, K.~Kamgar, C.-C.~M. Yeh, Y.~Zhu, S.~Gharghabi, C.~A.
  Ratanamahatana, E.~Keogh, The {UCR} time series archive, IEEE/CAA Journal of
  Automatica Sinica 6~(6) (2019) 1293--1305.

\bibitem{rakthanmanon2013data}
T.~Rakthanmanon, E.~Keogh, Data mining a trillion time series subsequences
  under dynamic time warping, in: Twenty-Third International Joint Conference
  on Artificial Intelligence, 2013, pp. 3047--3051.

\end{thebibliography}

}

\end{document}